\documentclass{article}

\usepackage{microtype}
\usepackage{graphicx}
\usepackage{hyperref}
\usepackage{subfigure}
\usepackage{booktabs} 
\usepackage{amsmath,amsfonts,amsthm}
\usepackage{multirow}
\usepackage{multicol}
\usepackage{amssymb}
\usepackage{stackengine}
\usepackage{scalerel}
\usepackage{cleveref}
\usepackage{textcomp}
\usepackage{enumitem}

\usepackage{pifont}
\newcommand{\xmark}{\ding{55}}%

\newcommand{\mt}{\widetilde{M}}
\newcommand{\E}{\mathbb{E}}
\newcommand{\R}{\mathbb{R}}
\newcommand{\mc}{\mathcal}
\newcommand{\mb}{\boldsymbol}

\DeclareMathOperator*{\argmin}{arg\,min}

\def\transpose{\top} 

\newtheorem{theorem}{Theorem}

\newtheorem{proposition}{Proposition}
\newtheorem{assumption}{Assumption}
\newtheorem{corollary}{Corollary}

\def\0{\boldsymbol{0}}
\def\1{\boldsymbol{1}}

\def\x{\boldsymbol{x}}

\usepackage[accepted]{icml2020}

\icmltitlerunning{Rethinking Bias-Variance Trade-off for Generalization of Neural Networks}

\usepackage[normalem]{ulem}

\newcommand{\scrcmd}{\textsuperscript}
\newcommand{\scr}[1]{\scrcmd{#1}}

\begin{document}
\twocolumn[
\icmltitle{Rethinking Bias-Variance Trade-off for Generalization of Neural Networks}

\icmlsetsymbol{equal}{*}

\begin{icmlauthorlist}
\icmlauthor{Zitong Yang}{equal,to}
\icmlauthor{Yaodong Yu}{equal,to}
\icmlauthor{Chong You}{to}
\icmlauthor{Jacob Steinhardt}{to,goo}
\icmlauthor{Yi Ma}{to}
\end{icmlauthorlist}

\icmlaffiliation{to}{Department of Electrical Engineering and Computer Sciences, University of California, Berkeley.}
\icmlaffiliation{goo}{Department of Statistics, University of California, Berkeley}

\icmlcorrespondingauthor{Zitong Yang}{zitong@berkeley.edu}
\icmlcorrespondingauthor{Yaodong Yu}{yyu@eecs.berkeley.edu}
\icmlkeywords{Machine Learning, ICML}

\vskip 0.3in
]
\newpage

\printAffiliationsAndNotice{\icmlEqualContribution}
\begin{abstract}
The classical bias-variance trade-off predicts that bias decreases and variance increases with model complexity, leading to a U-shaped risk curve.
Recent work calls this into question for neural networks and other over-parameterized models, for which it is often observed that larger models generalize better.
We provide a simple explanation for this by measuring the bias and variance of neural networks: while the bias is {\em monotonically decreasing} as in the classical theory, the variance is {\em unimodal} or {\em bell-shaped}: it increases then decreases with the width of the network. 
We vary the network architecture, loss function, and choice of dataset and confirm that variance unimodality occurs robustly for all models we considered.
The risk curve is the sum of the bias and variance curves and displays different qualitative shapes depending on the relative scale of bias and variance, with the double descent curve observed in recent literature as a special case.
We corroborate these empirical results with a theoretical analysis of two-layer linear networks with random first layer.
Finally, evaluation on out-of-distribution data shows that most of the drop in accuracy comes from increased bias while variance increases by a relatively small amount.
Moreover, we find that deeper models decrease bias and increase variance for both in-distribution and out-of-distribution data.
\vskip -0.1in
\end{abstract}

\section{Introduction}
\vskip -0.05in
\begin{figure*}[ht]
  \begin{center}
    \subfigure[\label{fig:Mvariance}Case 1]{\includegraphics[width=0.33\textwidth]{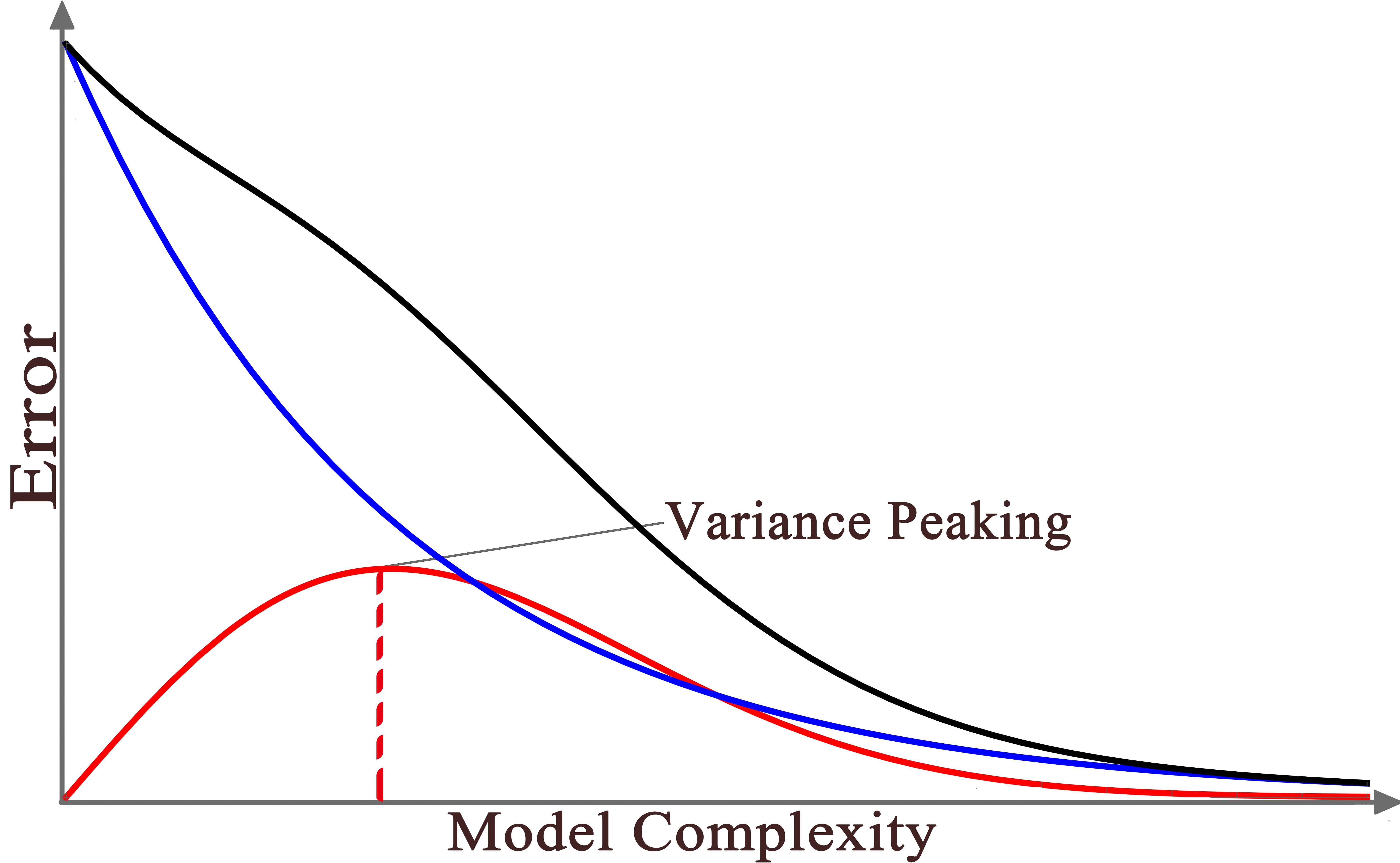}}
    \subfigure[\label{fig:Lvariance}Case 2]{\includegraphics[width=0.33\textwidth]{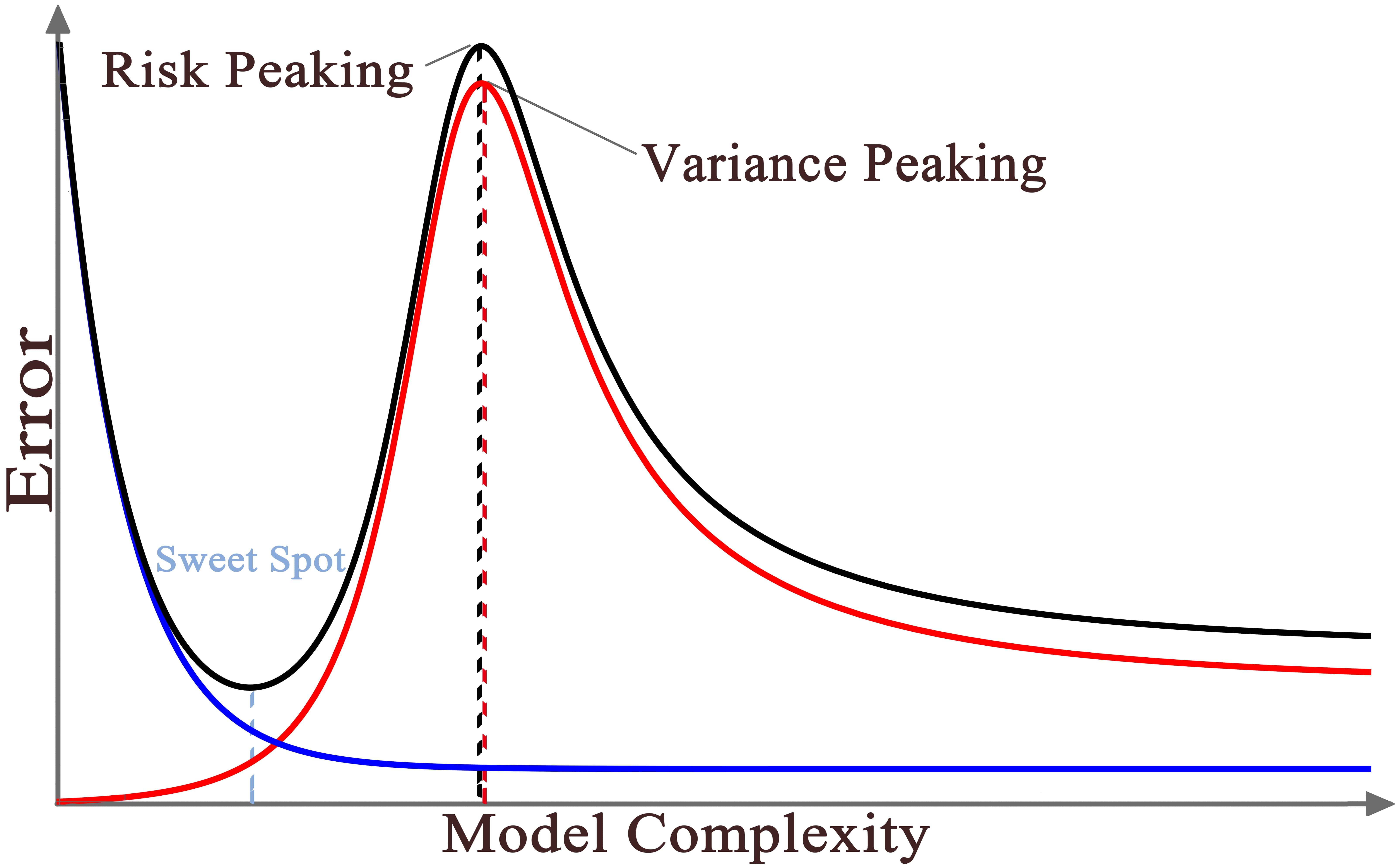}}
    \subfigure[\label{fig:Svariance}Case 3]{\includegraphics[width=0.33\textwidth]{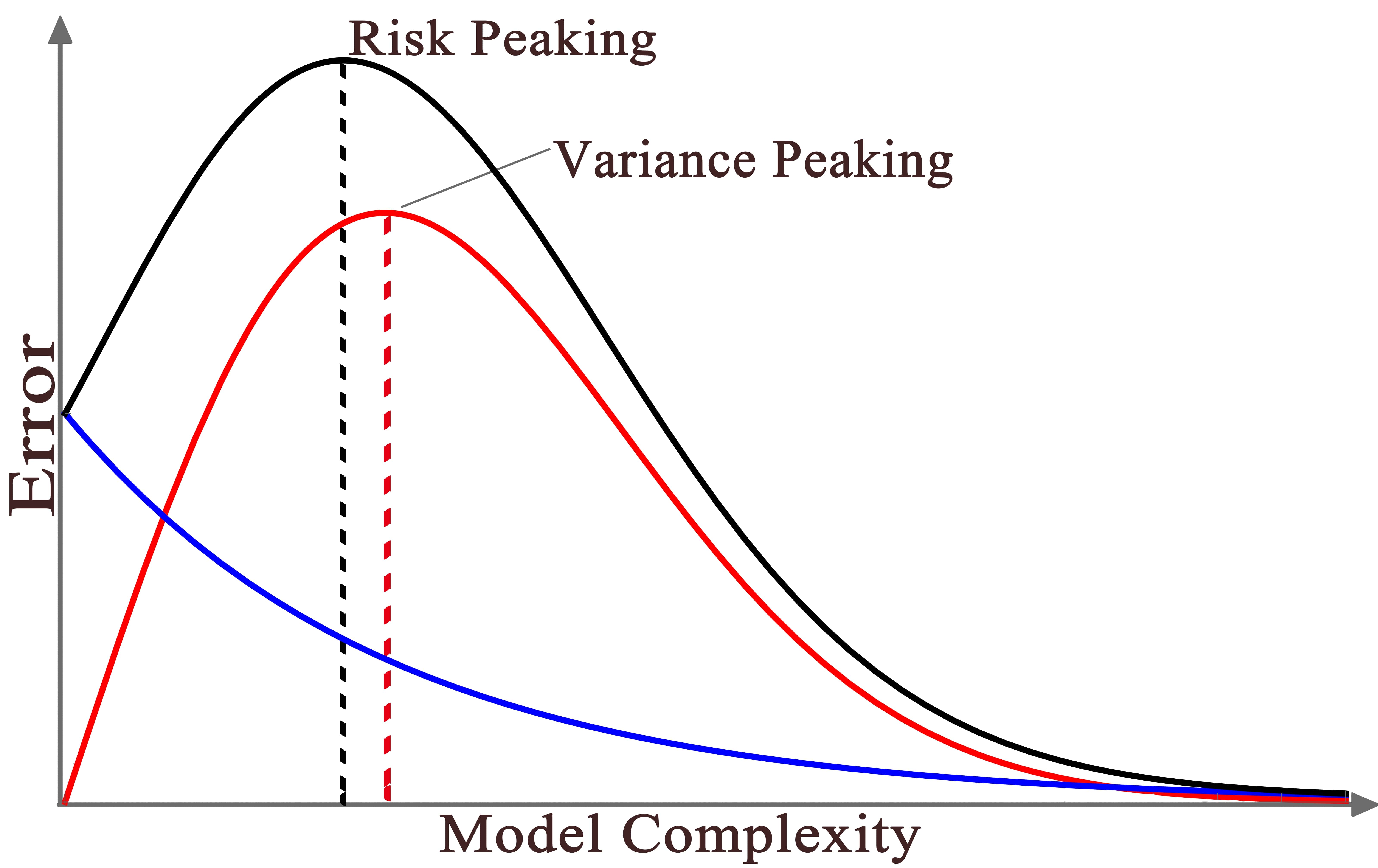}}
      \vskip -0.15in
    \caption{Typical cases of expected risk curve (in black) in neural networks. Blue: squared bias curve. Red: variance curve.}
    \label{fig:conceptualdiagram}
  \end{center}
  \vskip -0.2in
\end{figure*}

Bias-variance trade-off is a fundamental principle for understanding the generalization of predictive learning models \cite{hastie01statisticallearning}.
The {\em bias} is an error term that stems from a mismatch between the model class and the underlying data distribution, and is typically monotonically non-increasing as a function of the complexity of the model. The {\em variance} measures sensitivity to fluctuations in the training set and is often attributed to a large number of model parameters.
Classical wisdom predicts that model variance increases and bias decreases {\em monotonically} with model complexity \cite{geman1992neural}.
Under this perspective, we should seek a model that has neither too little nor too much capacity and achieves the best trade-off between bias and variance. 

In contrast, modern practice for neural networks repeatedly demonstrates the benefit of increasing the number of neurons~\cite{krizhevsky2012imagenet, simonyan2014very, zhang2017understanding}, even up to the point of saturating available memory.
{The inconsistency between classical theory and modern practices suggests that some arguments in the classical theory can not be applied to modern neural networks.}

\citet{geman1992neural} first studied the bias and variance of the neural networks and give experimental evidence that the variance is indeed increasing as the width of the neural network increases.
Since \citet{geman1992neural}, \citet{Neal:arxiv18} first experimentally measured the variance of modern neural network architectures and shown that the variance can actually be decreasing as the width increases to a highly overparameterized regime.
Recently, \citet{belkindouble, belkin2018understand, belkin2019does} directly studied the risk of modern machine learning models and proposed a {\em double descent} risk curve, which has also been analytically characterized for certain regression and classification models \cite{meisong, trevor18, Spigler:IOP19, deng2019model, Advani2017HighdimensionalDO, Bartlett201907378, chatterji2020finitesample}. 
However, there exists two mysteries around the double descent risk curve. 
First, the double descent phenomenon can not be robustly observed~\cite{nakkiran2019deep, Ba2020Generalization}. In particular, to observe it in modern neural network architectures, we sometimes have to artificially inject label noise~\cite{nakkiran2019deep}. 
Second, there lacks an explanation for {\em why} the double descent risk curve should occur. 
In this work, we offer an simple explanation for these two mysteries by proposing an unexpected {\em unimodal} variance curve.

Specifically, we measure the bias and variance of modern deep neural networks trained on commonly used computer vision datasets. 
Our main finding is that while the bias is {monotonically decreasing} with network width as in the classical theory, the variance curve is \emph{unimodal} or \emph{bell-shaped}: it first increases and then decreases~(see Figure \ref{fig:mainline}). 
Therefore, the unimodal variance is consistent with the finding of \citet{Neal:arxiv18}, who observed that the variance eventually decreases in the overparameterized regime. In particular, the unimodal variance curve can also be observed in \citet[Figure 1, 2, 3]{Neal:arxiv18}. However, \citet{Neal:arxiv18} did not point out the characteristic shape of the variance or connect it to double descent. More importantly, we demonstrate that the unimodal variance phenomenon can be robustly observed for varying network architecture and dataset.
Moreover, by using a generalized bias-variance decomposition for Bregman divergences~\cite{pfau2013bergman}, we verify that it occurs for both squared loss and cross-entropy loss.

This unimodal variance phenomenon initially appears to contradict recent theoretical work suggesting that both bias and variance are non-monotonic and exhibit a peak in some regimes~\cite{meisong, trevor18} . 
The difference is that this previous work considered the \emph{fixed-design} bias and variance, while we measure the \emph{random-design} bias and variance (we describe the differences in detail in \textsection\ref{section:background}). 
Prior to our work, \citet{nakkiran2019more} also considered the variance of linear regression in the random-design setting, and \citet{RossetTib2017RandomFixed} discussed additional ways to decompose risk into the bias and the variance term.

A key finding of our work is that the complex behavior of the risk curve arises due to the simple but non-classical variance unimodality phenomenon.
Indeed, since the expected risk (test loss) is the sum of bias and variance, monotonic bias and unimodal variance can lead to three characteristic behaviors, illustrated in Figure~\ref{fig:conceptualdiagram}, depending on the relative size of the bias and variance. 
If the bias completely dominates, we obtain monotonically decreasing risk curve~(see Figure~\ref{fig:Mvariance}). 
Meanwhile, if the variance dominates, we obtain a bell-shaped risk curve that first increases then decreases (see Figure~\ref{fig:Svariance}). 
The most complex behavior is if bias and variance dominate in different regimes, leading to the double-descent 
risk curve in Figure~\ref{fig:Lvariance}. 
All three behaviors are well-aligned with the empirical observation in deep learning that larger models typically perform better. The most common behavior in our experiments is the first case (monotonically decreasing risk curve) as bias is typically larger than variance. {We can observe the double-descent risk curve when label noise is added to the training set (see \textsection\ref{section:doubledescent}), and can observe the unimodal risk curve when we use the generalized bias-variance decomposition for cross-entropy loss (see \textsection\ref{section:phenomenonrobust}).}

\paragraph{Further Implications.} 
The investigations described above characterize bias and variance as a function of network width, but we can explore the dependence on other quantities as well, such as model depth (\textsection\ref{section:depth}). Indeed, we find that deeper models tend to have lower bias but higher variance. 
Since bias is larger at current model sizes, this confirms the prevailing wisdom that we should generally use deeper models when possible. 
On the other hand, it suggests that this process may have a limit---eventually very deep models may have low bias but high variance such that increasing the depth further harms performance.

We also investigate the commonly observed drop in accuracy for models evaluated on out-of-distribution data, and attribute it primarily to increased bias. 
Combined with the previous observation, this suggests that increasing model depth may help combat the drop in out-of-distribution accuracy, which is supported by experimental findings in \citet{hendrycks2019benchmarking}.

\paragraph{Theoretical Analysis of A Two-Layer Neural Network.} 
Finally, we conduct a theoretical study of a two-layer linear network with a random Gaussian first layer. While this model is much simpler than those used in practice, we nevertheless observe the same characteristic behaviors for the bias and variance. 
In particular, by working in the asymptotic setting where the input data dimension, amount of training data, and network width go to infinity with fixed ratios, we show that the bias is monotonically decreasing while the variance curve is unimodal.  
Our analysis also characterizes the location of the variance peak as the point where the number of hidden neurons is {approximately half} of the dimension of the input data.

\section{Preliminaries}
In this section we present the bias-variance decomposition for squared loss. We also present a generalized bias-variance decomposition for cross-entropy loss in \textsection\ref{section:prelim_estimating}. 
The task is to learn a function $f:\R^d\rightarrow\R^c,$ based on i.i.d. training samples $\mc{T}=\{(\mb x_i,\mb y_i)\}_{i=1}^n$ drawn from a joint distribution $P$ on $\R^d\times\R^c$, such that the mean squared error $\E_{\mb x, \mb y}\left[\|\mb y-f(\mb x, \mc{T})\|_2^2\right]$ is minimal, where $(\mb x, \mb y)\sim P$.
Here we denote the learned function by $f(\mb x; \mc{T})$ to make the dependence on the training samples clear.

Note that the learned predictor $f(\mb x; \mc{T})$ is a random quantity depending on $\mc{T}$. 
We can assess its performance in two different ways. 
The first way, random-design, takes the expectation over $\mc{T}$ such that we consider the expected error $\E_{\mc T}\left[\|\mb y-f(\mb x, \mc{T})\|_2^2\right]$. 
The second way, fixed-design, holds the training covariates $\{\mb x_i\}_{i=1}^n$ fixed and only takes expectation over $\{\mb y_i\}_{i=1}^n$, i.e., $\E_{\mc T}\left[\|\mb y-f(\mb x, \mc{T})\|_2^2\mid\{\mb x_i\}_{i=1}^n\right]$. The choice of random/fixed-design leads to different bias-variance decompositions. Throughout the paper, we focus on random-design, as opposed to fixed-design studied in \citet{meisong, trevor18, Ba2020Generalization}.

\subsection{Bias Variance Decomposition}\label{section:background}
\paragraph{Random Design.} 
In the random-design setting, decomposing the quantity $\E_{\mc T}\left[\|\mb y-f(\mb x, \mc{T})\|_2^2\right]$ gives the usual bias-variance trade-off from machine learning, e.g. \citet{geman1992neural, hastie01statisticallearning}.
\begin{align*}
  &\E_{\mb x, \mb y}\E_{\mc T}\left[\|\mb y-f(\mb x, \mc{T})\|_2^2\right] =\\ &\underbrace{\E_{\mb x, \mb y}\left[\|\mb y-\bar{f}(\mb x)\|_2^2\right]}_{\textbf{Bias}^2} + \underbrace{\E_{\mb x}\E_{\mc T}\left[\|f(\mb x, \mc T)-\bar{f}(\mb x)\|_2^2\right]}_{\textbf{Variance}},
\end{align*}
where $\bar{f}(\mb x) = \E_{\mc T}f(\mb x, \mc T)$. Here $\E_{\mc T}\left[\|(\mb y-f(\mb x, \mc{T})\|_2^2\right]$ measures the average prediction error over different realizations of the training sample. In addition to take the expectation $\E_{\mc T}$, we also average over $\E_{\mb x, \mb y}$, as discussed in \citet[\textsection 3.2]{bishop:2006:PRML}. For future reference, we define
\begin{align}
  \textbf{Bias}^2 &= \E_{\mb x, \mb y}\left[\|\mb y-\bar{f}(\mb x)\|_2^2\right]\label{eq:bias},\\
  \textbf{Variance} &= \E_{\mb x}\E_{\mc T}\left[\|f(\mb x, \mc T)-\bar{f}(\mb x)\|_2^2\right].\label{eq:variance}
\end{align}
In \textsection\ref{section:prelim_estimating}, we present our estimator for bias and variance in equation \eqref{eq:bias} and \eqref{eq:variance}. 

\paragraph{Fixed Design.} 
In fixed-design setting, the covariates $\{\mb x_i\}_{i=1}^n$ are held be fixed, and the only randomness in the training set $\mc T$ comes from $\mb y_i \sim P(\mb Y\mid \mb X=\mb x_i)$. 
As presented in \citet{meisong, trevor18, Ba2020Generalization}, a more natural way to present the fixed-design assumption is to hold $\{\mb x_i\}_{i=1}^n$ to be fixed and let $\mb y_i = f_0(\mb x)+\mb \epsilon_i$ for $i=1,\dots, n$, where $f_0(\mb x)$ is a ground-truth function and $\mb \epsilon_i$ are random noises. 
Under this assumption, the randomness in $\mc T$ all comes from the random noise $\mb \epsilon_i$. To make this clear, we write $\mc T$ as $\mc T_{\mb\epsilon_i}$.
Then, we obtain the \emph{fixed-design} bias-variance decomposition
\begin{align*}
  &\E_{\mb\epsilon_i}\left[\|(\mb y-f(\mb x, \mc T_{\mb\epsilon_i})\|_2^2\right] =\\ &\underbrace{\left[\|(\mb y-\bar{f}(\mb x)\|_2^2\right]}_{\textbf{Bias}^2} + \underbrace{\E_{\mb \epsilon_i}\left[\|(f(\mb x, \mc T_{\mb\epsilon_i})-\bar{f}(\mb x)\|_2^2\right]}_{\textbf{Variance}},
\end{align*}
where $\bar{f}(\mb x) = \E_{\mb \epsilon_i}f(\mb x, \mc T_{\mb\epsilon_i})$. In most practical settings, the expectation $\E_{\mb \epsilon_i}f(\mb x, \mc T_{\mb\epsilon_i})$ {\em cannot be estimated} from training samples $\mc T = \{(\mb x_i, \mb y_i)\}_{i=1}^n$, because we do not have access to independent copies of $f(\mb x_i)+\mb\epsilon_i$.
In comparison to the random-design setting, the fixed-design setting tends to have larger bias and smaller variance, since more ``randomness'' is introduced into the variance term.

\subsection{Estimating Bias and Variance} \label{section:prelim_estimating}
In this section, we present the estimator we use to estimate the bias and variance as defined in equation \eqref{eq:bias} and \eqref{eq:variance}. 
The high level idea is to approximate the expectation $\E_{\mc T}$ by computing the sample average using multiple training sets $\mc T_1, \dots, \mc T_N$.
When evaluating the expectation $\E_{\mc T}$, there is a trade-off between having larger training sets ($n$) within each training set and having larger number of splits ($N$), since $n\times N=$ total number of training samples.

\paragraph{Mean Squared Error (MSE).} 
To estimate bias and variance in equation \eqref{eq:bias} and \eqref{eq:variance}, we introduce an {\em unbiased} estimator for variance, and obtain bias by subtracting the variance from the risk. 
Let $\mc{T} = \mc{T}_1\cup \cdots \cup\mc{T}_N$ be a random disjoint split of training samples.
In our experiment, we mainly take $N=2$ (for CIFAR10 each $\mc{T}_i$ has 25k samples). 
To estimate the variance, we use the unbiased estimator
\begin{equation*}
    \widehat{\textsf{var}}(\mb x, \mc T) = \frac{1}{N-1}\sum_{j=1}^N \Big\| f(\mb x, \mc{T}_j)- {
    \sum_{j=1}^N \frac{1}{N}f(\mb x, \mc{T}_j)}
    \Big\|_2^2, 
\end{equation*}
where $\textsf{var}$ depends on the test point $\mb x$ and on the random training set $\mc T$.
While $\textsf{var}$ is unbiased, its 
variance can be reduced by using multiple random splits to obtain estimators $\widehat{\textsf{var}}_1, \dots, \widehat{\textsf{var}}_k$ and taking their average. This reduces the variance of the variance estimator since:
\begin{align*}
    \text{Var}_{\mc T}\Big(\frac{1}{k}\sum_{i=1}^k \widehat{\textsf{var}}_i\Big) =\frac{\sum_{ij}\text{Cov}_{\mc T}(\widehat{\textsf{var}}_i, \widehat{\textsf{var}}_j)}{k^2}\leq \text{Var}_{\mc T}(\widehat{\textsf{var}}_1),
\end{align*}
where the $\{\widehat{\textsf{var}}_i\}_{i=1}^{k}$ are identically distributed but not independent, and we used the Cauchy-Schwarz inequality. 

\paragraph{Cross-Entropy Loss (CE).} 
In addition to the classical bias-variance decomposition for MSE loss, we also consider a generalized bias-variance decomposition for cross-entropy loss.
Let $\pi(\mb x, \mc T) \in \R^c$ be the output of the neural network (a probability distribution over the class labels).
$\pi(\mb x, \mc T)$ is a random variable since the training set $\mc{T}$ is random.
Let $\pi_0(\mb x)\in\R^c$ be the one-hot encoding of the ground-truth label.
Then, omitting the dependence of $\pi$ and $\pi_0$ on $\mb x$ and $\mc T$, the cross entropy loss
\begin{equation*}
    H(\pi_0, \pi) = \sum_{l=1}^{c} \pi_0[l]\log(\pi[l])
\end{equation*}
can be decomposed as 
\begin{algorithm}[t]\label{algorithm:bergvar}
   \caption{Estimating Generalized Variance}
   \label{alg:example}
\begin{algorithmic}
 \STATE {\bfseries Input:} Test point $\mb x$, Training set $\mc T$.
 \FOR{$i=1$ {\bfseries to} $k$}
 \STATE{Split the $\mc T$ into $\mc T_1^{(i)}, \dots, \mc T_N^{(i)}$.}
   \FOR{$j=1$ {\bfseries to} $N$}
   \STATE{Train the model using $\mc T_j^{(i)}$;}
   \STATE{Evaluate the model at $\mb x$; call the result $\pi_j^{(i)}$;}
   \ENDFOR
 \ENDFOR
 \STATE{Compute $\widehat{\pi} = \exp\left\{\frac{1}{N\cdot k} \sum_{ij} \log\left(\pi_j^{(i)}\right)\right\}$}
 \STATE{(\textit{using element-wise log and exp; $\widehat{\pi}$ estimates $\bar{\pi}$}).}
 \STATE{Normalize $\widehat{\pi}$ to get a probability distribution.}
 \STATE{Compute the variance $\frac{1}{N\cdot k} \sum_{ij} D_{\text{KL}}\left(\widehat{\pi}\Vert \pi_j^{(i)}\right)$.}
\end{algorithmic}
\end{algorithm}
  \vskip -0.2in
  
\begin{align}
\label{eq:generalized-variance}
    \E_{\mc{T}}\left[H(\pi_0, \pi)\right] =  \underbrace{D_\text{KL}(\pi_0\Vert\bar{\pi})}_{\textbf{Bias}^2} + \underbrace{\E_{\mc{T}} \left[D_\text{KL}(\bar{\pi}\Vert\pi)\right]}_{\textbf{Variance}},
\end{align}
where $\pi[l]$ is the $l$-th element of  $\pi$, and $\bar{\pi}$ is the average of log-probability after normalization, i.e.,
\begin{equation*}
  \bar{\pi}[l] \propto \exp\{\E_{\mc T}\log(\pi[l])\} \text{\,\, for \,\,} l = 1, \ldots, c.
\end{equation*}
This decomposition is a special case of the general decomposition for Bregman divergence discussed in \citet{pfau2013bergman}.

We apply Algorithm \ref{alg:example} to estimate the generalized variance in \eqref{eq:generalized-variance}. Here we could not obtain an unbiased estimator, but the estimate is better if we take more random splits (larger $k$). {In practice, we choose $k$ to be large enough so that the estimated variance stabilizes when we further increase $k$ (see \textsection\ref{section:error}).}
Similar to the case of squared loss, we estimate the bias by subtracting the variance from the risk.

\section{Measuring Bias and Variance for Neural Networks}\label{section:bvfornn}
\begin{figure*}[ht]
  \begin{center}
    \subfigure{
    \includegraphics[width=.31\textwidth]{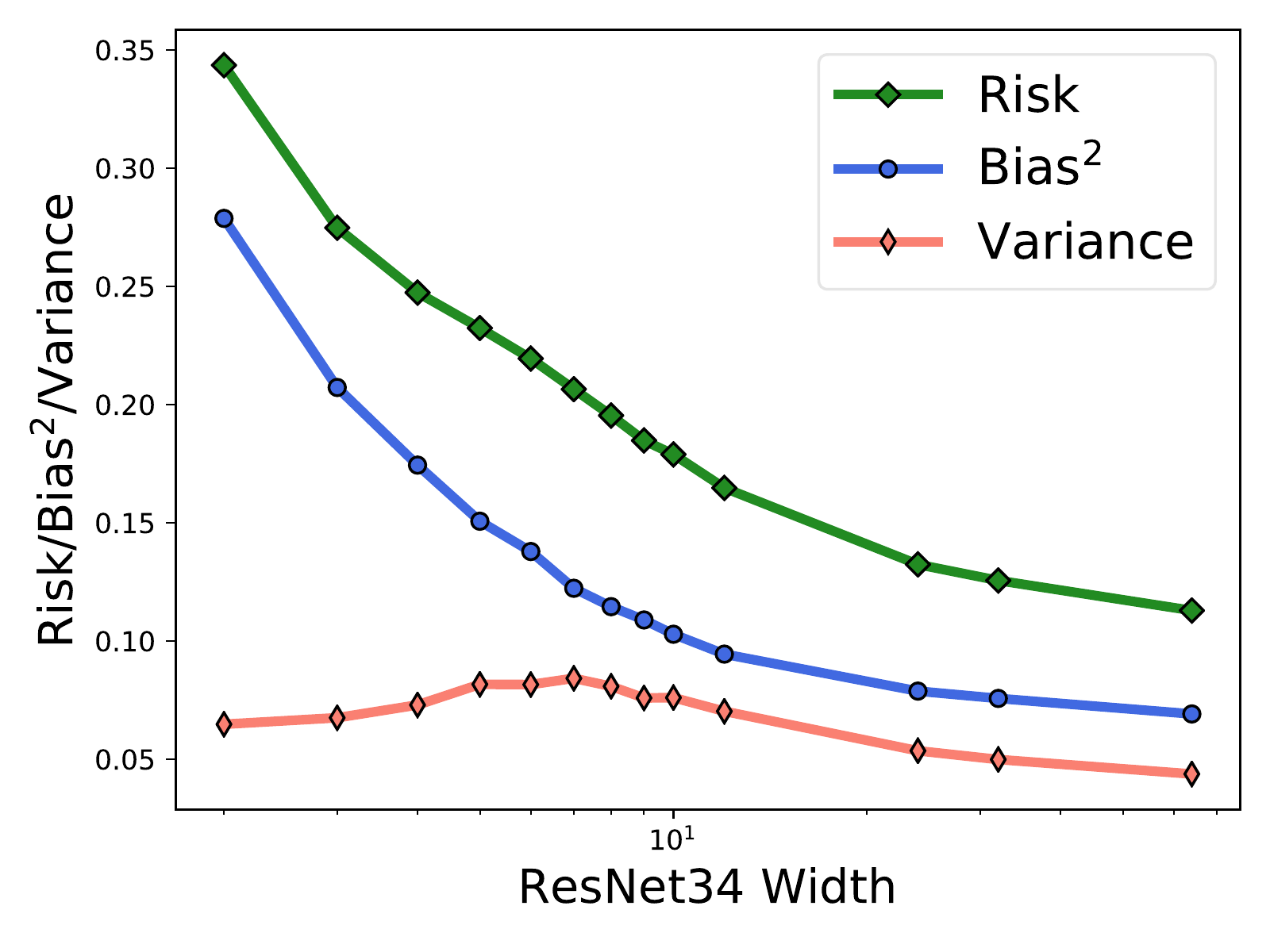}
    }
    \subfigure{
    \includegraphics[width=.31\textwidth]{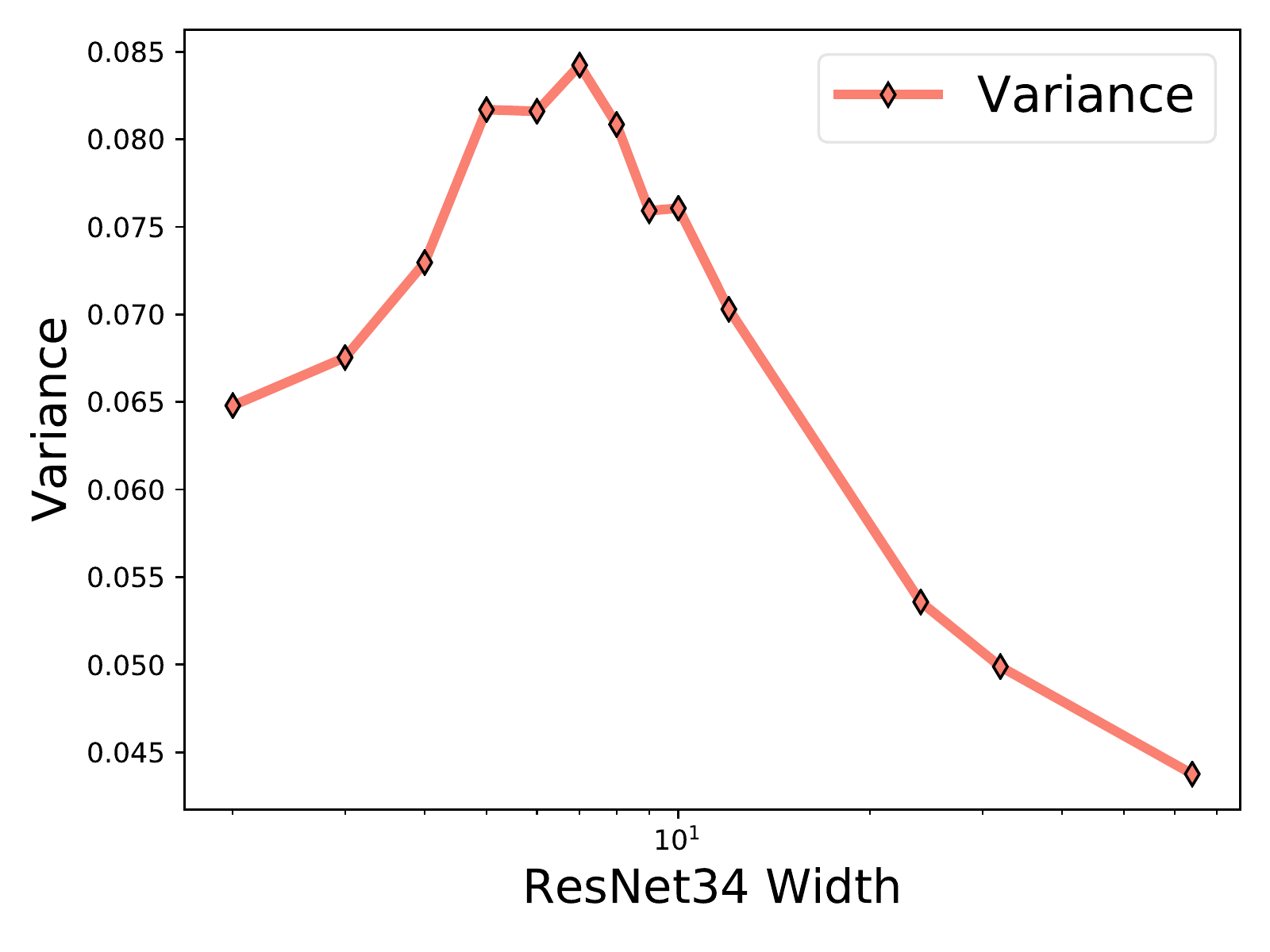}
    }
    \subfigure{
    \includegraphics[width=.31\textwidth]{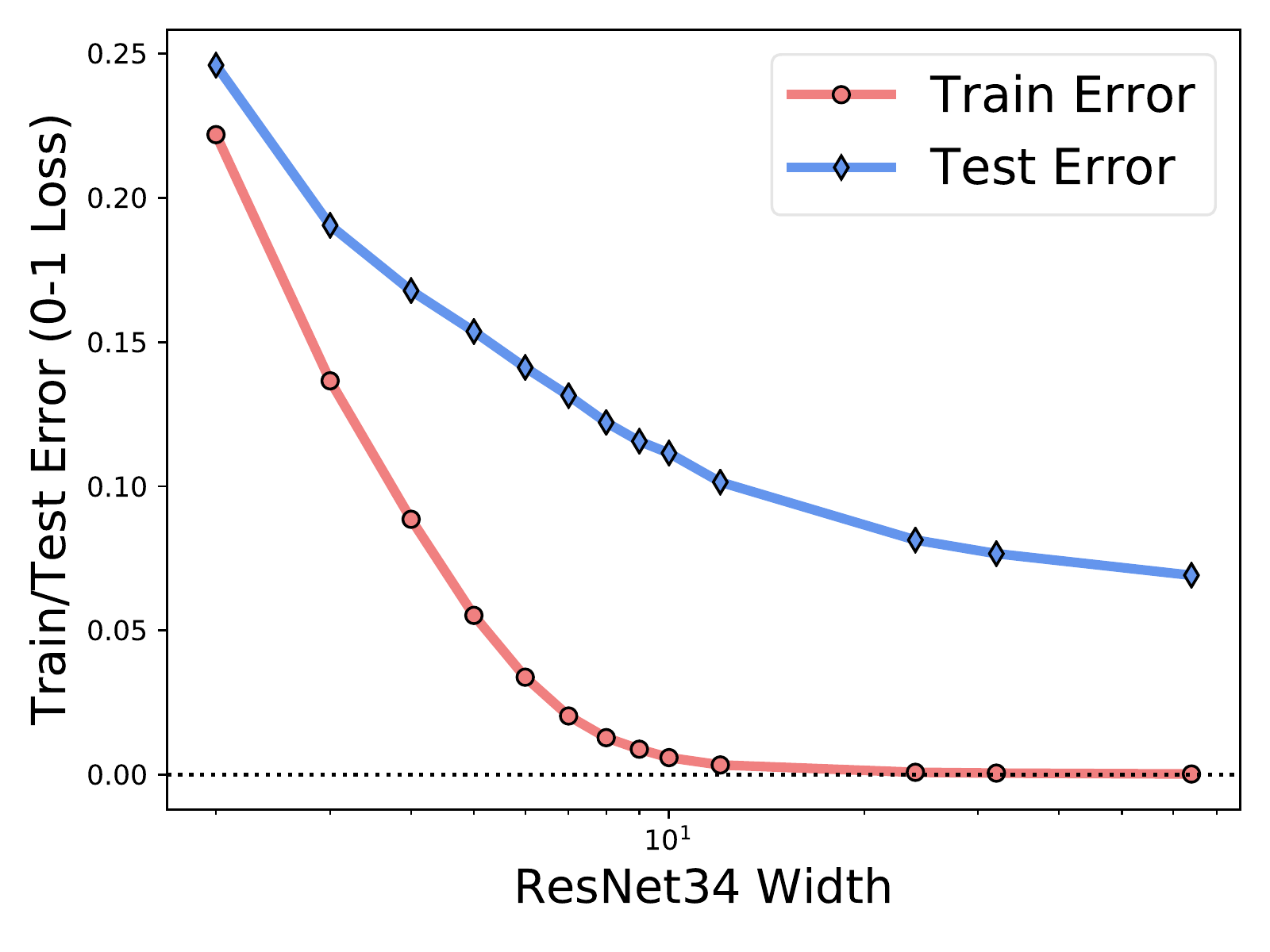}
    }
      \vskip -0.2in
    \caption{Mainline experiment on ResNet34, CIFAR10 dataset (25,000 training samples). (\textbf{Left}) Risk, bias, and variance for ResNet34. (\textbf{Middle}) Variance for ResNet34. (\textbf{Right}) Train error and test error for ResNet34. 
    }
  \label{fig:mainline}
  \end{center}
  \vskip -0.2in
\end{figure*}

\begin{figure*}[ht]
  \begin{center}
    \subfigure[\label{fig:arch}ResNext29, MSE loss, CIFAR10]{\includegraphics[width=0.31\textwidth]{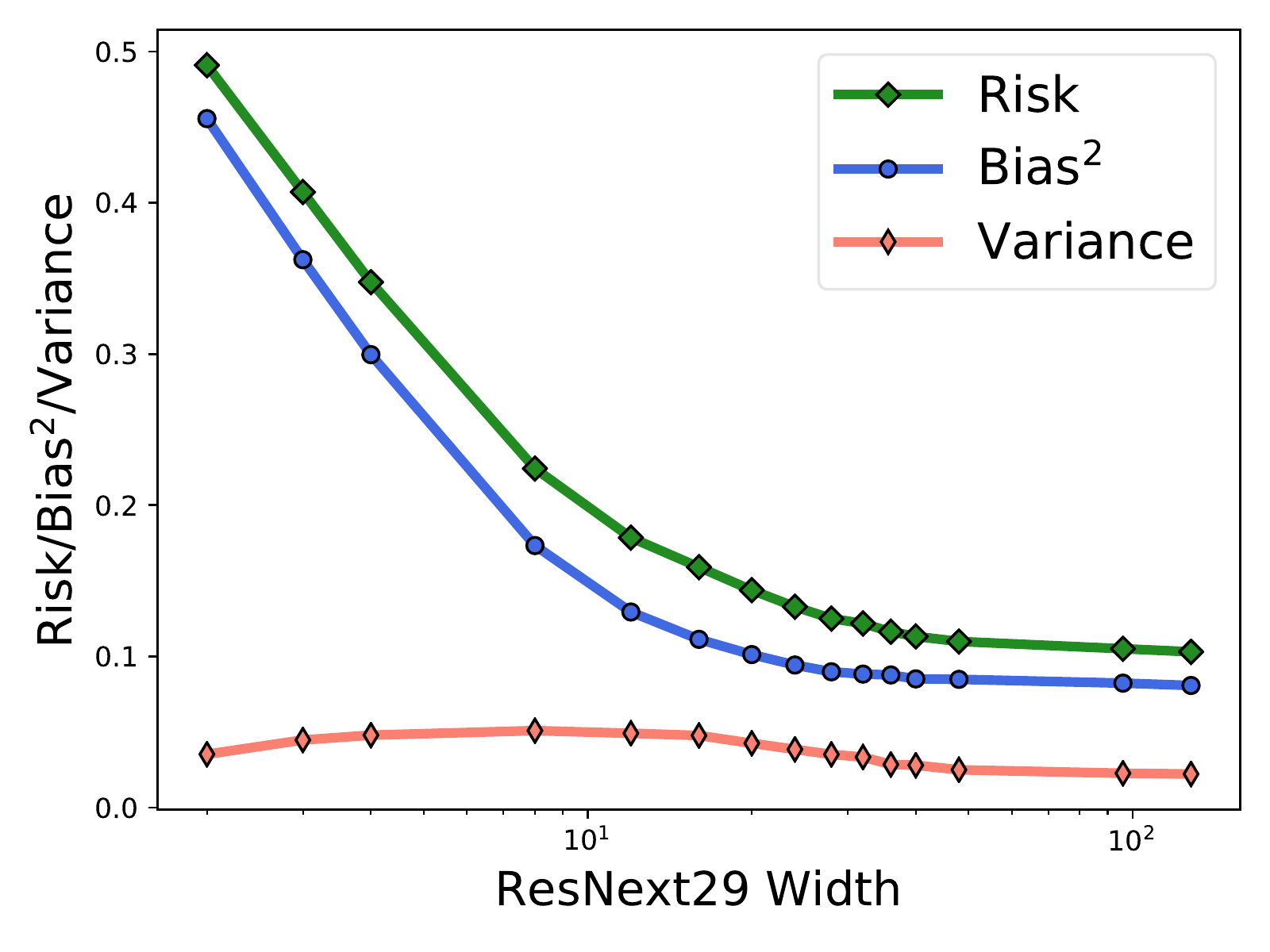}}
    \subfigure[\label{fig:loss}ResNet34, CE loss, CIFAR10]{\includegraphics[width=0.31\textwidth]{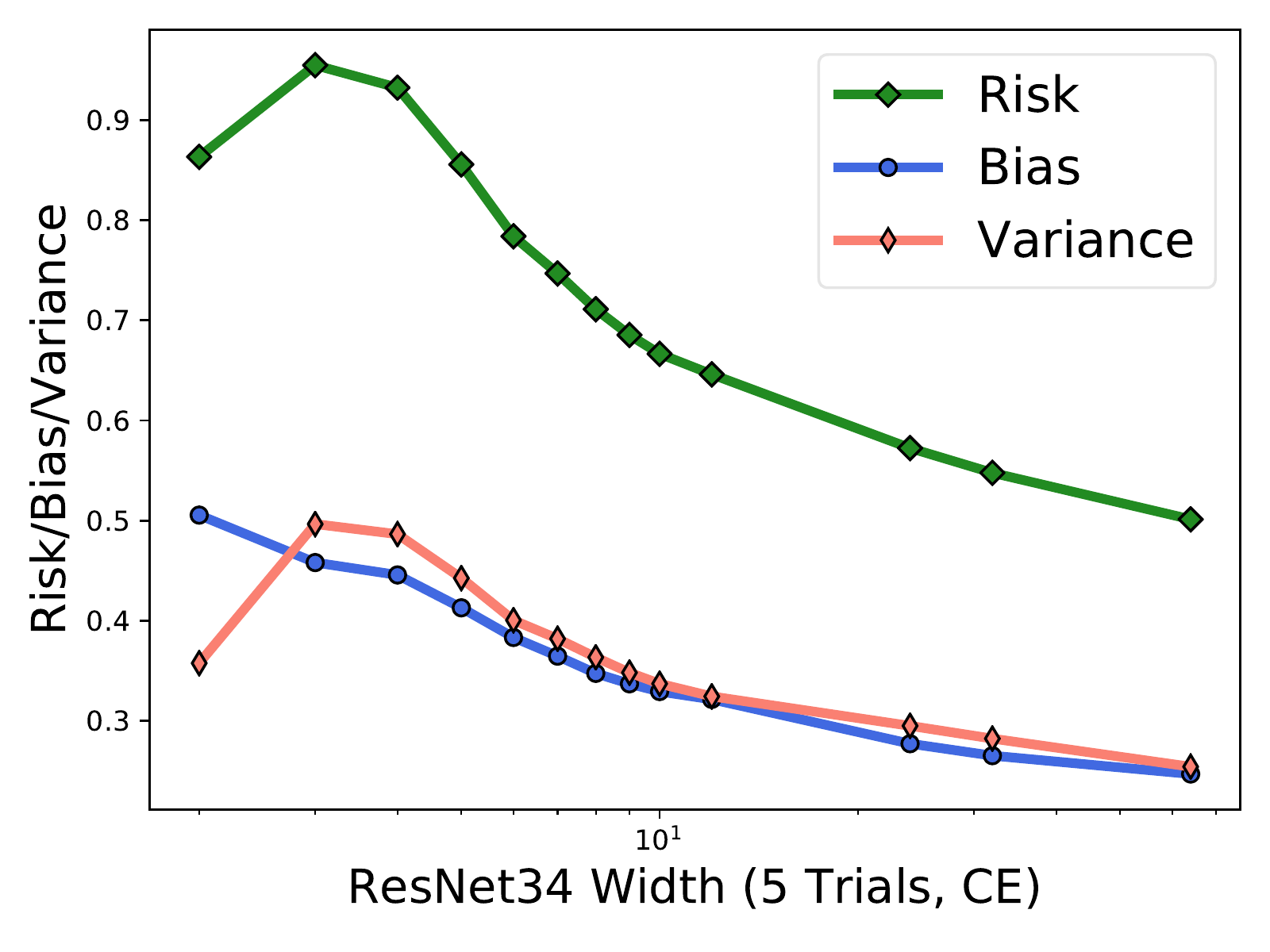}}
    \subfigure[\label{fig:data_mnist}DNN, MSE loss, MNIST]{\includegraphics[width=0.31\textwidth]{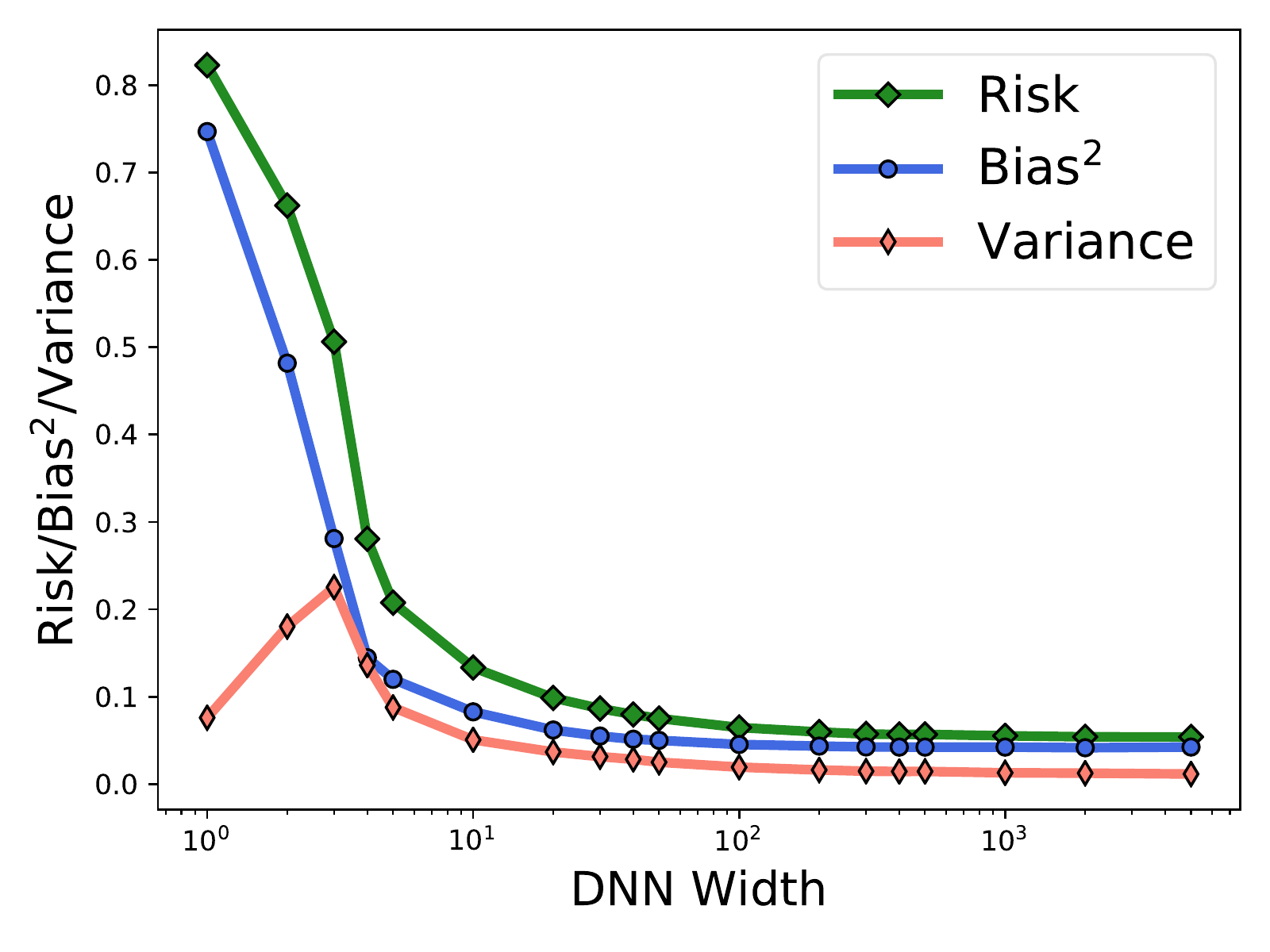}}
      \vskip -0.1in
    \caption{Risk, bias, and variance with respect to different network architectures, training loss functions, and datasets. (a). ResNext29 trained by MSE loss on CIFAR10 dataset (25,000 training samples). (b). ResNet34 trained by CE loss (estimated by generalized bias-variance decomposition using Bregman divergence) on CIFAR10 dataset (10,000 training samples). (c). Fully connected network with one hidden layer and ReLU activation trained by MSE loss on MNIST dataset (10,000 training samples).}
    \label{fig:robust-arch-loss-data}
  \end{center}
  \vskip -0.2in
\end{figure*}

In this section, we study the bias and variance (equations~\eqref{eq:bias} and \eqref{eq:variance}) of deep neural networks. While the bias is monotonically decreasing as folk wisdom would predict, the variance is unimodal (first increases to a peak and then decreases). We conduct extensive experiments to verify that this phenomenon appears robustly across architectures, datasets, optimizer, and loss function. Our code can be found at \url{https://github.com/yaodongyu/Rethink-BiasVariance-Tradeoff}.

\subsection{Mainline Experimental Setup}\label{section:mainlinesetup}
\begin{figure*}[ht]
  \begin{center}
    \subfigure{\includegraphics[width=.48\textwidth]{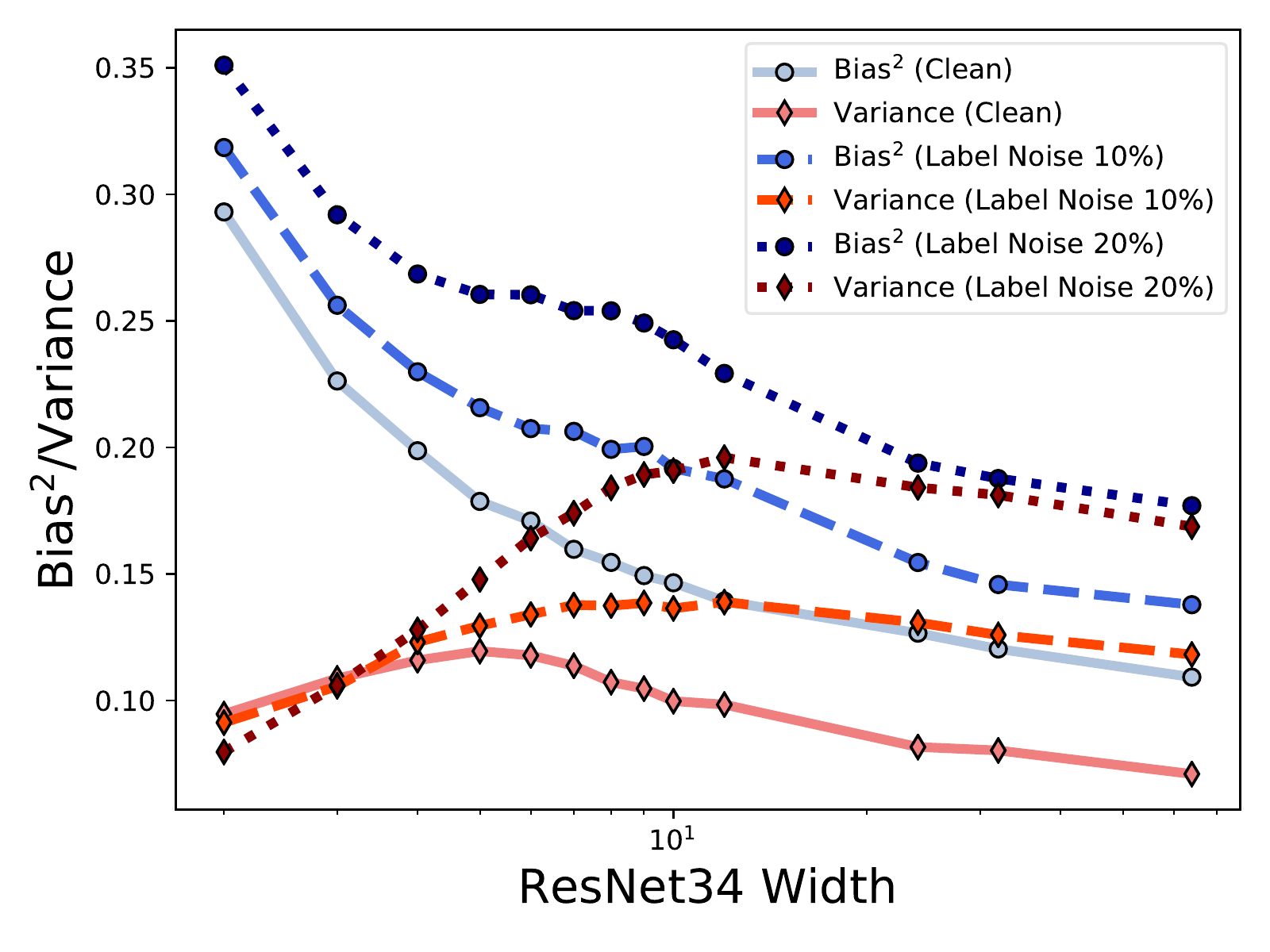}
    }
    \subfigure{
    \includegraphics[width=.48\textwidth]{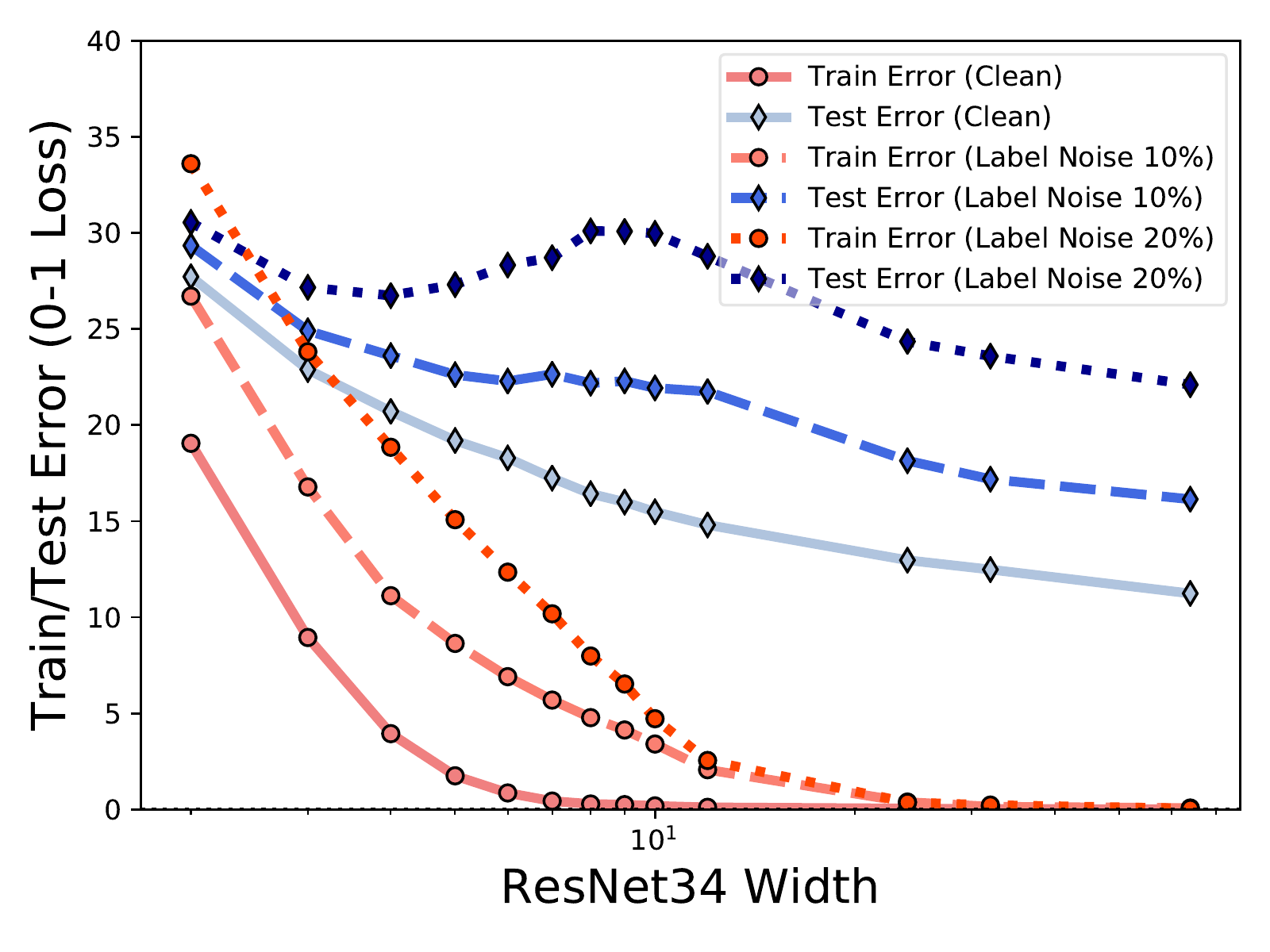}
    }
    \vskip -0.2in
    \caption{Increasing label noise leads to double-descent. 
(\textbf{Left}) Bias and variance under different label noise percentage. (\textbf{Right}) Training error and test error under different label noise percentage. }
    \label{fig:noise}
  \end{center}
  \vskip -0.2in
\end{figure*}

We first describe our mainline experimental setup. In the next subsection, we vary each design choice to check robustness of the phenomenon. More extensive experimental results are given in the appendix.

For the mainline experiment, we trained a ResNet34~\cite{he2016deep} on the CIFAR10 dataset~\cite{krizhevsky2009learning}. 
We trained using stochastic gradient descent (SGD) with momentum $0.9$. The initial learning rate is 0.1. We applied stage-wise training (decay learning rate by a factor of 10 every 200 epochs), and used weight decay $5 \times 10^{-4}$. 
To change the model complexity of the neural network, we scale the number of filters (i.e., width) of the convolutional layers. 
More specifically, with $\text{width}=w$, the number of filters are $[w, 2w, 4w, 8w]$.
We vary $w$ from 2 to 64 (the width $w$ of a regular ResNet34 designed for CIFAR10 in \citet{he2016deep} is 16).

Relative to the standard experimental setup~\cite{he2016deep}, there are two main differences. 
First, since bias-variance is usually defined for the squared loss (see \eqref{eq:bias} and \eqref{eq:variance}), our loss function is the squared error (squared $\ell_2$ distance between the softmax probabilities and the one-hot class vector) rather than the log-loss. 
In the next section we also consider models trained with the log-loss and estimate the bias and variance by using a generalized bias-variance decomposition, as described in \textsection\ref{section:prelim_estimating}.
Second, to measure the variance (and hence bias), we need two models trained on independent subsets of the data as discussed in \textsection\ref{section:prelim_estimating}. 
Therefore, the training dataset is split in half and each model is trained on only $n = 25,000 = 50,000/2$ data points. 
We estimate the variance by averaging over $N = 3$ such random splits (i.e., we train $6 = 3 \times 2$ copies of each model).

In Figure \ref{fig:mainline}, we can see that the variance as a function of the width is unimodal and the bias is monotonically decreasing. Since the scale of the variance is small relative to the bias, the overall behavior of the risk is {monotonically decreasing}.

\subsection{Varying Architectures, Loss Functions, Datasets} \label{section:phenomenonrobust}

\textbf{Architectures.} We observe the same monotonically descreasing bias and unimodal variance phenomenon for ResNext29~\cite{xie2017aggregated}. To scale the ``width'' of the ResNext29, we first set the number of channels to 1 and increase the \textit{cardinality}, defined in \cite{xie2017aggregated}, from 2 to 4, and then fix the {cardinality} at 4 and increase channel size from 1 to 32. 
Results are shown in Figure~\ref{fig:arch}, where the width on the $x$-axis is defined as the {cardinality} times the filter size.

\textbf{Loss Function.} 
In addition to the bias-variance decomposition for MSE loss, we also considered a similar decomposition for cross-entropy loss as described in \textsection\ref{section:prelim_estimating}. We train  with cross-entropy loss and use $n = 10,000$ training samples (5 splits), repeating 
$N = 4$ times with independent random splits. As shown in Figure \ref{fig:loss}, the behavior of the generalized bias and variance for cross entropy is consistent with our earlier observations: the bias is monotonically decreasing and the variance is unimodal. The risk first increases and then decreases, corresponding to the unimodal risk pattern in Figure~\ref{fig:Svariance}. 

\textbf{Datasets.} In addition to CIFAR10, we study bias and variance on MNIST~\cite{lecun1998mnist} and Fashion-MNIST~\cite{xiao2017fashion}. 
For these two datasets, we use a fully connected neural network with one hidden layer with ReLU activation function. 
The ``width'' of the network is the number of hidden nodes. {We use 10,000 training samples ($N=5$).}
As seen in Figure \ref{fig:data_mnist} and \ref{fig:data_fmnist} (in Appendix~\ref{sec:appendix-exp}), for both MNIST and Fashion-MNIST, the variance is again unimodal and the bias is monotonically decreasing.

In addition to the above experiments, we also conduct experiments on the CIFAR100 dataset, the VGG network architecture~\cite{simonyan2014very}, various training sample sizes, and different weight decay regularization and present the results in Appendix~\ref{sec:appendix-exp}. We observe the same {monotonically descreasing bias} and {unimodal variance} phenomenon in {\em all} of these experiments.

\subsection{Connection to Double-Descent Risk}\label{section:doubledescent}
When the relative scale of bias and variance changes, the risk displays one of the three patterns, \textit{monotonically decreasing}, \textit{double descent}, and \textit{unimodal}, as presented in Figure \ref{fig:Mvariance}, \ref{fig:Lvariance} and \ref{fig:Svariance}. 
In particular, the recent stream of observations on double descent risk \cite{belkindouble} can be explained by unimodal variance and monotonically decreasing bias. 
In our experiments, including the experiments in previous sections, we typically observe monotonically decreasing risk; but with more label noise, the variance will increase and we observe the double descent risk curve.

\paragraph{Label Noise.}
Similar to the setup in \citet{nakkiran2019more}, for each split, we sample training data from the whole training dataset, and replace the label of each training example with a uniform random class with independent probability $p$. 
Label noise increases the variance of the model and hence leads to double-descent risk as seen in Figure~\ref{fig:noise}. 
If the variance is small, the risk does not have the double-descent shape because the variance peak is not large enough to overwhelm the bias, as observed in Figures \ref{fig:mainline}, \ref{fig:arch}, \ref{fig:data_mnist} and \ref{fig:data_fmnist}. 
\begin{figure*}[ht]
  \begin{center}
    \subfigure[\label{fig:ood}OOD Example]{\includegraphics[width=0.31\textwidth]{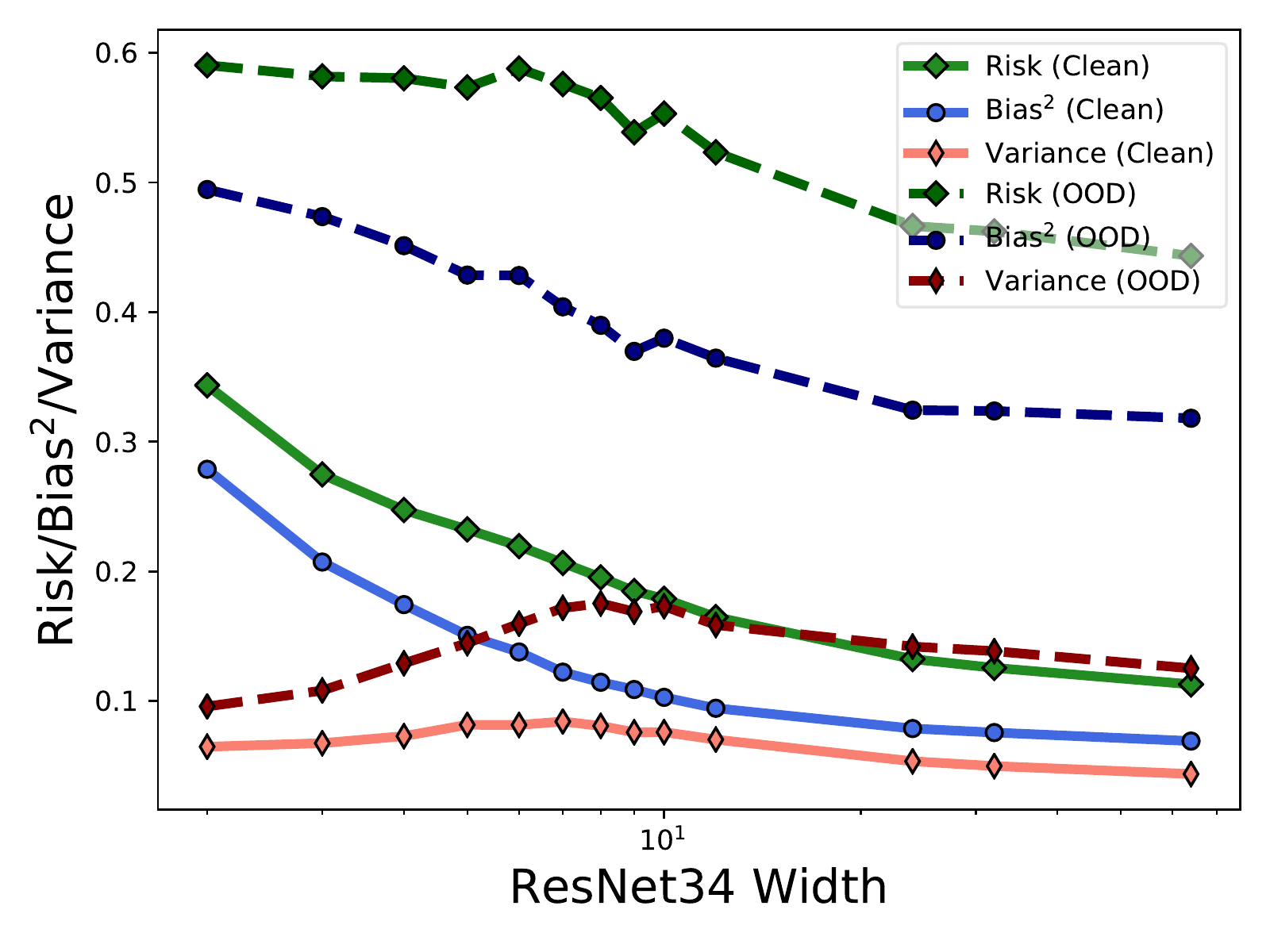}}
    \subfigure[\label{fig:depth-b}Bias of model with different depth]{\includegraphics[width=0.31\textwidth]{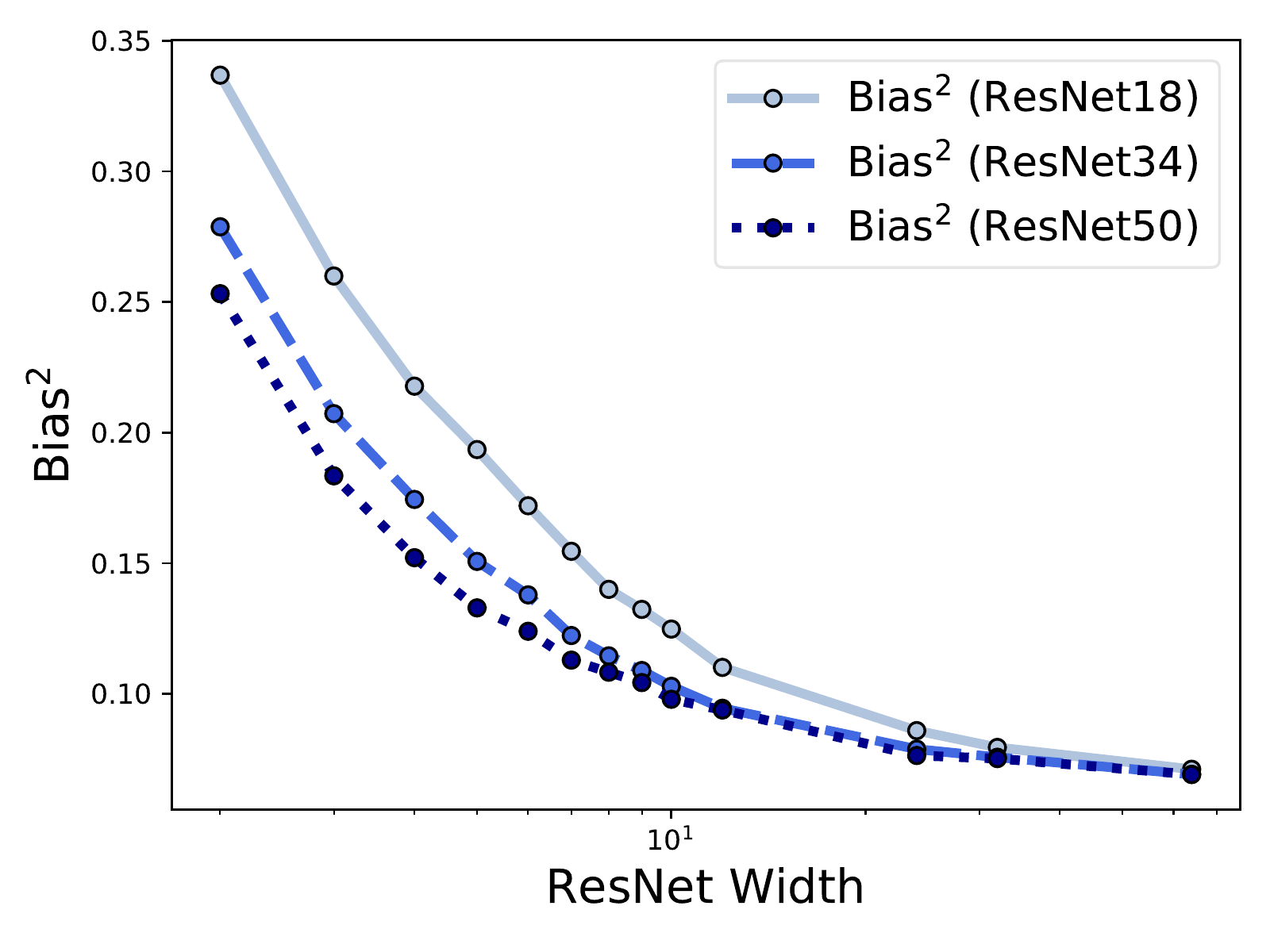}}
    \subfigure[\label{fig:depth-v}Variance of model with different depth]{\includegraphics[width=0.31\textwidth]{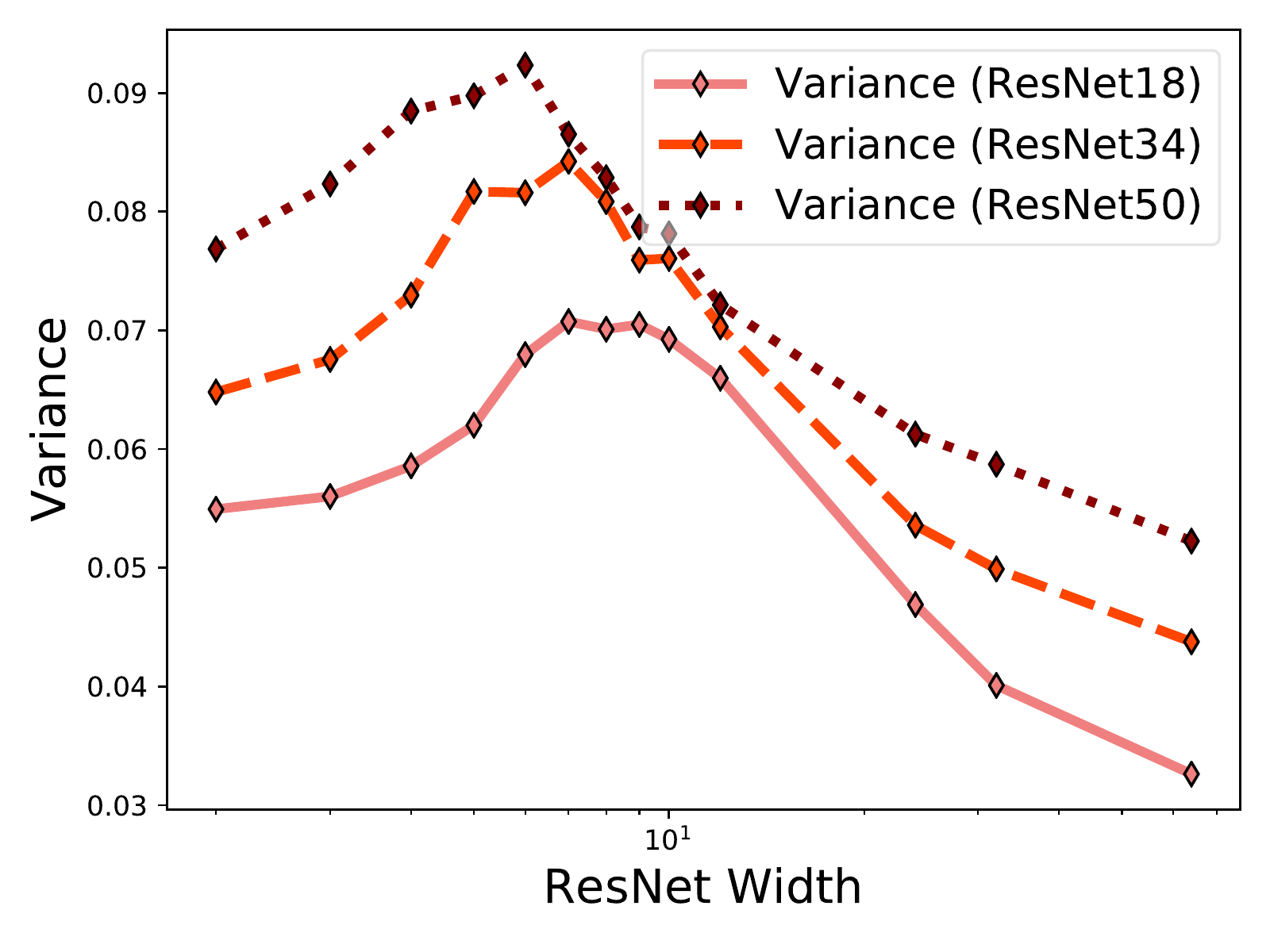}}
      \vskip -0.1in
    \caption{(a). Risk, bias, and variance for ResNet34 on out-of-distribution examples (CIFAR10-C dataset). (b)-(c). Bias and variance for ResNet with different depth  trained by MSE loss on CIFAR10 (25,000 training samples).}
    \label{fig:ood-plus-depth}
  \end{center}
  \vskip -0.2in
\end{figure*}

\subsection{Discussion of Possible Sources of Error}\label{section:error}
In this section, we briefly describe the possible sources of error in our estimator defined in  \textsection\ref{section:prelim_estimating}. 

\textbf{Mean Squared Error.} As argued in  \textsection\ref{section:prelim_estimating}, the variance estimator  is unbiased. 
To understand the variance of the estimator, we first split the data into two parts. For each part, we compute the bias and variance for varying network width by using our estimator. Averaging across different model width, the relative difference between the two parts is 0.6\% for bias and 3\% for variance, so our results for MSE are minimally sensitive to finite-sample effects. The complete experiments can be found in the appendix~(see Figure \ref{fig:mse_error}).

\textbf{Cross Entropy Loss.} 
For cross entropy loss, we are currently unable to obtain an unbiased estimator. We can assess the quality of our estimator using the following scheme. We partition the dataset into five parts $\mc T_1, \dots, \mc T_5$, i.e., set $N=5$ in Algorithm~\ref{alg:example}. Then, we sequentially plot the estimate of bias and variance using $k=1, 2, 3, 4$ as described in Algorithm~\ref{alg:example}. We observe that using larger $k$ gives better estimates. 
In Figure \ref{fig:ce_error} of Appendix \ref{appendix:ce_error}, we observe that as $k$ increases, the bias curve systematically decreases and the variance curve increases.
Therefore our estimator over-estimates the bias and under-estimates the variance, but the overall behaviors of the curves remain consistent.

\section{What Affects the Bias and Variance?}\label{section:what-affect-bv}
In this section, through the Bias-Variance decomposition analyzed in \textsection\ref{section:bvfornn}, we investigate the role of depth for neural networks and the robustness of neural networks on out-of-distribution examples.

\subsection{Bias-Variance Tradeoff for Out-of-Distribution (OOD) Example}
For many real-world computer vision applications, inputs can be corrupted by random noise, blur, weather, etc. 
These common occurring corruptions are shown to significantly decrease model performance~\cite{azulay2019deep, hendrycks2019benchmarking}. 
To better understand the ``generalization gap'' between in-distribution test examples and out-of-distribution test examples, we empirically evaluate the bias and variance on the CIFAR10-C dataset developed by~\citet{hendrycks2019benchmarking}, which is a common corruption benchmark and includes 15 types of corruption.

By applying the models trained in the mainline experiment, we are able to evaluate the bias and variance on CIFAR10-C test dataset according to the definitions in \eqref{eq:bias} and \eqref{eq:variance}. 
As we can see from Figure \ref{fig:ood}, both the bias and variance increase relative to the original CIFAR10 test set.
Consistent with the phenomenon observed in the mainline experiment, the bias dominates the overall risk. 
The results indicate that the  ``generalization gap'' mainly comes from increased bias, with relatively less contribution from variance as well.

\subsection{Effect of Model Depth on Bias and Variance}\label{section:depth}
{In addition to the ResNet34 considered in the mainline experiment, we also evaluate the bias and variance for ResNet18 and ResNet50.
Same as the mainline experiment setup, we estimate the bias and variance for ResNet using 25,000 training samples ($N=2$) and three independent random splits ($k=3$). The standard building block of ResNet50 architecture in \citet{he2016deep} is bottleneck block, which is different from the basic block used in ResNet18 and ResNet34.
To ensure that depth is the only changing variable across three architectures, we apply the basic block for ResNet50. 
Same training epochs and learning rate decays are applied to three models.} 

From Figure~\ref{fig:depth-b} and \ref{fig:depth-v}, we observe that {\em the bias decreases as the depth increases,  while the variance increases as the depth increases}. For each model, the bias is monotonically decreasing and the  variance is unimodal. 
The differences in variance are small (around 0.01) compared with the changes in bias. Overall, the risk typically decreases as the depth increases. 
Our experimental results suggest that the improved generalization for deeper models, with the same network architecture, are mainly attributed to lower bias. 

For completeness, we also include the bias and variance versus depth when basic blocks in ResNet are replaced by bottleneck blocks~(see Figure \ref{fig:depth-bottle} in the appendix). We observe similar qualitative trend of bias and variance.

Note that at high width, the bias of ResNet50 is slightly higher than the bias of ResNet18 and ResNet34. We attribute this inconsistency to difficulties when training ResNet50 without bottleneck blocks at high width.
Lastly, we also include the bias and variance versus depth for out-of-distribution test samples, in which case we also observed decreased bias and increased variance as depth increases, as shown in Figure \ref{fig:ood_depth} of Appendix \ref{appendix:ood_depth}.

\section{Theoretical Insights from a Two-layer Linear Model} \label{section:two-layer}
\begin{figure*}[ht]
  \begin{center}
    \subfigure[\label{fig:risk_synthetic}Risk v.s. $\gamma$ for different $n$]{\includegraphics[width=.32\textwidth]{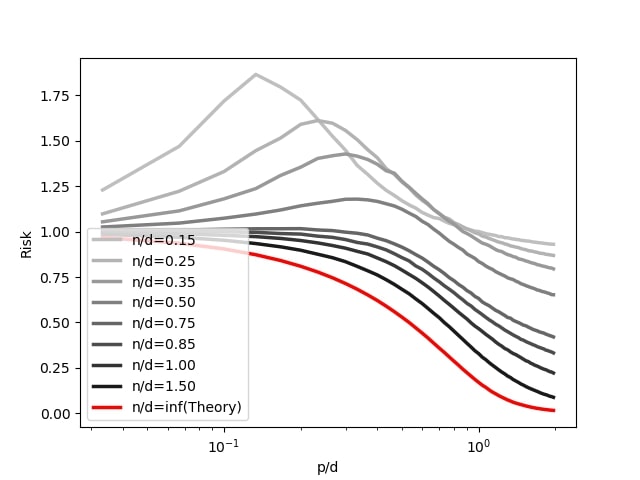}
    }
    \subfigure[\label{fig:bias_synthetic}Bias v.s. $\gamma$ for different $n$]{
    \includegraphics[width=.32\textwidth]{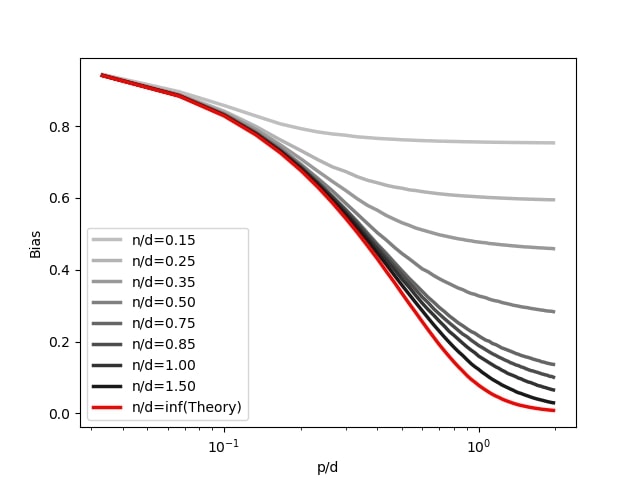}
    }
    \subfigure[\label{fig:variance_synthetic}Variance v.s. $\gamma$ for different $n$]{
    \includegraphics[width=.32\textwidth]{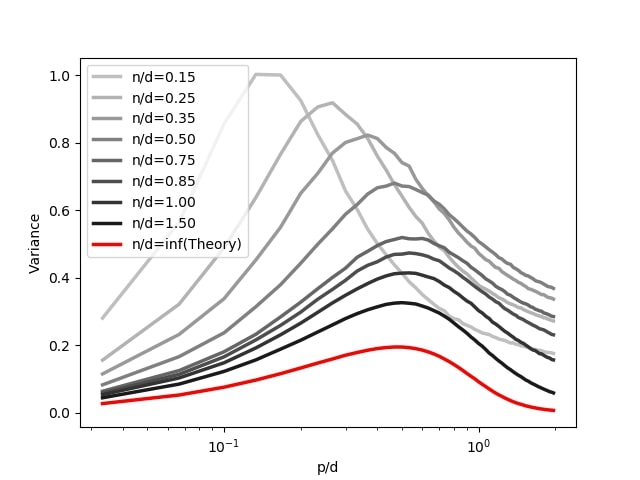}
    }
    \vskip -0.1in
    \caption{Risk, bias, and variance for a two-layer linear neural network.}
    \label{fig:synthetic}
  \end{center}
  \vskip -0.2in
\end{figure*} 

While the preceding experiments show that the bias and variance robustly exhibit monotonic-unimodal behavior in the \emph{random-design} setting, existing theoretical analyses hold instead for the \emph{fixed-design} setting, where the behavior of the bias and variance are more complex, with both the bias and variance exhibiting a peak and the risk exhibiting double descent pattern (\citet[Figure 6]{meisong}). 
In general, while the risk should be the same (in expectation) for the random and fixed design setting, the fixed-design setting has lower bias and higher variance. 

Motivated by the more natural behavior in the random-design setting, we work to extend the existing fixed-design theory to the random-design case. 
Our starting point is \citet{meisong}, who consider two-layer non-linear networks with random hidden layer weights. However, the randomness in the design complicates the analysis, so we make two points of departure to help simplify: 
first, we consider two-layer \emph{linear} rather than non-linear networks, and second, we consider a different scaling limit ($n/d \to \infty$ rather than $n/d$ going to some constant). 
In this setting, we rigorously show that the variance is indeed unimodal and the bias is monotonically decreasing (Figure \ref{fig:synthetic}). 
Our precise assumptions are given below.

\subsection{Model Assumptions}\label{section:model}

We consider the task of learning a function $y = f(\mb x)$ that maps each input vector $\mb x \in \R^d$ to an output (label) value $y \in \R$. 
The input-output pair $(\mb x, y)$ is assumed to be drawn from a distribution where $\mb x \sim \mc{N}(0, \mb I_d/d)$ and 
\begin{equation}\label{eq:f0}
  y = f_0(\mb x) := \mb x^\transpose \mb \theta,
\end{equation}
where $\mb \theta \in \R^d$ is a weight vector. 
Given a training set $\mc T:=\{(\mb x_i, y_i)\}_{i=1}^{n}$ with training samples drawn independently from the data distribution, we learn a two-layer linear neural network parametrized by $\mb W \in \R^{p \times d}$ and $\mb \beta \in \R^p$ as
\begin{equation*}
  f(\mb x) = (\mb W \mb x)^\transpose \mb \beta,
\end{equation*}
where $p$ is the number of hidden units in the network.
In above, we take $\mb W$ as a parameter independent of the training data $\mc T$ whose entries are drawn from i.i.d. Gaussian distribution $\mc{N}(0, 1/d)$. 
Given $\mb W$, the parameter $\mb \beta$ is estimated by solving the following {\em ridge regression}\footnote{$\ell_2$ regularization on weight parameters is arguably the most widely used technique in training neural network, known for improving generalization \cite{krogh1992simple}. Other regularization such as $\ell_1$ can also be used and leads to qualitatively similar behaviors.} problem
\begin{equation}\label{eq:beta_hat}
  \mb\beta_\lambda(\mc T, \mb W) = \argmin_{\mb\beta\in\R^p} \|(\mb W\mb X)^\transpose \mb\beta-\mb y\|_2^2 + \lambda\|\mb \beta\|_2^2,
\end{equation}
where $\mb X = [\mb x_1, \ldots, \mb x_n]\in\R^{d\times n}$ denotes a matrix that contains training data vectors as its columns, $\mb y = [y_1, \ldots, y_n] \in \R^n$ denotes a vector containing training labels as its entries, and $\lambda\in\R^+$ is the regularization parameter. 
By noting that the solution to \eqref{eq:beta_hat} is given by
\begin{equation*}\label{eq:betahat_explicit}
  \mb\beta_\lambda(\mc T, \mb W) = (\mb W\mb X\mb X^\transpose \mb W^\transpose+\lambda \mb I)^{-1}\mb W\mb X\mb y,
\end{equation*}
our estimator $f:\R^d \rightarrow \R$ is given as 
\begin{equation}\label{eq:exprf}
  f_\lambda(\mb x; \mc T, \mb W) = \mb x^\transpose \mb W^\transpose \mb\beta_\lambda(\mc T, \mb W).
\end{equation}

\subsection{Bias-Variance Analysis}
We may now calculate the bias and variance of the model described above via the following formulations:
\begin{align*}
  \textbf{Bias}_\lambda(\mb\theta)^2 &= \E_{\mb x} \left[\E_{\mc T, \mb W} f_\lambda(\mb x; \mc T, \mb W)- f_0(\mb x)\right]^2,\\
  \textbf{Variance}_\lambda(\mb\theta) &= \E_{\mb x}\text{Var}_{\mc T,\mb W} \left[f_\lambda(\mb x; \mc T, \mb W)\right],
\end{align*}
where $f_0(\mb x)$ and $f_\lambda(\mb x; \mc T, \mb W)$ are defined in \eqref{eq:f0} and \eqref{eq:exprf}, respectively. 
Note that the bias and variance are functions of the model parameter $\mb\theta$. To simplify the analysis, we introduce a prior $\mb\theta\sim\mc{N}(0, \mb I_d)$ and calculate the expected bias and expected variance as
\begin{align}
  \textbf{Bias}_\lambda^2 &:= \E_{\mb\theta} \textbf{Bias}_\lambda(\mb\theta)^2\label{eq:Ebias},  \\
  \textbf{Variance}_\lambda &:= \E_{\mb\theta} \textbf{Variance}_\lambda(\mb\theta). \label{eq:Evariance}
\end{align}
The precise formulas for the expected bias and the expected variance are parametrized by the dimension of the input feature $d$, the number of training points $n$, the number of hidden units $p$ and also $\lambda$. 

Previous literatures~\cite{meisong} suggests that both the risk and the variance achieves a peak at the interpolation threshold ($n=p$). 
In the regime when $n$ is very large, the risk no longer exhibits a peak, but the unimodal pattern of variance still holds. 
In the rest of the section, we consider the regime where the $n$ is large (monotonically decreasing risk), and derive the precise expression for the bias and variance of the model. From our expression, we obtain the location where the variance achieves the peak.
For this purpose, we consider the following asymptotic regime of $n, p$ and $d$:
\begin{assumption}\label{assumption:largesample}
Let $\{(d, n(d), p(d))\}_{d=1}^\infty$ be a given sequence of triples. We assume that there exists a $\gamma > 0$ such that 
\begin{equation*}
  \lim_{d \to \infty} \frac{p(d)}{d} = \gamma, \quad \text{and} \quad \lim_{d\to\infty} \frac{n(d)}{d} = \infty.
\end{equation*}
For simplicity, we will write $n := n(d)$ and $p := p(d)$.  
\end{assumption}
With the assumption above, we have the expression of the expected bias, variance and risk as a function of $\gamma$ and $\lambda$.
\begin{theorem}\label{thm:largesample}
Given $\{(d, n(d), p(d))\}_{d=1}^\infty$ that satisfies Assumption \ref{assumption:largesample}, let $\lambda=\frac{n}{d}\lambda_0$ for some fixed $\lambda_0 > 0$.
The asymptotic expression of expected bias and variance are given by 
\begin{align}
  &\lim_{d \to \infty}\textbf{Bias}_{\lambda}^2 = \frac{1}{4}\Phi_3(\lambda_0, \gamma)^2,\label{eq:largesample_bias}\\
  &\lim_{d \to \infty}\textbf{Variance}_{\lambda} =\nonumber\\
  &\begin{cases}
    \frac{\Phi_1(\lambda_0, \frac{1}{\gamma})}{2\Phi_2(\lambda_0, \frac{1}{\gamma})} - \frac{(1-\gamma)(1-2\gamma)}{2\gamma} - \frac{1}{4}\Phi_3(\lambda_0, \gamma)^2, & \gamma\le1,\\
    \frac{\Phi_1(\lambda_0, \gamma)}{2\Phi_2(\lambda_0, \gamma)} -
    \frac{\gamma-1}{2} - \frac{1}{4}\Phi_3(\lambda_0, \gamma)^2, &\gamma > 1,\nonumber
  \end{cases}
\end{align} 
where
\begin{equation*}
  \begin{split}
    \Phi_1(\lambda_0, \gamma) &= \lambda_0(\gamma+1) + (\gamma-1)^2,\\
    \Phi_2(\lambda_0, \gamma) &= \sqrt{(\lambda_0+1)^2 + 2(\lambda_0-1)\gamma + \gamma^2}, \\
    \Phi_3(\lambda_0, \gamma) &=\Phi_2(\lambda_0, \gamma) - \lambda_0 - \gamma+1.
  \end{split}
\end{equation*}
\end{theorem}
The proof is given in  Appendix \ref{appendix:proof}. 

The risk can be obtained through $\textbf{Bias}_\lambda^2+\textbf{Variance}_\lambda$.
The expression in Theorem \ref{thm:largesample} is plotted as the red curves in Figure \ref{fig:synthetic}. 
In addition to the case when $n/d\rightarrow\infty$, we also plot the shape of bias, variance and risk when $n/d\rightarrow\{0.15, 0.25, 0.35, \dots, 1.00, 1.50\}$.
We find that the risk of the model grows from unimodal to monotonically decreasing as the number of samples increased (see Figure \ref{fig:risk_synthetic}). 
Moreover, the bias of the model is monotonically decreasing (see Figure \ref{fig:bias_synthetic}) and the variance is unimodal (see Figure \ref{fig:variance_synthetic}).\\
\begin{corollary}[Monotonicity of Bias]
  The derivative of the limiting expected Bias in \eqref{eq:largesample_bias} can be calculated as
  \begin{equation}\label{eq:bias-derivative}
    -\frac{\left(\sqrt{2 (\gamma +1) \lambda _0+(\gamma -1)^2+\lambda _0^2}-\gamma -\lambda _0+1\right)^2}{2 \sqrt{\gamma ^2+2 \gamma  \left(\lambda _0-1\right)+\left(\lambda _0+1\right){}^2}}.
  \end{equation}
\end{corollary}
When $\lambda_0 \geq 0$, the expression in \eqref{eq:bias-derivative} is strictly non-positive, therefore the limiting expected bias is monotonically non-increasing as a function of $\gamma$, as classical theories predicts. 

To gain further insight into the above formulas, we also consider the case when the ridge regularization amount $\lambda_0$ is small. In particular, we consider the first order effect of $\lambda_0$ on the bias and variance term, and compute the value of $\gamma$ where the variance attains the peak. 
\begin{corollary}[Unimodality of Variance -- small $\lambda_0$ limit]\label{cor:smalllbd}
Under the assumptions of Theorem \ref{thm:largesample}, the first order effect of $\lambda_0$ on variance is given by
\begin{equation*}
  \lim_{d \to \infty}\E\textbf{Variance}_{\lambda} =\\
  \begin{cases}
  O\left(\lambda _0^2\right), \quad \gamma > 1,\\
  -(\gamma -1) \gamma -2 \gamma  \lambda _0+O\left(\lambda _0^2\right), \text{o.w.}\\
  \end{cases}
\end{equation*}
and the risk is given by
\begin{equation*}
  \lim_{d \to \infty}\E\textbf{Risk}_{\lambda} = 
  \begin{cases}
    1-\gamma + O\left(\lambda _0^2\right), & \gamma\le1,\\
    O\left(\lambda _0^2\right),& \gamma > 1.
  \end{cases}
\end{equation*}
Moreover, up to first order, the peak in the variance is
\begin{equation*}
  \textbf{Peak} = \frac{1}{2} - \lambda_0 + O\left(\lambda _0^2\right).
\end{equation*}
\end{corollary}
Theorem \ref{cor:smalllbd} suggests that when $\lambda_0$ is sufficiently small, the variance of the model is maximized when $p=d/2$, and the effect of $\lambda_0$ is to shift the peak slightly to $d/2 - \lambda_0d$. 

From a technical perspective, to compute the variance in the random-design setting, we need to compute the element-wise expectation of certain random matrix. For this purpose, we apply the combinatorics of counting non-cross partitions to characterize the asymptotic expectation of products of Wishart matrices.

\section{Conclusion and Discussion}
In this paper we re-examine the classical theory of bias and variance trade-off as the width of a neural network increases. Through extensive experimentation, our main finding is that, while the bias is monotonically decreasing as classical theory would predict, the variance is unimodal. This combination leads to three typical risk curve patterns, all observed in practice.  Theoretical analysis of a two-layer linear network corroborates these experimental observations. 

The seemingly varied and baffling behaviors of modern neural networks are thus in fact consistent, and explainable through classical bias-variance analysis. 
The main unexplained mystery is the unimodality of the variance. We conjecture that as the model complexity approaches and then goes beyond the data dimension, it is regularization in model estimation (the ridge penalty in our theoretical example) 
that helps bring down the variance. Under this account, the decrease in variance for large dimension comes from better conditioning of the empirical covariance, making it better-aligned with the regularizer.

In the future, it would be interesting to see if phenomena characterized by the simple two-layer model can be rigorously generalized to deeper networks with nonlinear activation, probably revealing other interplays between model complexity and  regularization (explicit or implicit). Such a study could also help explain another phenomenon we (and others) have observed: bias decreases with more layers as variance increases. We believe that the (classic) bias-variance analysis remains a powerful and insightful framework for understanding the behaviors of deep networks; properly used, it can guide practitioners to design more generalizable and robust networks in the future.    

\textbf{\large Acknowledgements.}  We would like to thank Emmanuel Cand\'{e}s for first bringing the double-descent phenomenon to our attention, Song Mei for helpful discussion regarding random v.s. fixed design regression, and Nikhil Srivastava  for pointing out to relevant references in random matrix theory. We would also like to thank Preetum Nakkiran, Mihaela Curmei, and Chloe Hsu  for valuable feedback during preparation of this manuscript.
The authors acknowledge support from Tsinghua-Berkeley Shenzhen Institute Research Fund and BAIR.

\bibliography{reference}
\bibliographystyle{icml2020}

\appendix
\newpage
\onecolumn
\section{Summary of Experiments}
We summarize the experiments in Table~\ref{tb1:exp}, each row corresponds to one experiment, some include several independent splits, in this paper. Every experiment is related to one or multiple figures, which is specified in the last column ``Figure''.
\begin{table}[htp]
    \centering
    {\small
        \begin{tabular}{ llcccccll}
    	\toprule
    	Dataset                        & Architecture & Loss &       Optimizer       & Train Size & $\#$Splits($k$) & Label Noise & Figure                                       & Comment                                          \\
    	\midrule
    	\multirow{1}{*}{CIFAR10}       & ResNet34     & MSE  & SGD(wd=\texttt{5e-4}) &   25000    &      3      &  {\xmark}   & \ref{fig:mainline}, \ref{fig:ood-plus-depth} & Mainline                                         \\
    	\midrule
    	\multirow{2}{*}{CIFAR10}       & ResNext29    & MSE  & SGD(wd=\texttt{5e-4}) &   25000    &      3      &  {\xmark}   & \ref{fig:arch}, \ref{fig:arch-resnext}       & \multirow{2}{*}{Architecture}                    \\
    	                               & VGG11        & MSE  & SGD(wd=\texttt{5e-4}) &   10000    &      1      &  {\xmark}   & \ref{fig:arch-vgg}                           &                                                  \\
    	\midrule
    	\multirow{1}{*}{CIFAR10}       & ResNet34     &  CE  & SGD(wd=\texttt{5e-4}) &   10000    &      4      &  {\xmark}   & \ref{fig:loss}, \ref{fig:appendix-loss}      & Loss                                             \\
    	\midrule
    	\multirow{1}{*}{MNIST}         & DNN          & MSE  & SGD(wd=\texttt{5e-4}) &   10000    &      1      &  {\xmark}   & \ref{fig:data_mnist}                         & \multirow{3}{*}{Dataset}                         \\
    	\multirow{1}{*}{Fashion-MNIST} & DNN          & MSE  & SGD(wd=\texttt{5e-4}) &   10000    &      1      &  {\xmark}   & \ref{fig:data_fmnist}                        &                                                  \\
    	\multirow{1}{*}{CIFAR100}      & ResNet34     &  CE  & SGD(wd=\texttt{5e-4}) &   10000    &      1      &  {\xmark}   & \ref{fig:dataset-cifar100}                   &                                                  \\
    	\midrule
    	\multirow{1}{*}{CIFAR10}       & ResNet34     & MSE  & SGD(wd=\texttt{5e-4}) &   10000    &      1      &  10\%/20\%  & \ref{fig:noise}                              & \multirow{1}{*}{Label noise}                     \\
    	\midrule
    	\multirow{2}{*}{CIFAR10}       & ResNet18     & MSE  & SGD(wd=\texttt{5e-4}) &   25000    &      3      &  {\xmark}   & \ref{fig:ood-plus-depth}                     & \multirow{2}{*}{Depth}                           \\
    	                               & ResNet50     & MSE  & SGD(wd=\texttt{5e-4}) &   25000    &      3      &  {\xmark}   & \ref{fig:ood-plus-depth}                     &                                                  \\
    	\midrule
    	\multirow{2}{*}{CIFAR10}       & ResNet34     & MSE  & SGD(wd=\texttt{5e-4}) &   10000    &      1      &  {\xmark}   & \ref{fig:sample-5split}                      & \multirow{2}{*}{Train size}                      \\
    	                               & ResNet34     & MSE  & SGD(wd=\texttt{5e-4}) &    2500    &      1      &  {\xmark}   & \ref{fig:sample-20split}                     &                                                  \\
    	\midrule
    	\multirow{1}{*}{CIFAR10}       & ResNet34     & MSE  & SGD(wd=\texttt{1e-4}) &   10000    &      1      &  {\xmark}   & \ref{fig:weightdecay-1e4}                    & Weight decay                                     \\
    	\midrule
    	\multirow{3}{*}{CIFAR10}       & ResNet26-B   & MSE  & SGD(wd=\texttt{5e-4}) &   25000    &      3      &  {\xmark}   & \ref{fig:depth-bottle}                       & \multirow{3}{6em}{Depth (with bottleneck block)} \\
    	                               & ResNet38-B   & MSE  & SGD(wd=\texttt{5e-4}) &   25000    &      3      &  {\xmark}   & \ref{fig:depth-bottle}                       &                                                  \\
    	                               & ResNet50-B   & MSE  & SGD(wd=\texttt{5e-4}) &   25000    &      3      &  {\xmark}   & \ref{fig:depth-bottle}                       &                                                  \\
    	     	\midrule
    	\multirow{2}{*}{CIFAR10}       & VGG9     & MSE  & SGD(wd=\texttt{5e-4}) &   25000    &      3      &  {\xmark}   & \ref{fig:depth-vgg}                 & \multirow{2}{*}{Depth}                           \\
    	                               & VGG11     & MSE  & SGD(wd=\texttt{5e-4}) &   25000    &      3      &  {\xmark}   & \ref{fig:depth-vgg}                     &                                                  \\
    	\bottomrule
    \end{tabular}
    }
    \vskip -0.05in
    \caption{Summary of Experiments.}
    \label{tb1:exp}
    \vskip -0.1in
\end{table}

\section{Additional Experiments}\label{sec:appendix-exp}
In this section, we provide additional experimental results, some of them are metioned in \textsection\ref{section:bvfornn} and  \textsection\ref{section:what-affect-bv}.

\textbf{Network Architecture:} The implementation of the deep neural networks used in this work is mainly adapted from \url{https://github.com/kuangliu/pytorch-cifar}.

\textbf{Training Details:} For CIFAR10 dataset and CIFAR100 dataset, when training sample size is 25,000, we use 500 epochs for training and decay by a factor of 10 the learning rate every 200 epoch. When training sample size is 10,000/5,000, we use 1000 epochs for training and decay by a factor of 10 the learning rate every 400 epoch.
For MNIST dataset and FMNIST dataset,  we use 200 epochs for training and decay by a factor of 10 the learning rate every 100 epoch. For all the experiments in this paper, we sampled data without replacement to train the models as described in \textsection\ref{section:prelim_estimating}.

\subsection{Architecture}\label{sec:app-arch}
\begin{figure}[h]
  \begin{center}
    \subfigure{
    \includegraphics[width=.31\textwidth]{Figures/robust/arch/resnext_rbv.pdf}
    }
    \subfigure{
    \includegraphics[width=.31\textwidth]{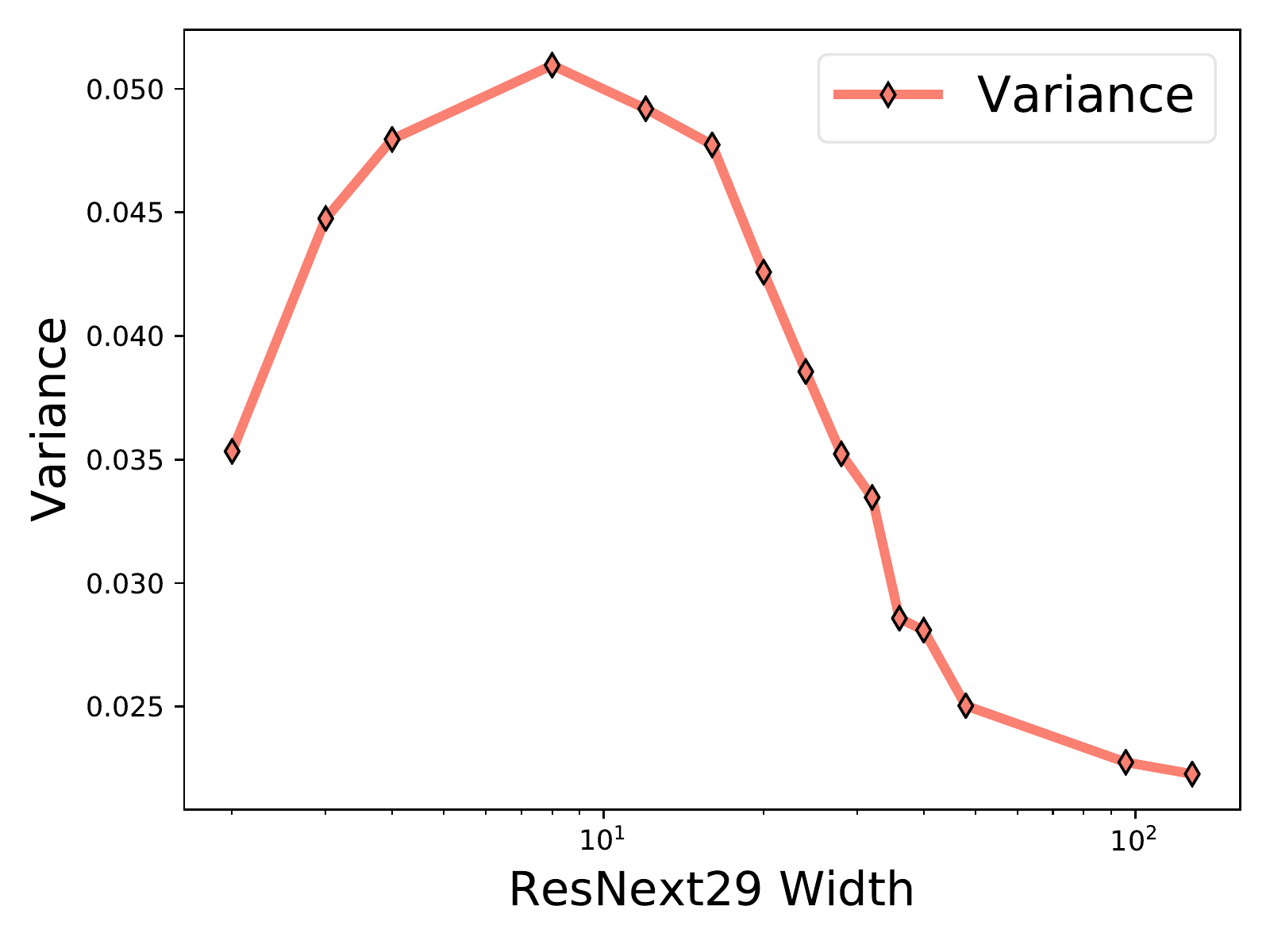}
    }
    \subfigure{
    \includegraphics[width=.32\textwidth]{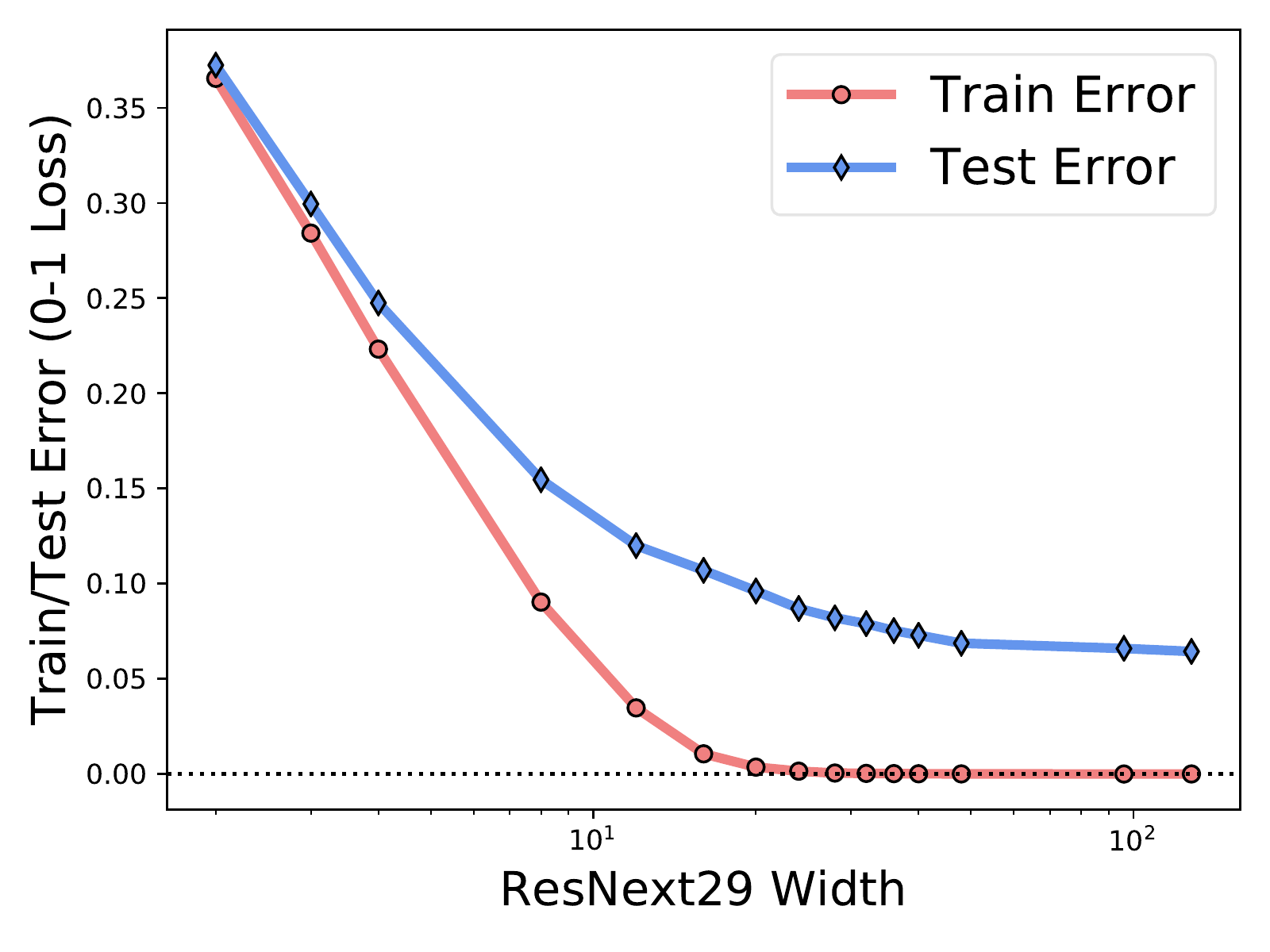}
    }
    \vskip -0.1in
    \caption{Risk, bias, variance, train/test error for ResNext29 trained by MSE loss on CIFAR10 dataset (25,000 training samples). (\textbf{Left}) Risk, bias, and variance for ResNext29. (\textbf{Middle}) Variance for ResNext29. (\textbf{Right}) Train error and test error for ResNext29. }
    \label{fig:arch-resnext}
  \end{center}
  \vskip -0.1in
\end{figure}

\begin{figure}[h]
  \begin{center}
    \subfigure{\includegraphics[width=.32\textwidth]{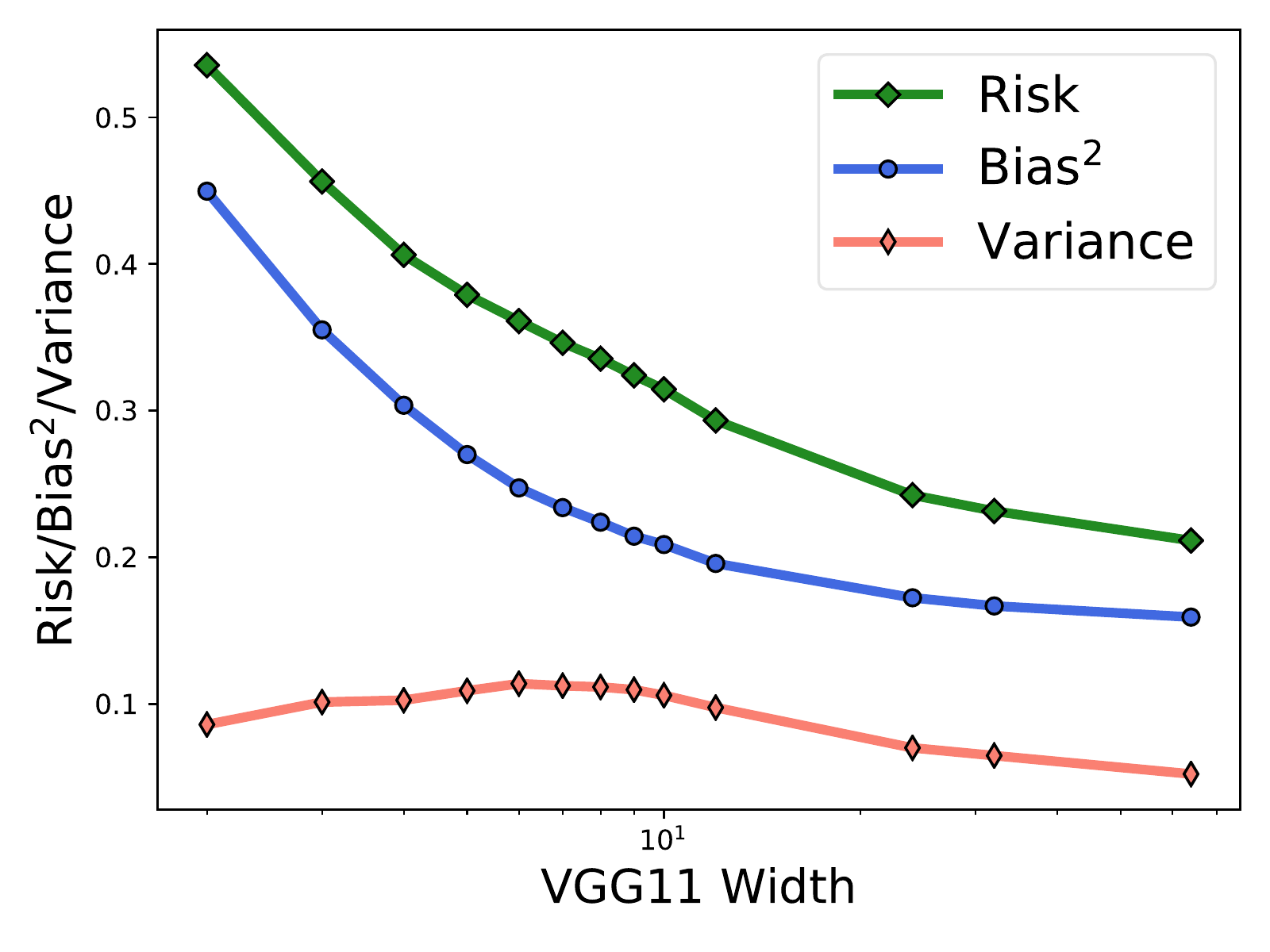}
    }
    \subfigure{
    \includegraphics[width=.32\textwidth]{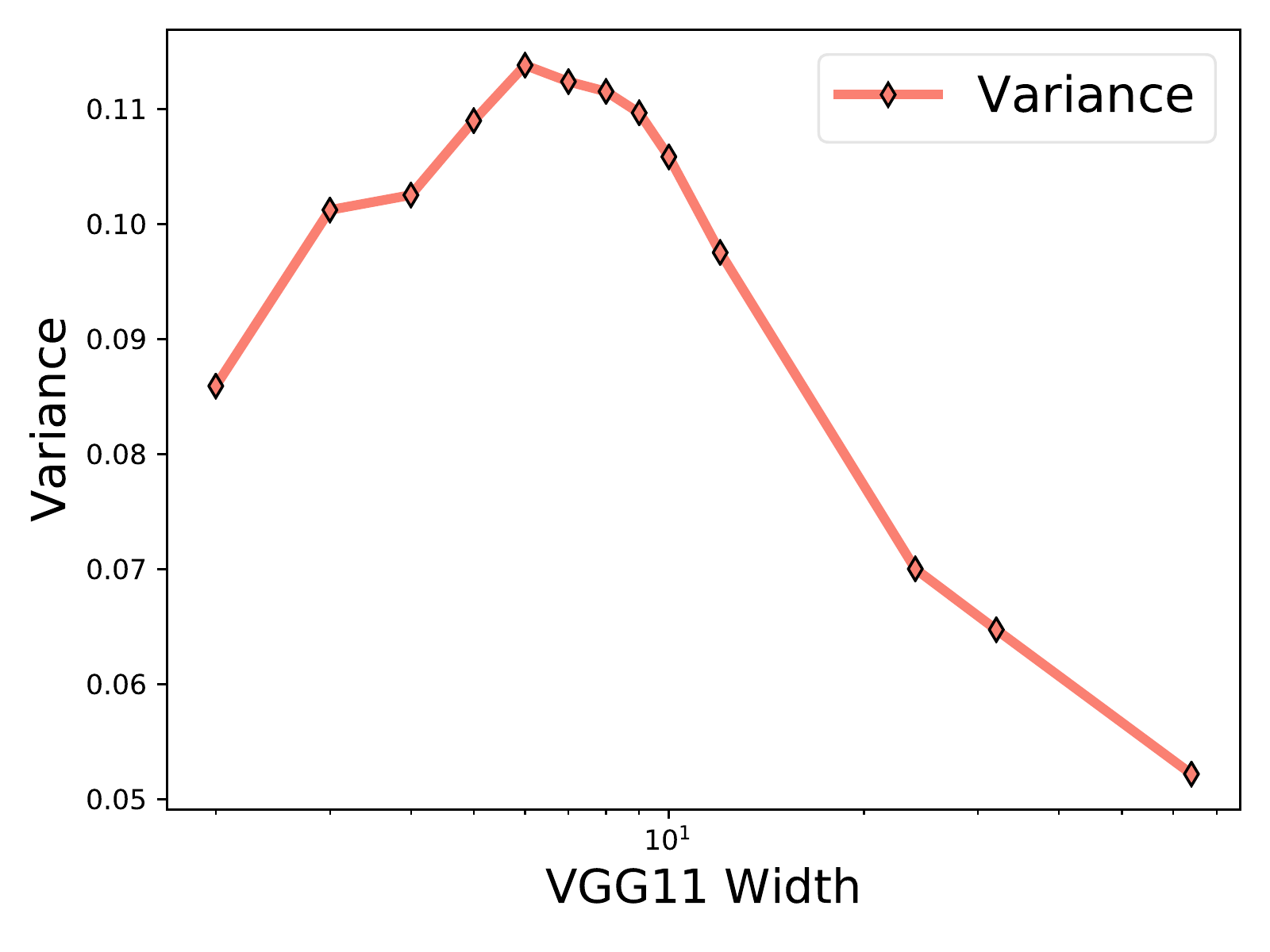}
    }
    \subfigure{
    \includegraphics[width=.32\textwidth]{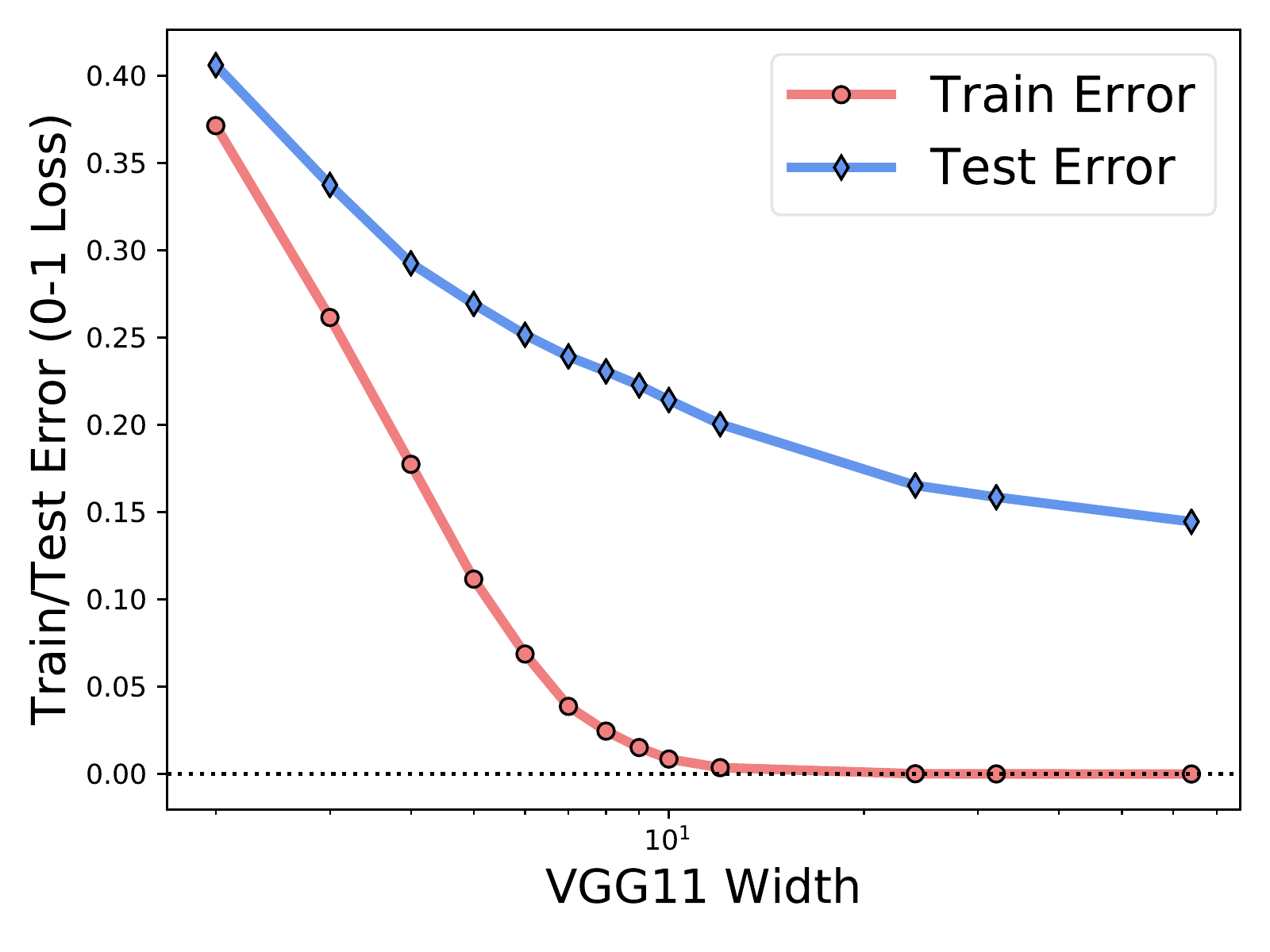}
    }
    \vskip -0.1in
    \caption{Risk, bias, variance, train/test error for VGG11 trained by MSE loss on CIFAR10 dataset (10,000 training samples). (\textbf{Left}) Risk, bias, and variance for VGG11. (\textbf{Middle}) Variance for VGG11. (\textbf{Right}) Train error and test error for VGG11. }
    \label{fig:arch-vgg}
  \end{center}
  \vskip -0.2in
\end{figure}

We provide additional results on ResNext29 presented in  \textsection\ref{section:phenomenonrobust}. The results are shown in Figure~\ref{fig:arch-resnext}. We also study the behavior of risk, bias, and variance of VGG network~\cite{simonyan2014very} on CIFAR10 dataset. Here we use VGG11 and the number of filters are $[k, 2k, 4k, 4k, 8k, 8k, 8k, 8k]$, where $k$ is the width in Figure~\ref{fig:arch-vgg}. The number of training samples of each split is 10,000. We use the same optimization setup as the mainline experiment (ResNet34 in Figure\ref{fig:mainline}).

\subsection{Loss}
\begin{figure}[h]
  \begin{center}
    \subfigure{\includegraphics[width=.32\textwidth]{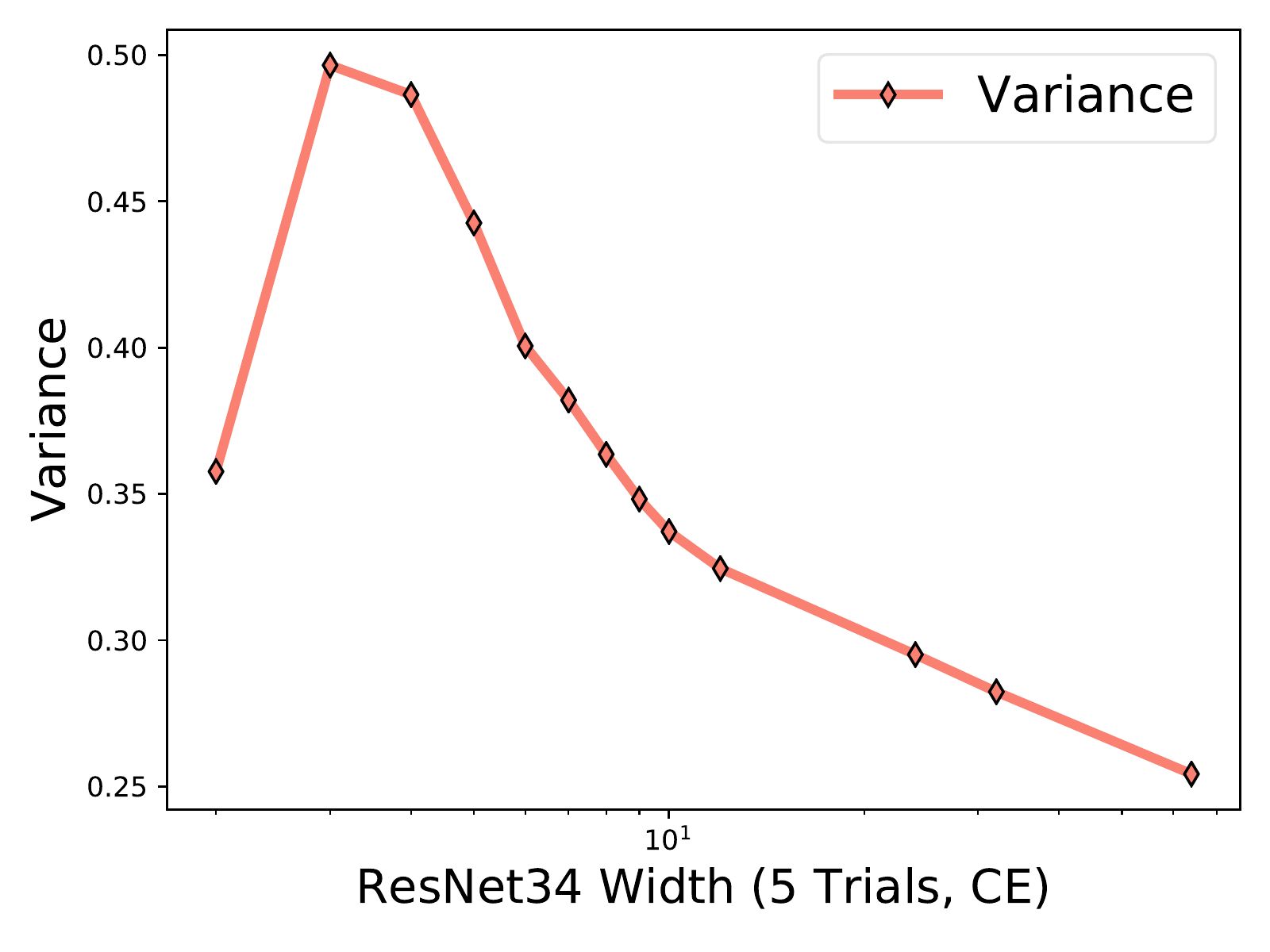}
    }
    \subfigure{
    \includegraphics[width=.32\textwidth]{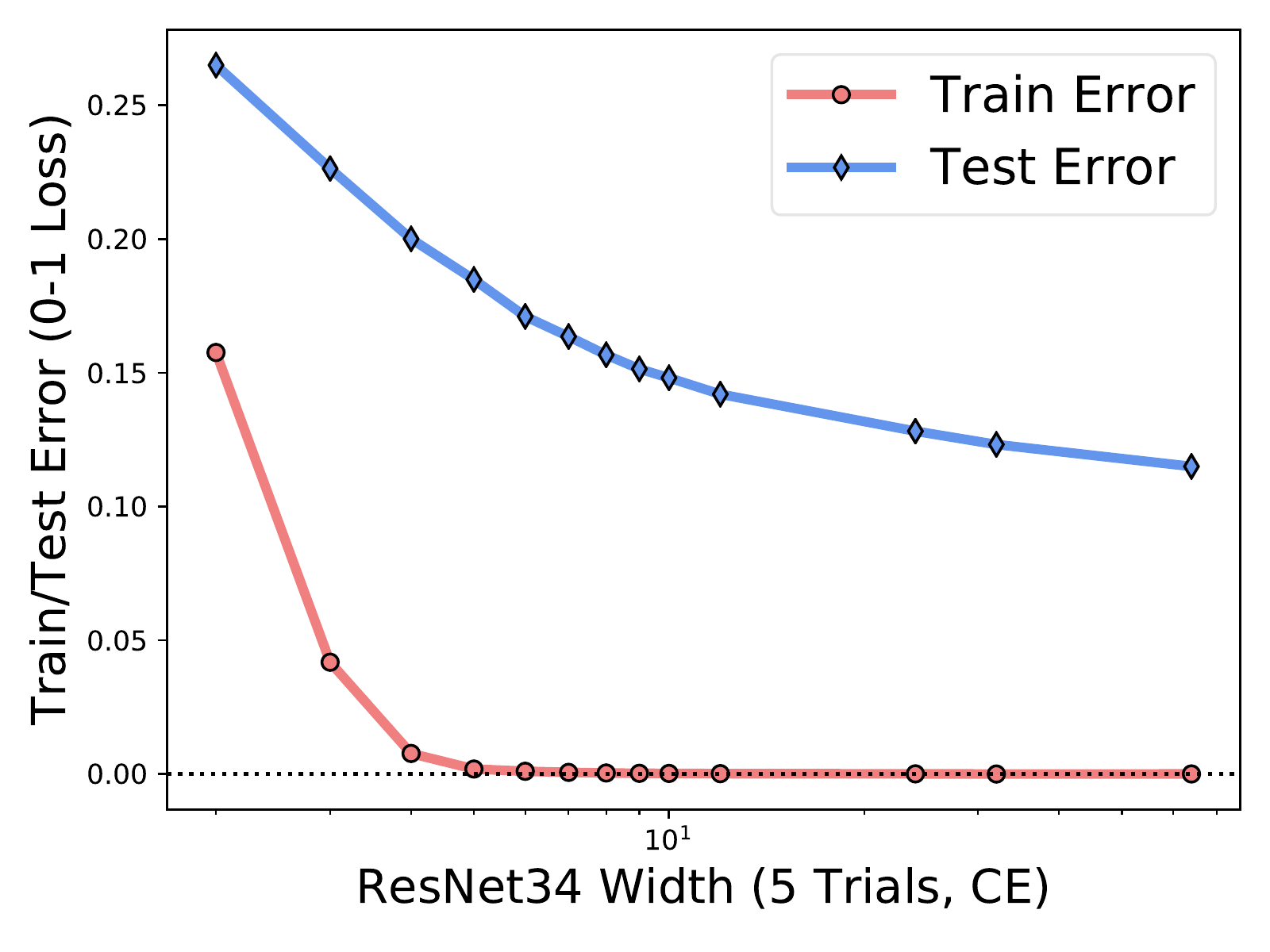}
    }
    \vskip -0.1in
    \caption{Variance and train/test error for ResNet34 trained by cross-entropy loss (estimated by generalized bias-variance decomposition using Bregman divergence) on CIFAR10 dataset (10,000 training samples). (\textbf{Left}) Variance for ResNet34. (\textbf{Right}) Train error and test error for ResNet34.}
    \label{fig:appendix-loss}
  \end{center}
\end{figure}

We provide additional results on cross-entropy loss presented in \textsection\ref{section:phenomenonrobust}, the results are shown in Figure~\ref{fig:appendix-loss}.

\subsection{Dataset}
\begin{figure}[h]
  \begin{center}
    {
    \includegraphics[width=.32\textwidth]{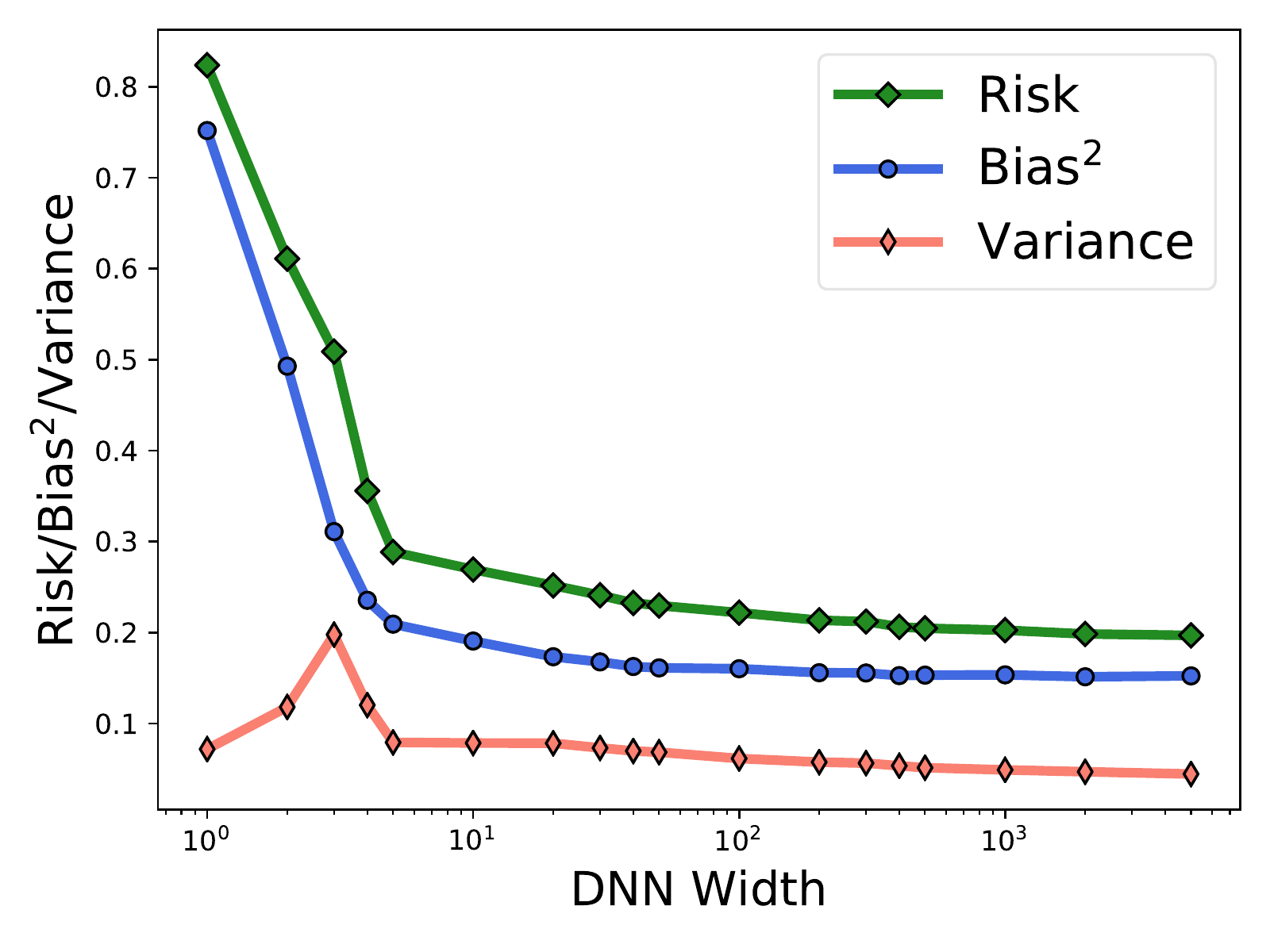}
    }
    \vskip -0.1in
    \caption{Fully connected network with one-hidden-layer and ReLU activation trained by MSE loss on Fashion-MNIST dataset (10,000 training samples).}
    \label{fig:data_fmnist}
  \end{center}
  \vskip -0.1in
\end{figure}

\begin{figure}[h]
  \begin{center}
    \subfigure{\includegraphics[width=.32\textwidth]{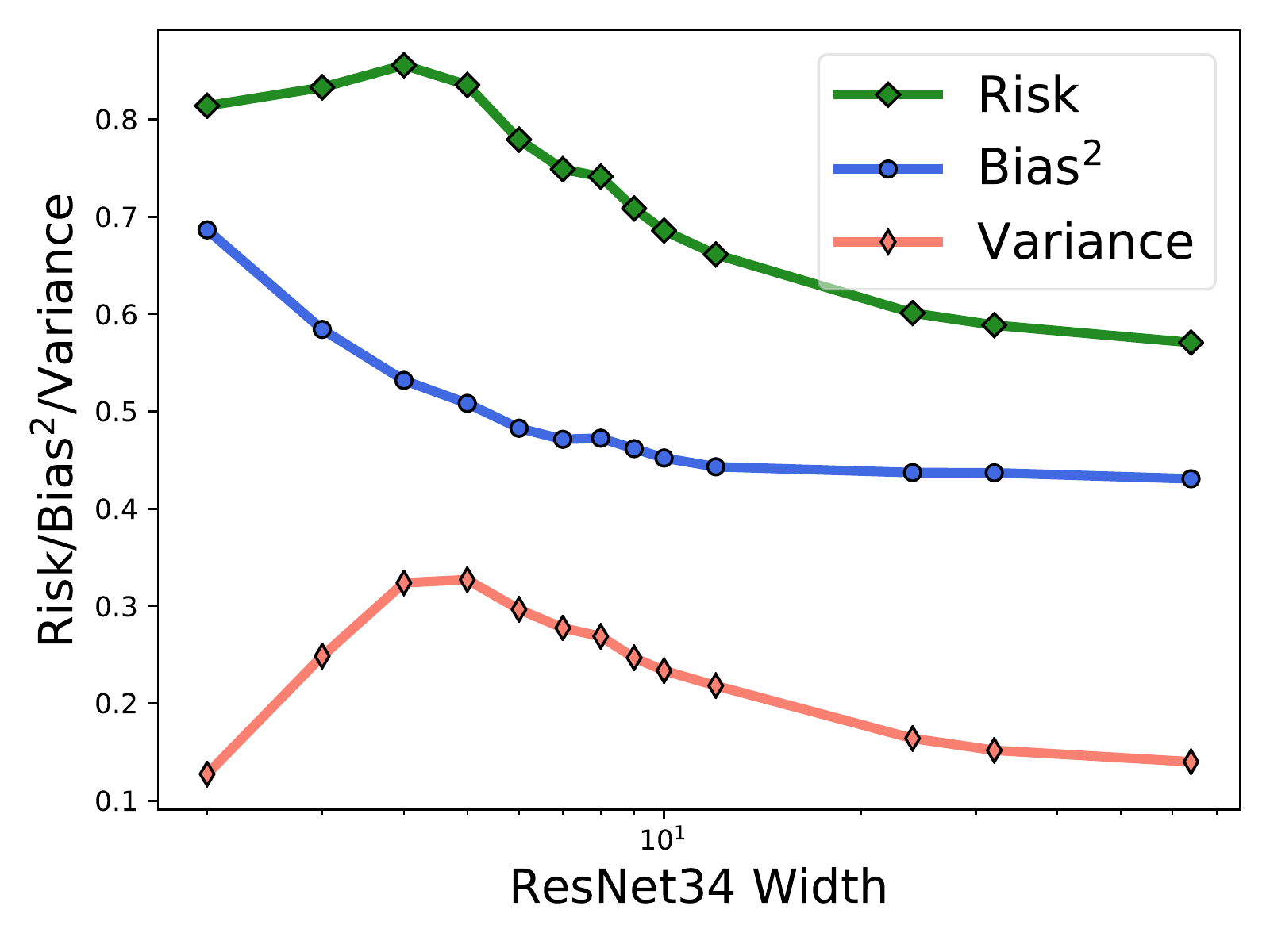}
    }
    \subfigure{
    \includegraphics[width=.32\textwidth]{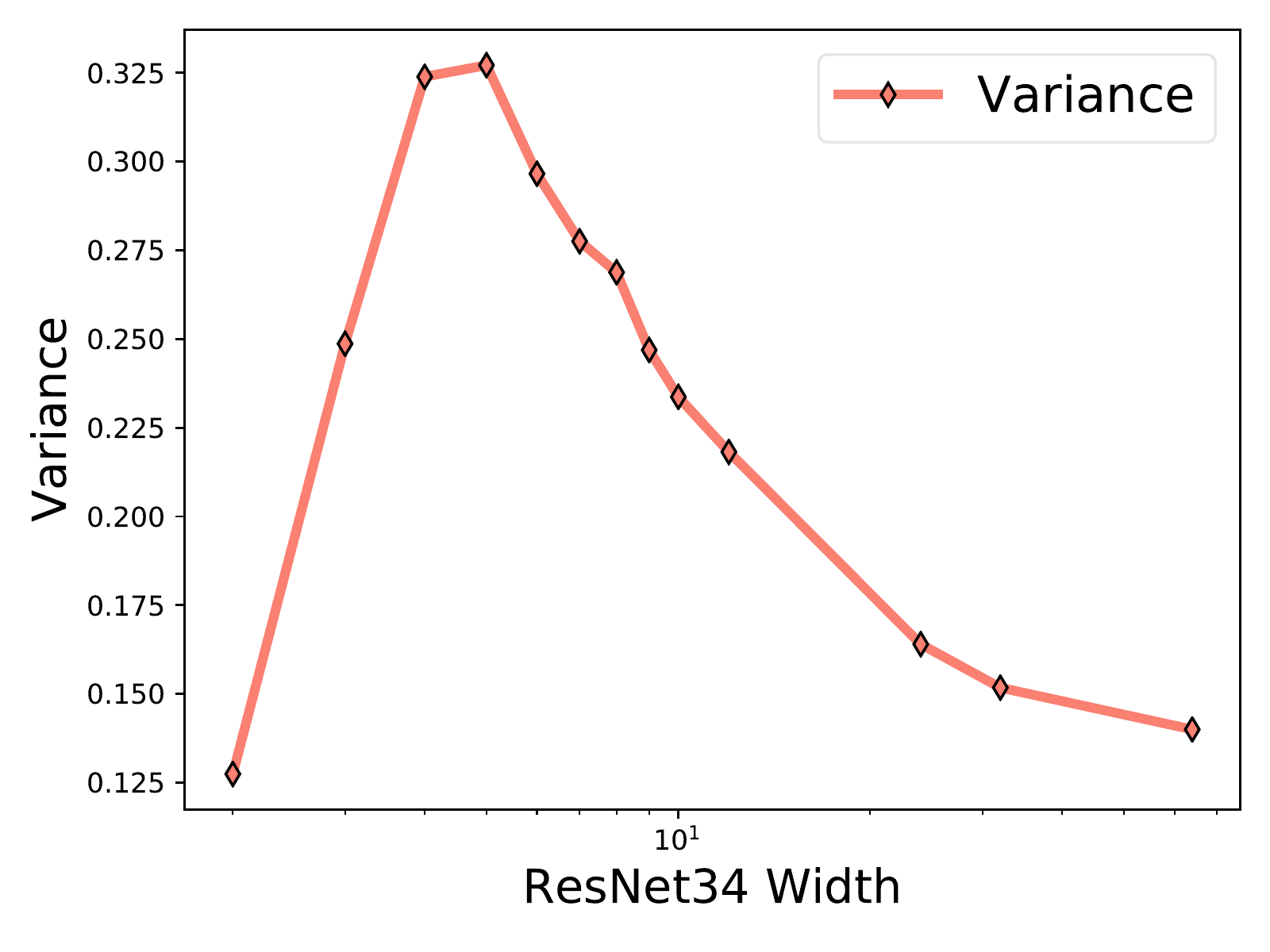}
    }
    \subfigure{
    \includegraphics[width=.32\textwidth]{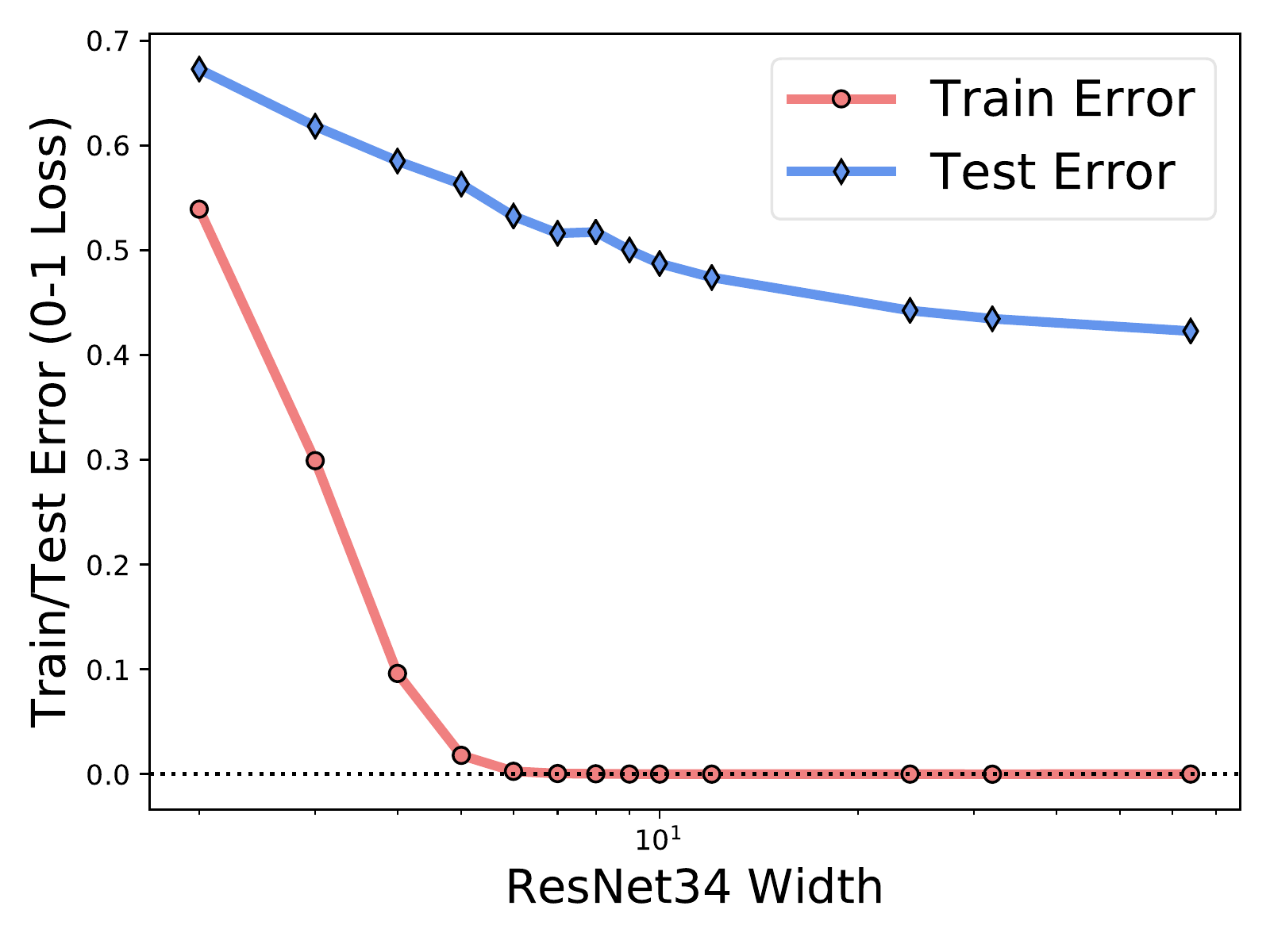}
    }
    \vskip -0.1in
    \caption{Risk, bias, variance, and train/test error for ResNet34 trained by cross-entropy loss (estimated by MSE bias-variance decomposition) on  CIFAR100 (10,000 training samples). (\textbf{Left}) Risk, bias, and variance for ResNet34. (\textbf{Middle}) Variance for ResNet34. (\textbf{Right}) Train error and test error for ResNet34.}
    \label{fig:dataset-cifar100}
  \end{center}
\end{figure}

We provide the results on Fashion-MNIST dataset in Figure~\ref{fig:data_fmnist}, which is mentioned in  \textsection\ref{section:phenomenonrobust}. 
We study the behavior of risk, bias, and variance of ResNet34 on CIFAR100 dataset. Because the number of class is large, we use cross-entropy during training, and apply the classical Bias-Vairance decomposition for MSE in \eqref{eq:bias} and \eqref{eq:variance} to estimate the risk, bias, and variance. As shown in Figure~\ref{fig:dataset-cifar100}, we observe the bell-shaped variance curve and the monotonically decreasing bias curve on CIFAR100 dataset.

\subsection{Training Size}\label{sec:app-train-size}
Appart from the 2 splits case in Figure~\ref{fig:mainline}, we also consider 5 splits (10,000 training samples) and  20 splits case (2,500 training samples). We present the 5 splits case (10,000 training samples) in Figure~\ref{fig:sample-5split}, which corresponds to the label 0$\%$ case in Figure~\ref{fig:noise}. We present the 20 splits (2,500 training samples) in Figure~\ref{fig:sample-20split}. With less number of training samples, both the bias and the variance will increase.

\subsection{Weight Decay}\label{sec:app-wd}
\begin{figure}[h]
  \begin{center}
    \subfigure{\includegraphics[width=.32\textwidth]{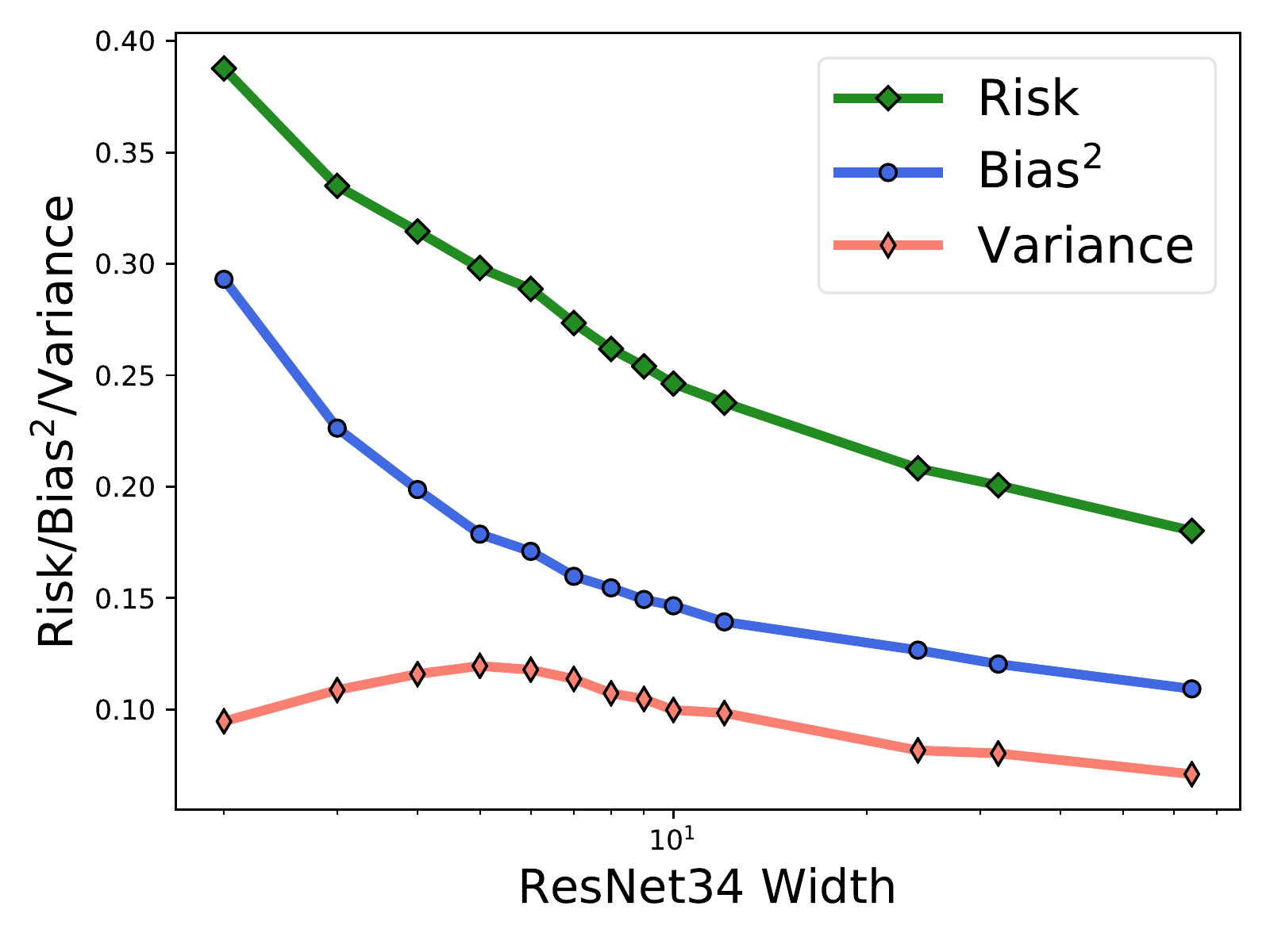}
    }
    \subfigure{
    \includegraphics[width=.32\textwidth]{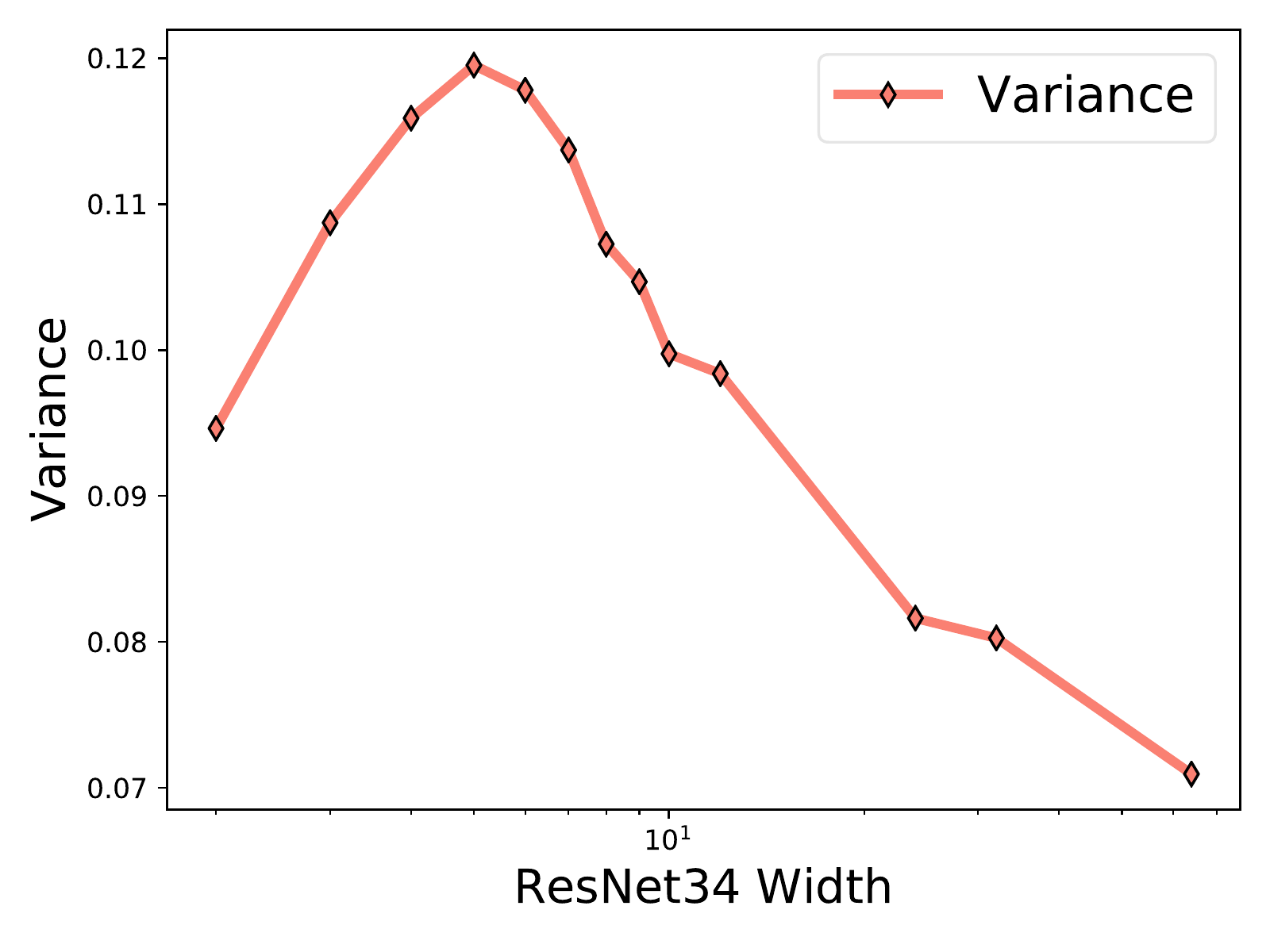}
    }
    \subfigure{
    \includegraphics[width=.32\textwidth]{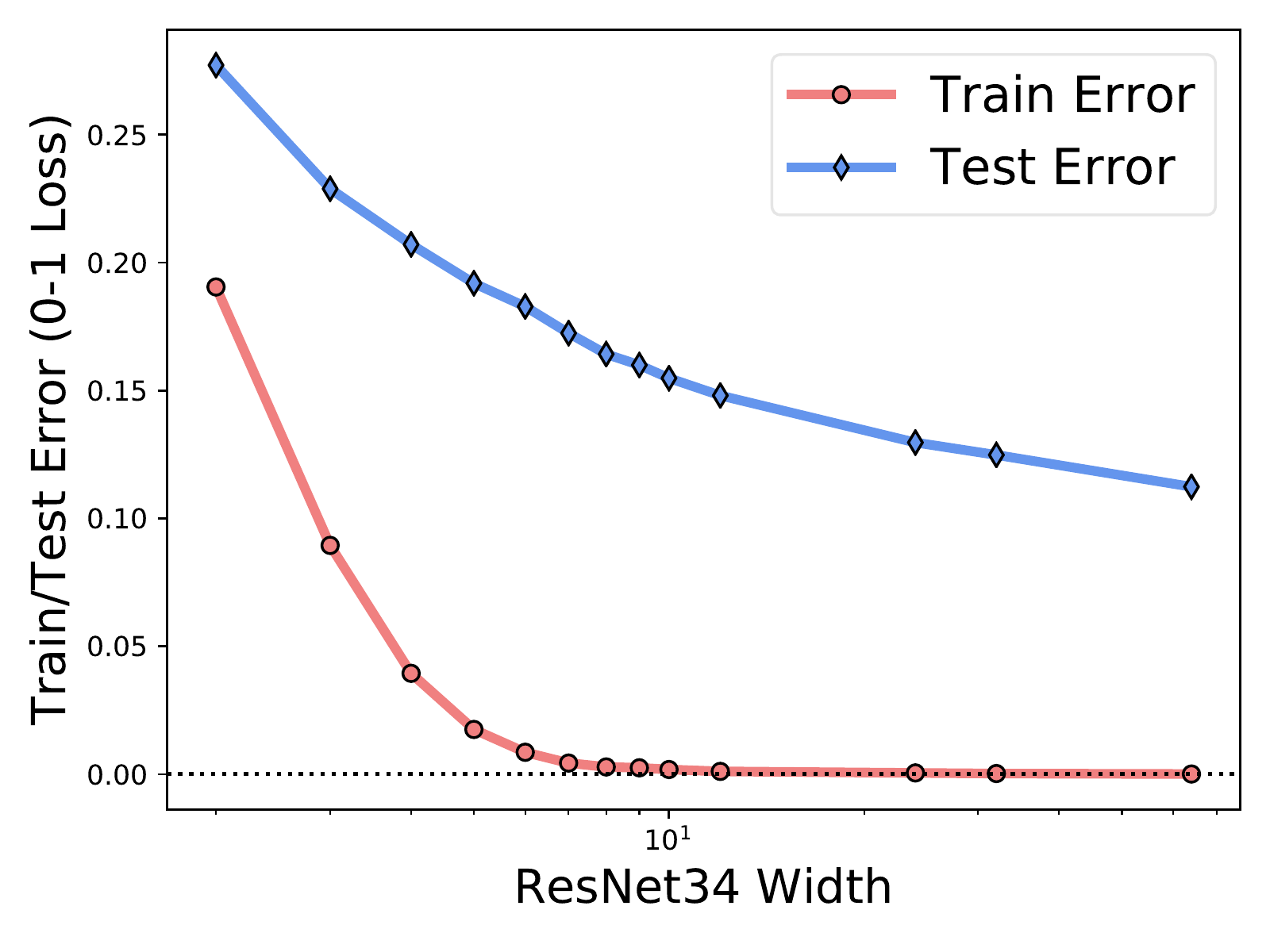}
    }
    \vskip -0.1in
    \caption{Risk, bias, variance, train/test error for ResNet34 trained by MSE loss on CIFAR10 dataset (10,000 training samples). (\textbf{Left}) Risk, bias, and variance for ResNet34. (\textbf{Middle}) Variance for ResNet34. (\textbf{Right}) Train error and test error for ResNet34.}
    \label{fig:sample-5split}
  \end{center}
  \vskip -0.2in
\end{figure}

\begin{figure}[h]
  \begin{center}
    \subfigure{\includegraphics[width=.32\textwidth]{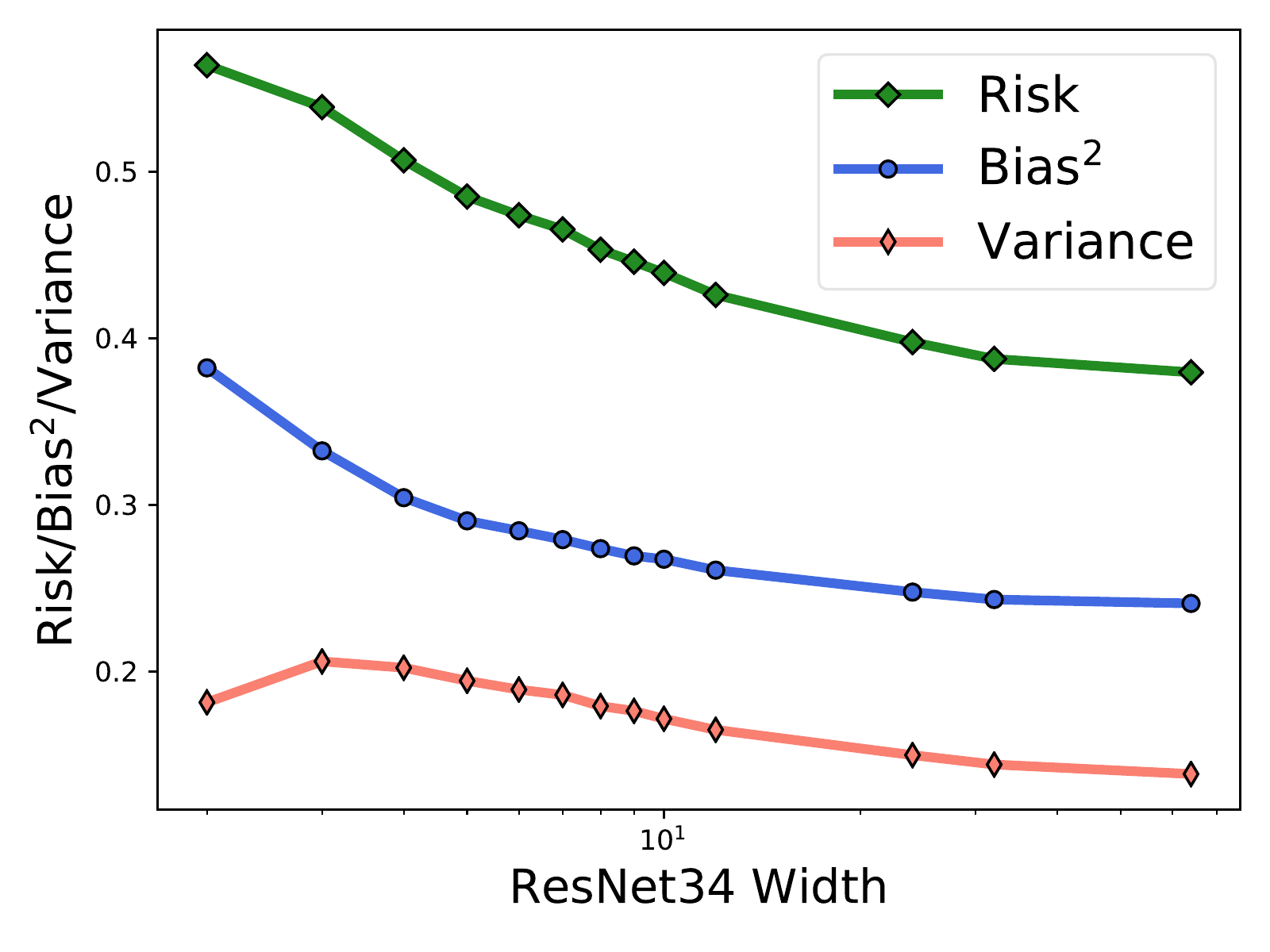}
    }
    \subfigure{
    \includegraphics[width=.32\textwidth]{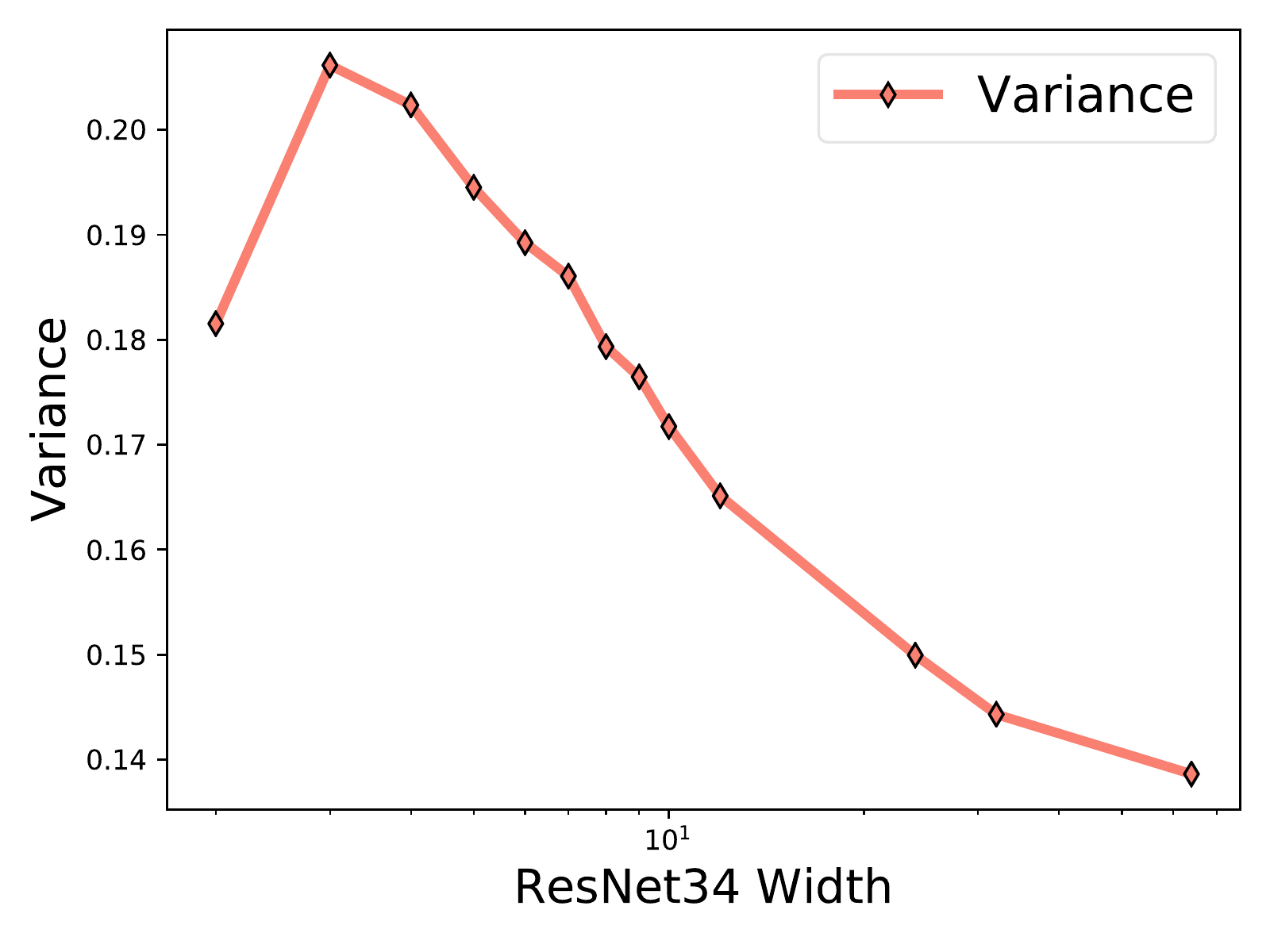}
    }
    \subfigure{
    \includegraphics[width=.32\textwidth]{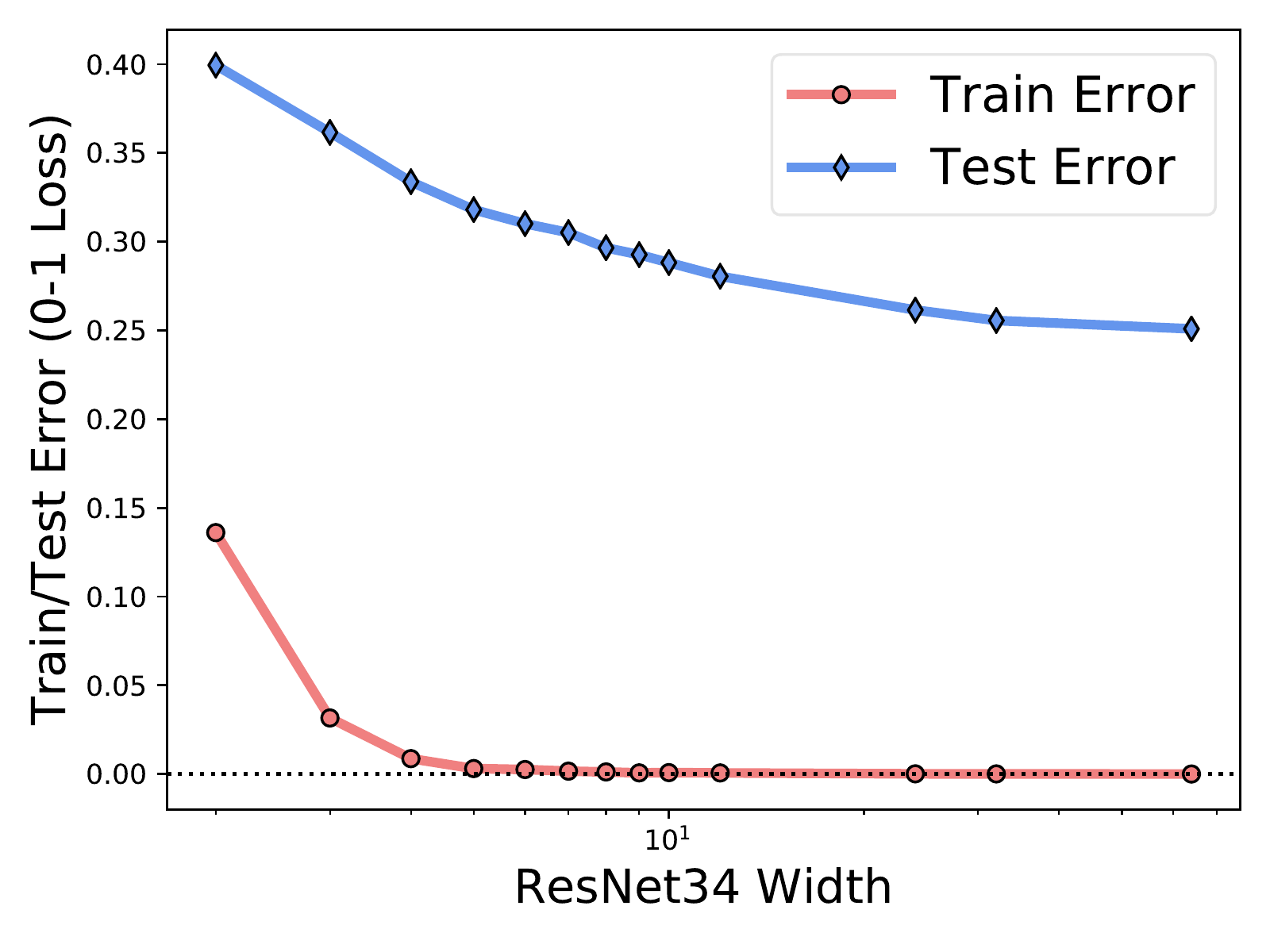}
    }
    \vskip -0.1in
    \caption{Risk, bias, variance, train/test error for ResNet34 trained by MSE loss on CIFAR10 dataset (2,500 training samples). (\textbf{Left}) Risk, bias, and variance for ResNet34. (\textbf{Middle}) Variance for ResNet34. (\textbf{Right}) Train error and test error for ResNet34.}
    \label{fig:sample-20split}
  \end{center}
  \vskip -0.1in
\end{figure}

\begin{figure}[h]
  \begin{center}
    \subfigure{\includegraphics[width=.32\textwidth]{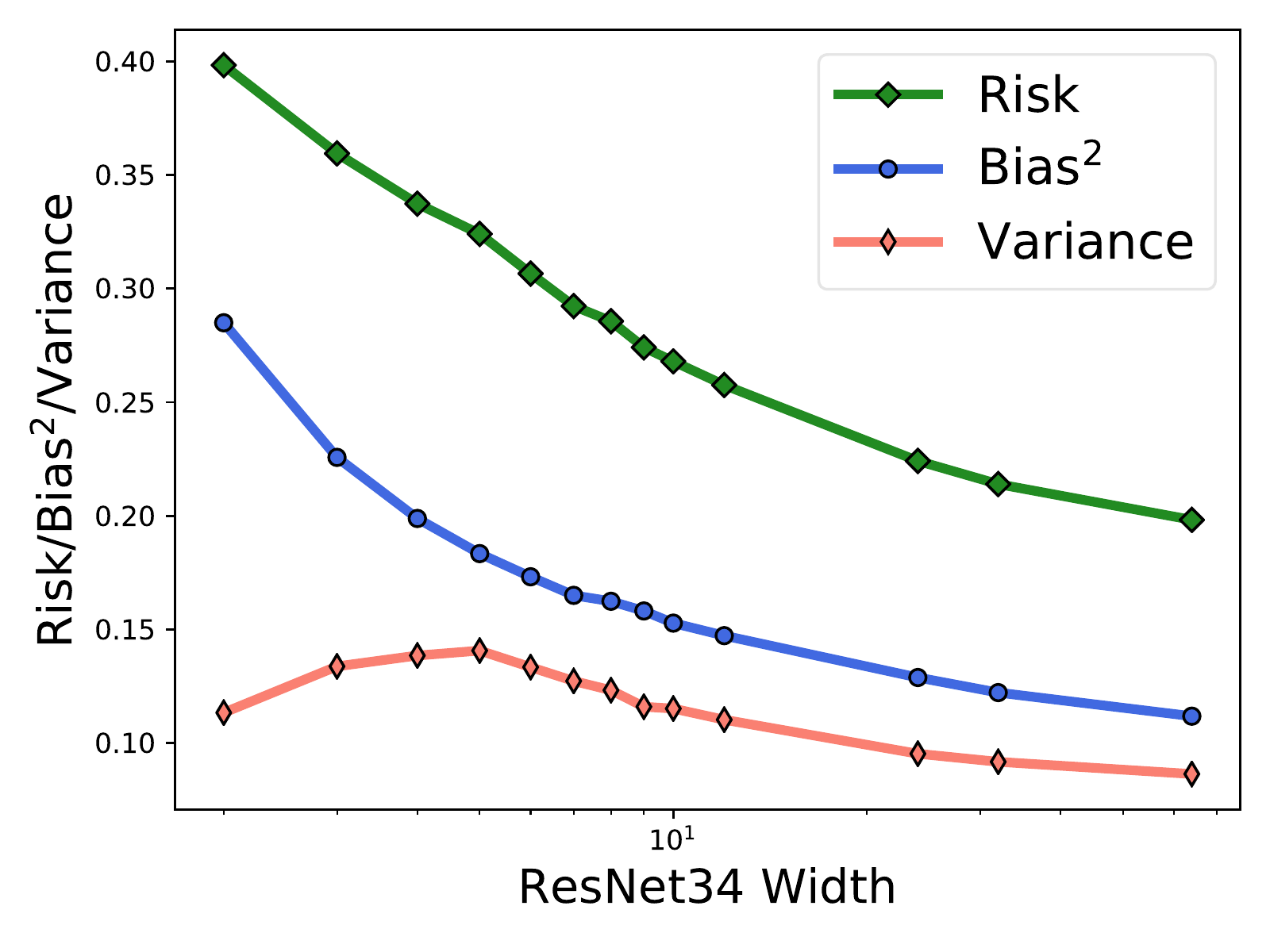}
    }
    \subfigure{
    \includegraphics[width=.32\textwidth]{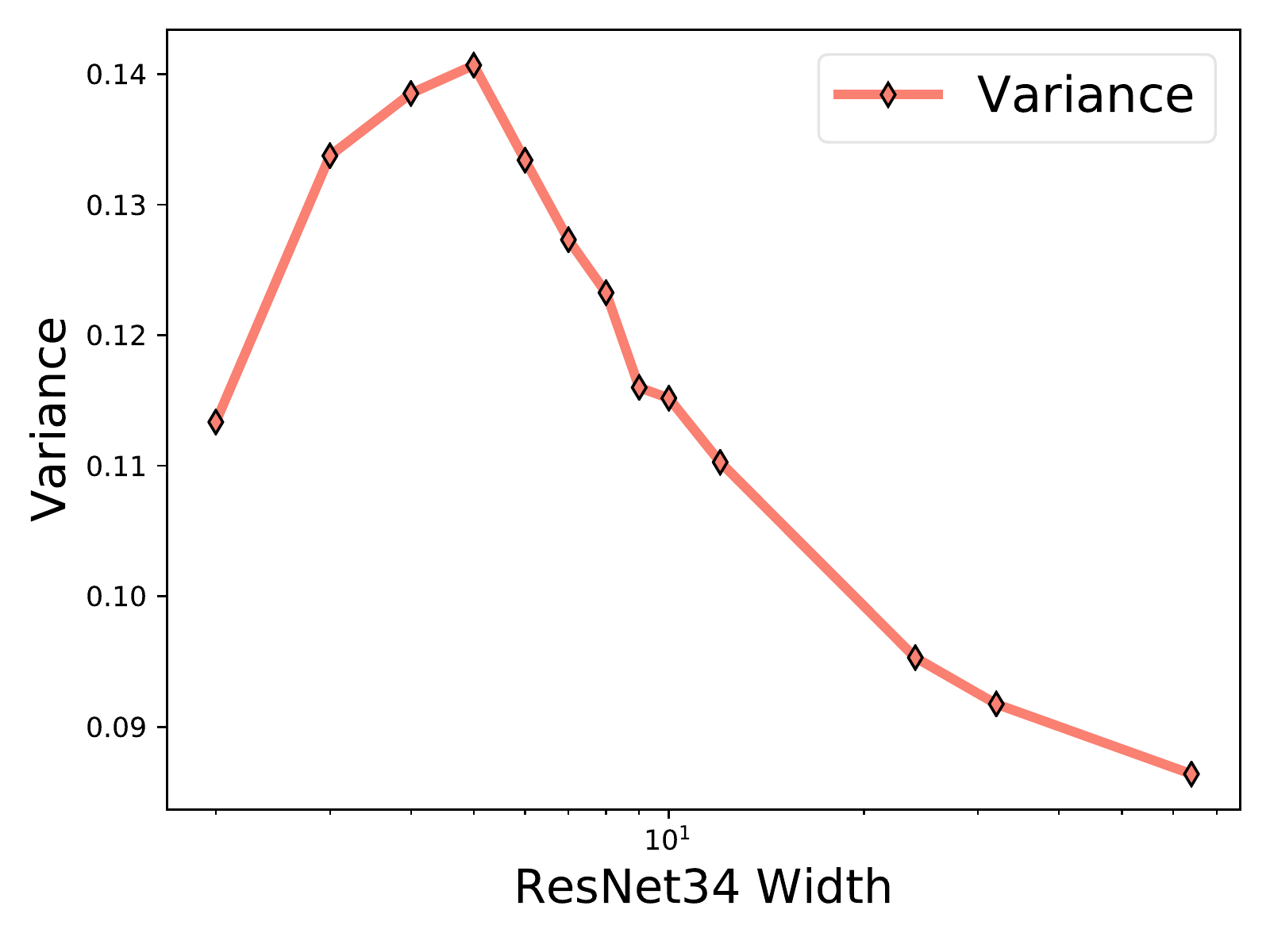}
    }
    \subfigure{
    \includegraphics[width=.32\textwidth]{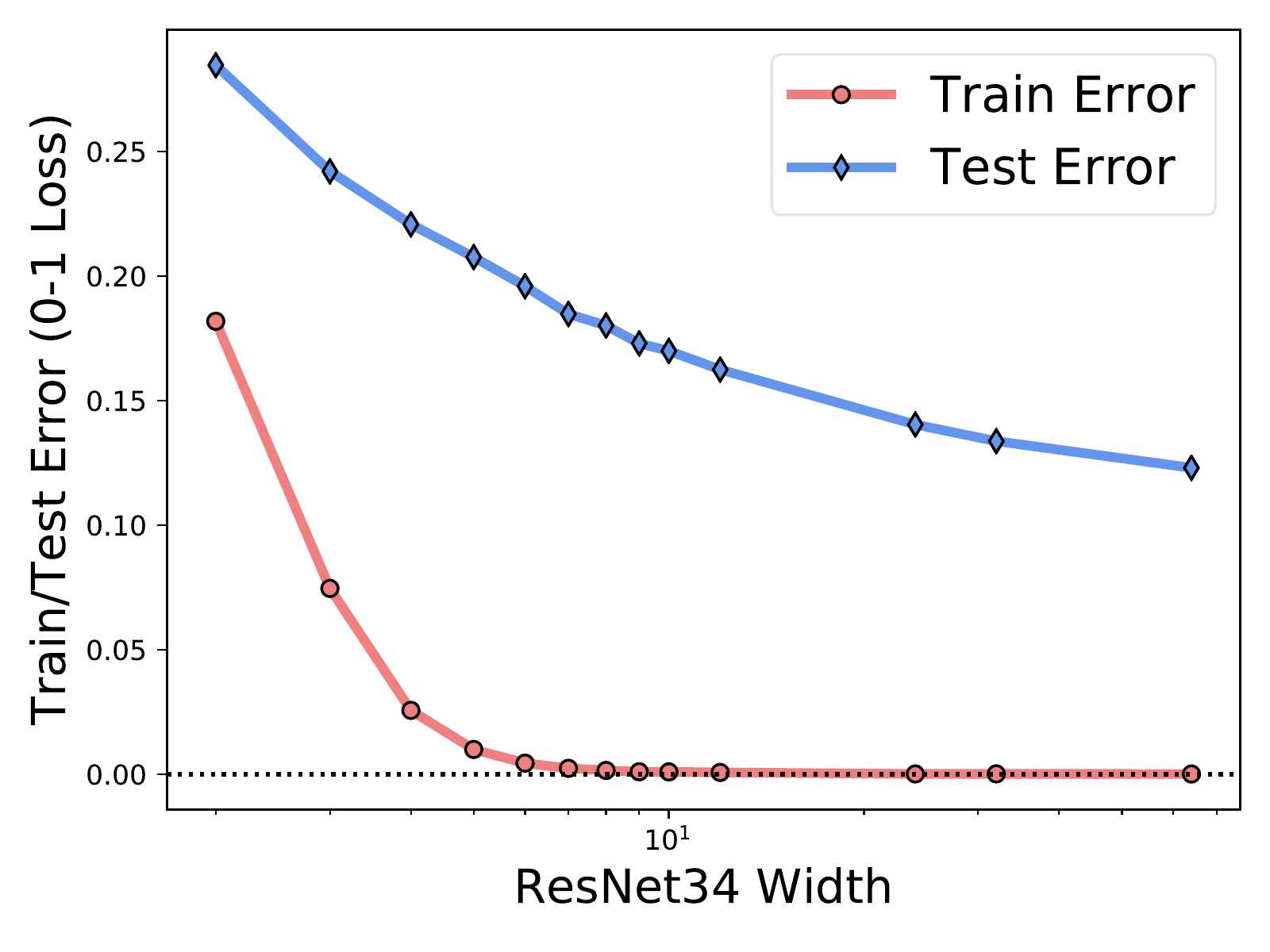}
    }
    \vskip -0.1in
    \caption{Risk, bias, variance, train/test error for ResNet34 trained by MSE loss on CIFAR10 dataset (10,000 training samples), the weight decay parameter of SGD is 1e-4. (\textbf{Left}) Risk, bias, and variance for ResNet34. (\textbf{Middle}) Variance for ResNet34. (\textbf{Right}) Train error and test error for ResNet34.}
    \label{fig:weightdecay-1e4}
  \end{center}
\end{figure}

We study another different weight decay parameter, (wd=\texttt{1e-4}) for ResNet34 on CIFAR10 dataset (10,000 training samples). The risk, bias, variance, and train/test error curves are shown in Figure~\ref{fig:weightdecay-1e4}. Compared with Figure~\ref{fig:sample-5split}, we observe that larger weight decay can decrease the variance.

\subsection{Label Noise}\label{sec:app-labelnoise}
\begin{figure}[h]
  \begin{center}
    {
    \includegraphics[width=.32\textwidth]{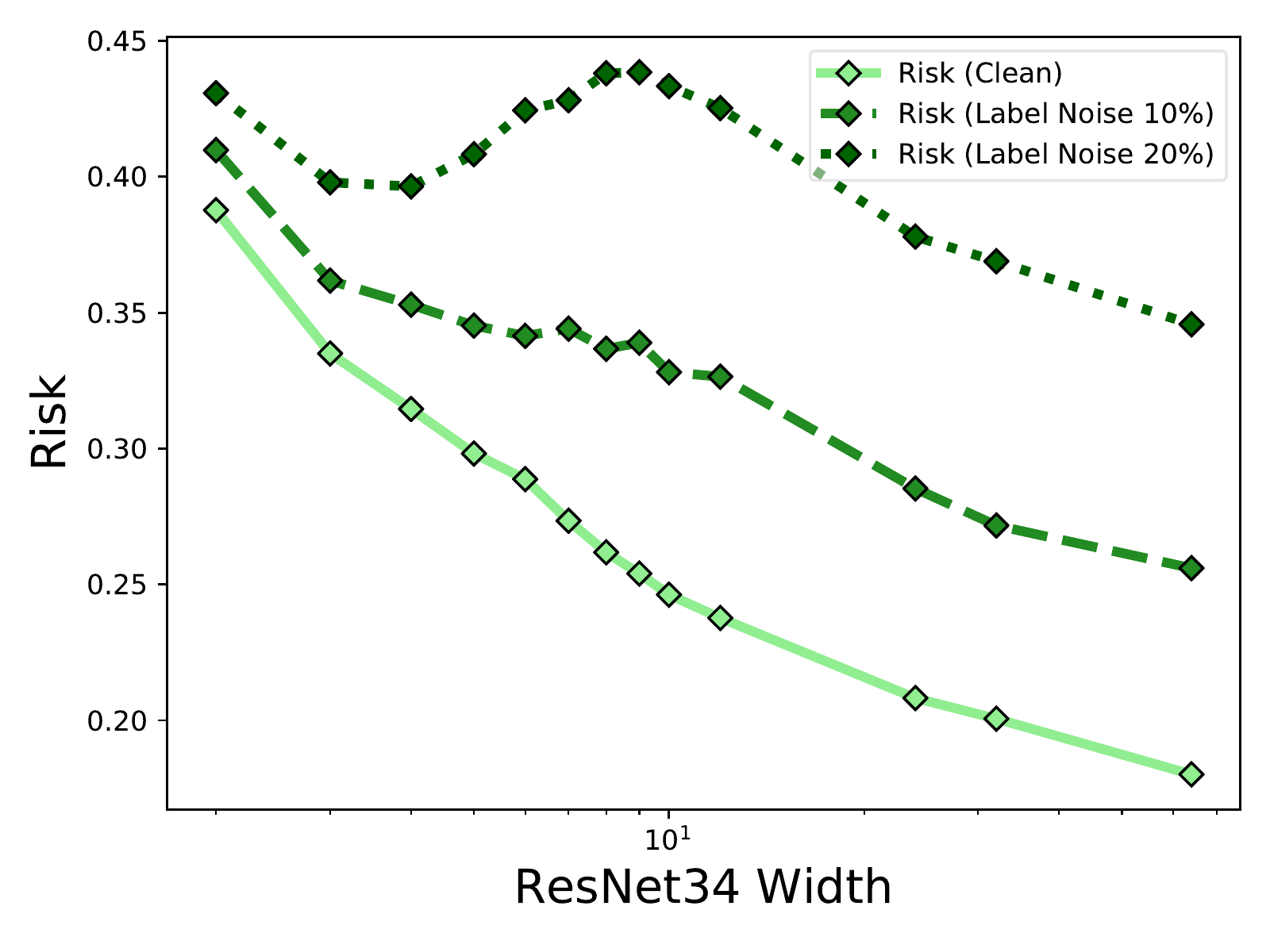}
    }
    \vskip -0.1in
    \caption{Risk under different label noise percentage. Increasing label noise leads to double descent risk curve.}
    \label{fig:appendix-risk-noise}
  \end{center}
  \vskip -0.1in
\end{figure}

We provide the risk curve for ResNet34 under different label noise percentage  as described in  \textsection\ref{section:doubledescent}, and the results are shown in Figure~\ref{fig:appendix-risk-noise}.

\subsection{0-1 Loss Bias-Variance Decomposition}
\begin{figure}[h]
  \begin{center}
    \subfigure{\includegraphics[width=.32\textwidth]{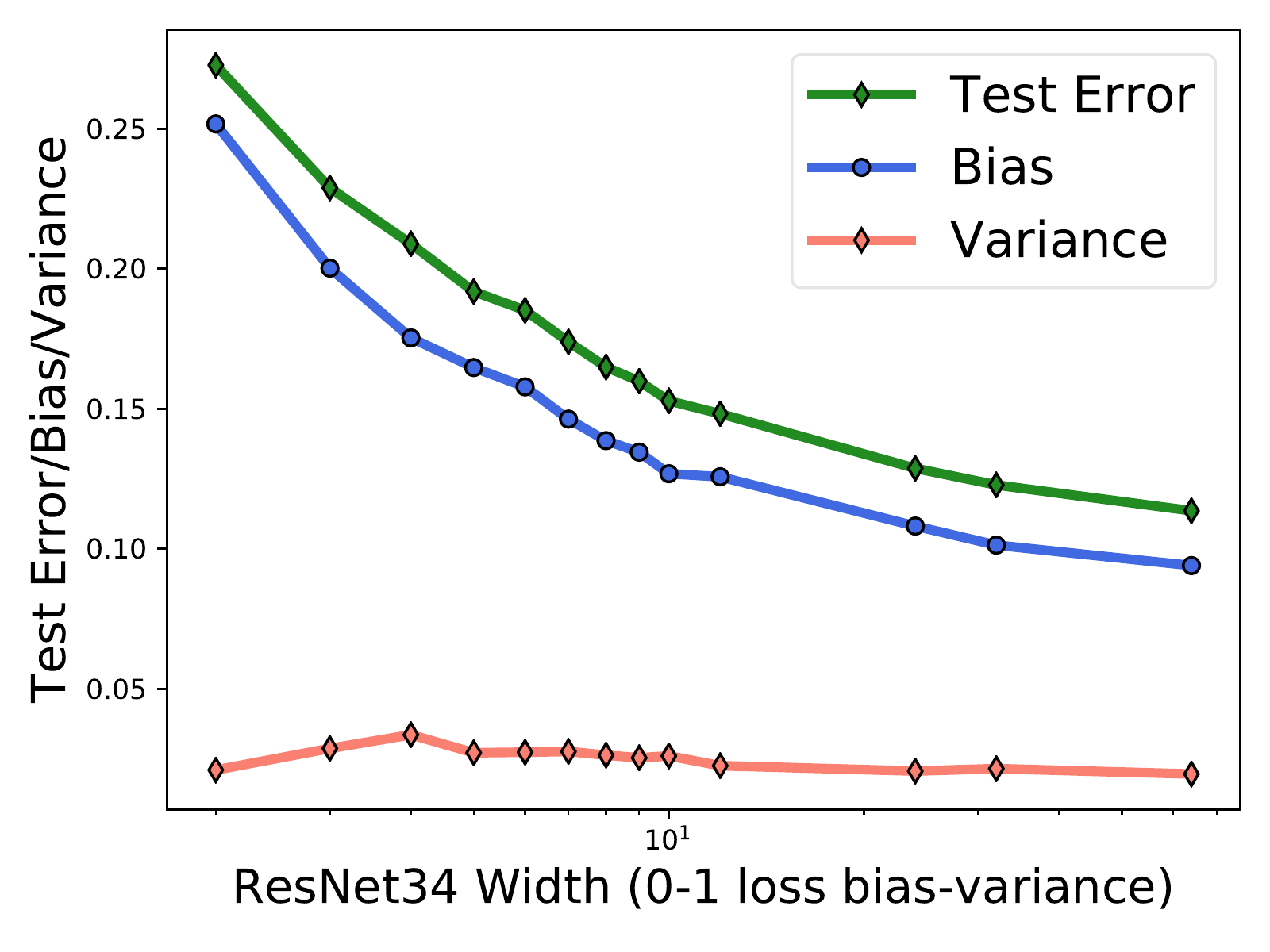}
    }
    \subfigure{\includegraphics[width=.32\textwidth]{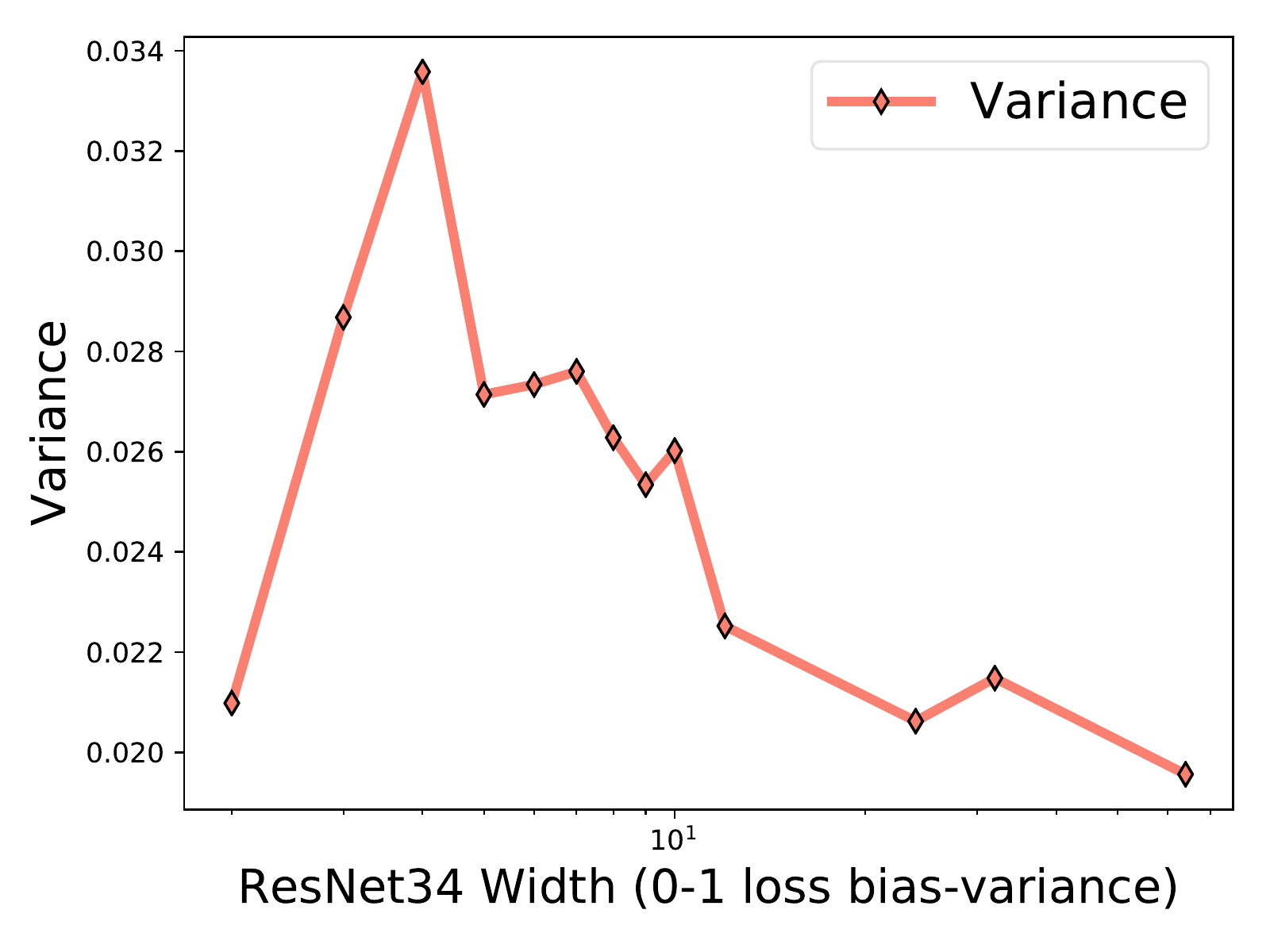}
    }
    \vskip -0.1in
    \caption{Bias-variance (0-1 loss), and test error for ResNet34 trained by MSE loss on CIFAR10 dataset (10,000 training samples). (\textbf{Left}) Bias and variance (0-1 loss), and test error for ResNet34. (\textbf{Right}) Variance (0-1 loss) for ResNet34. }
    \label{fig:0-1-bv}
  \end{center}
  \vskip -0.1in
\end{figure}

We evaluated the bias and variance for 0-1 loss (defined in \citet{dietterich1995machine}) on the CIFAR10 dataset with 10,000 training samples using ResNet34. The results are shown in Figure~\ref{fig:0-1-bv}. We can consistently observe that the bias is monotonically decreasing and the variance is unimodal.

\clearpage

\subsection{Sources of Error for Mean Squared Error (MSE)}\label{appendix:mse_error}
\begin{figure}[ht]
  \begin{center}
    \subfigure{\includegraphics[width=.45\textwidth]{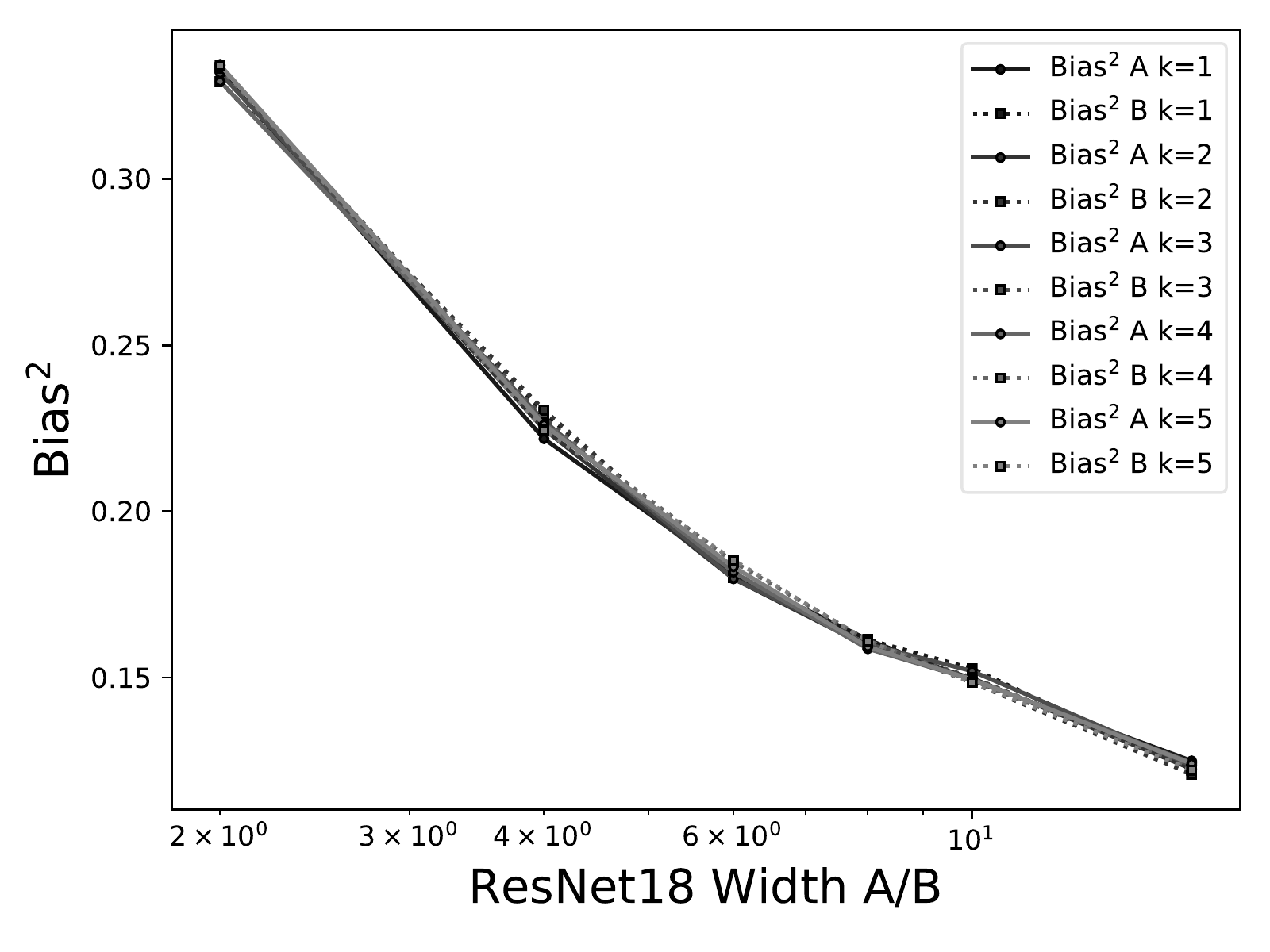}
    }
    \subfigure{
    \includegraphics[width=.45\textwidth]{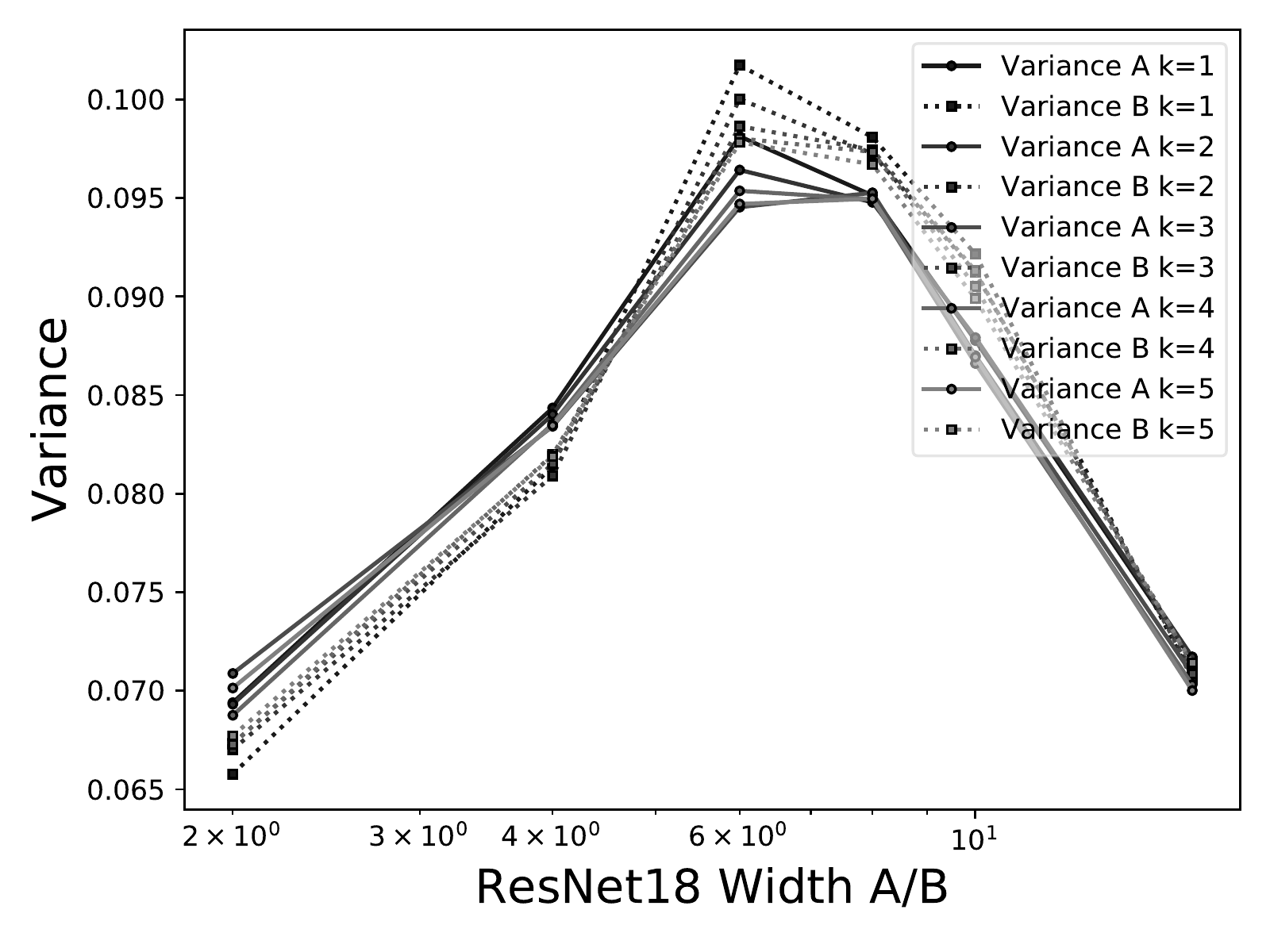}
    }
    \caption{Bias and variance for two portions of data with $k$ from 1 to 5. (\textbf{Left}) Bias for ResNet18. (\textbf{Right}) Variance for ResNet18.}
    \label{fig:mse_error}
  \end{center}
  \vskip -0.2in
\end{figure}

As argued in  \textsection\ref{section:prelim_estimating} the estimator for variance is unbiased estimator. To understand the variance of the estimator, we first split the data into two parts, $A$ and $B$. For each part,
we take multiple random splits ($k$) and estimate the variance by taking the average of those estimators, and vary the number of random splits $k$. The results are shown in Figure \ref{fig:mse_error}. We can see that the variation between to parts of data is small. Quantitatively, veraging across different model width, the relative difference between two parts of data is 0.65\% for bias and 3.15\% for variance. 

\subsection{Sources of Error for Cross Entropy Loss (CE)}\label{appendix:ce_error}

\begin{figure}[ht]
  \begin{center}
    \subfigure{\includegraphics[width=.32\textwidth]{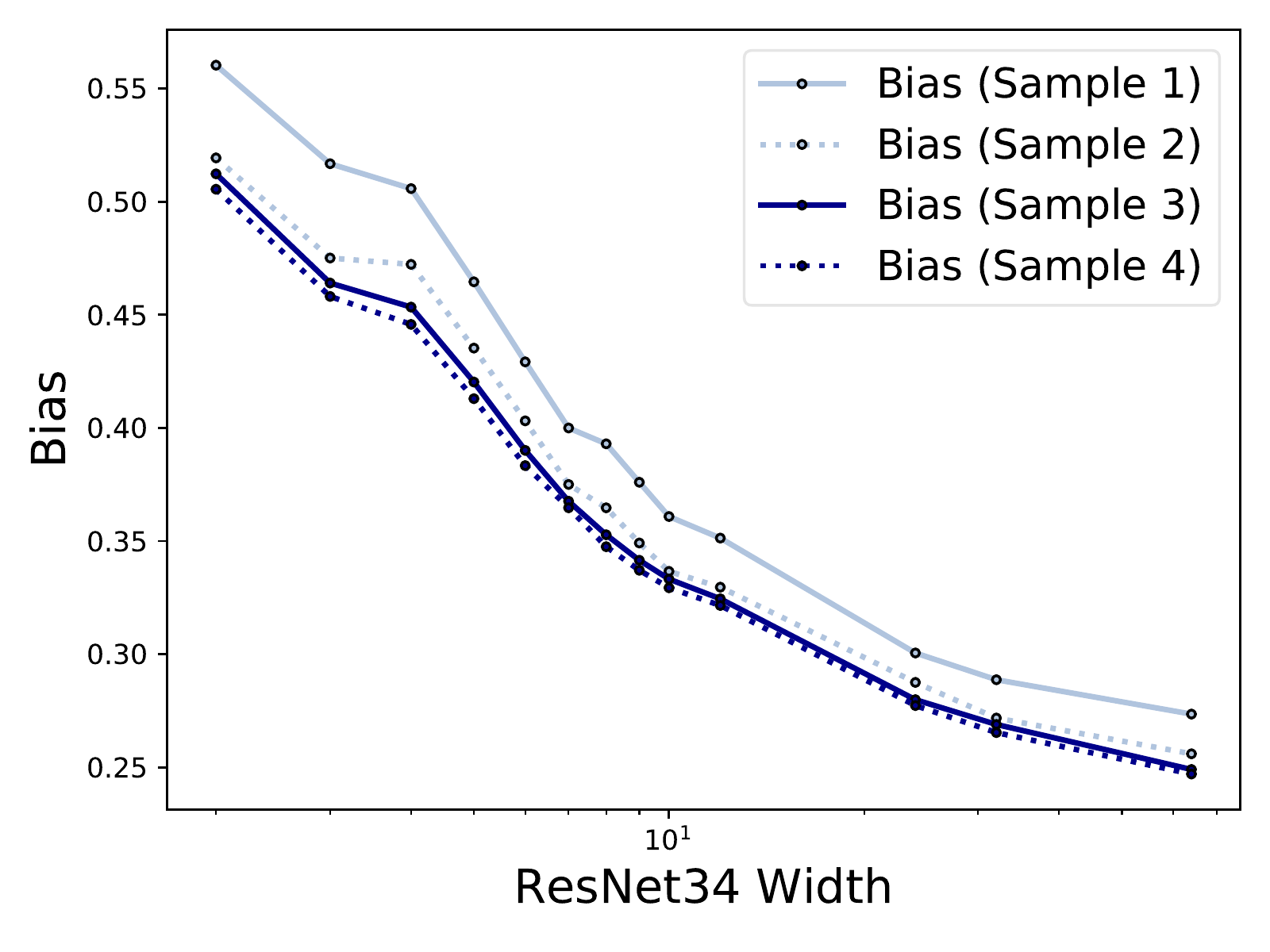}
    }
    \subfigure{
    \includegraphics[width=.32\textwidth]{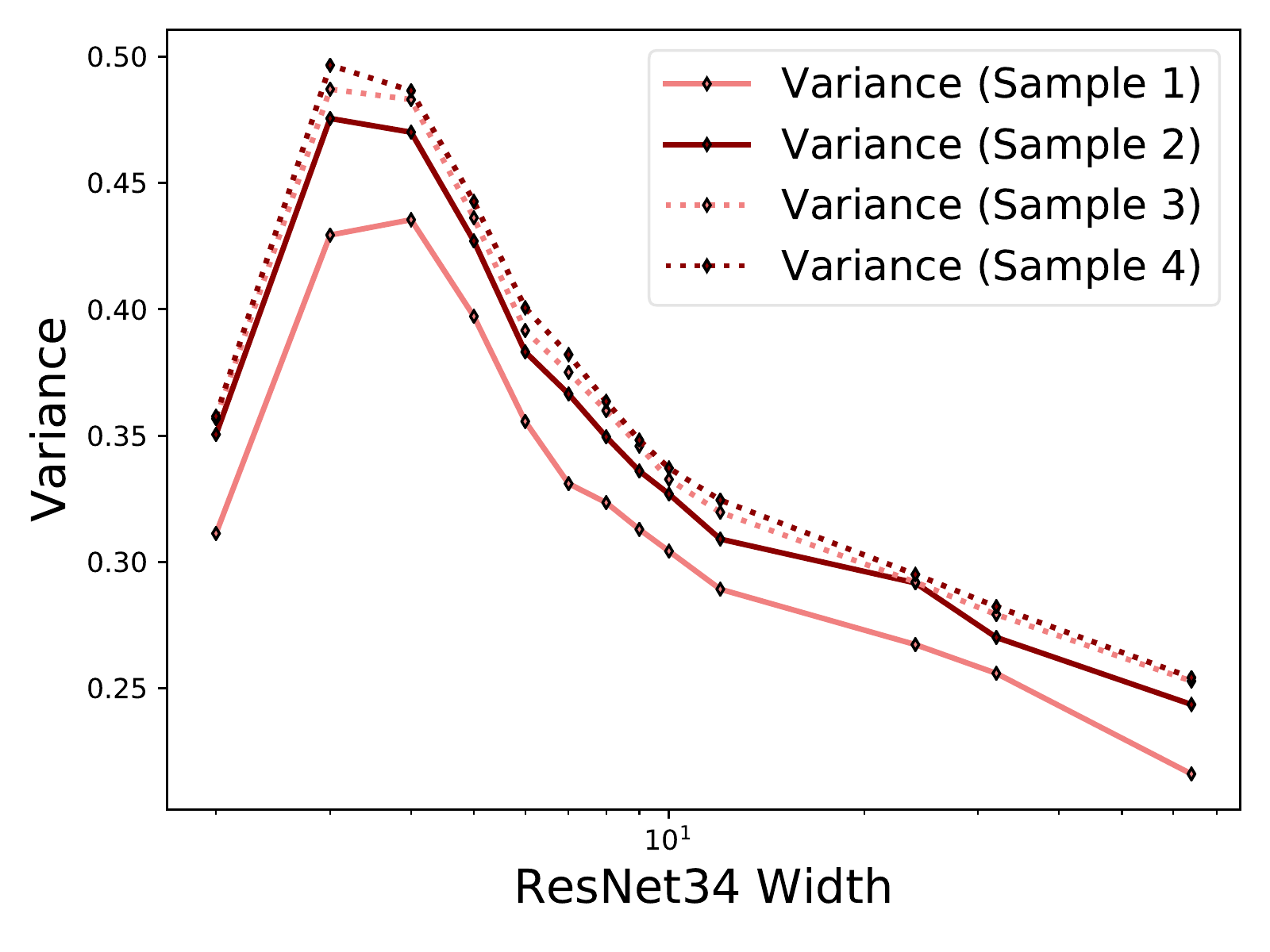}
    }
    \subfigure{
    \includegraphics[width=.32\textwidth]{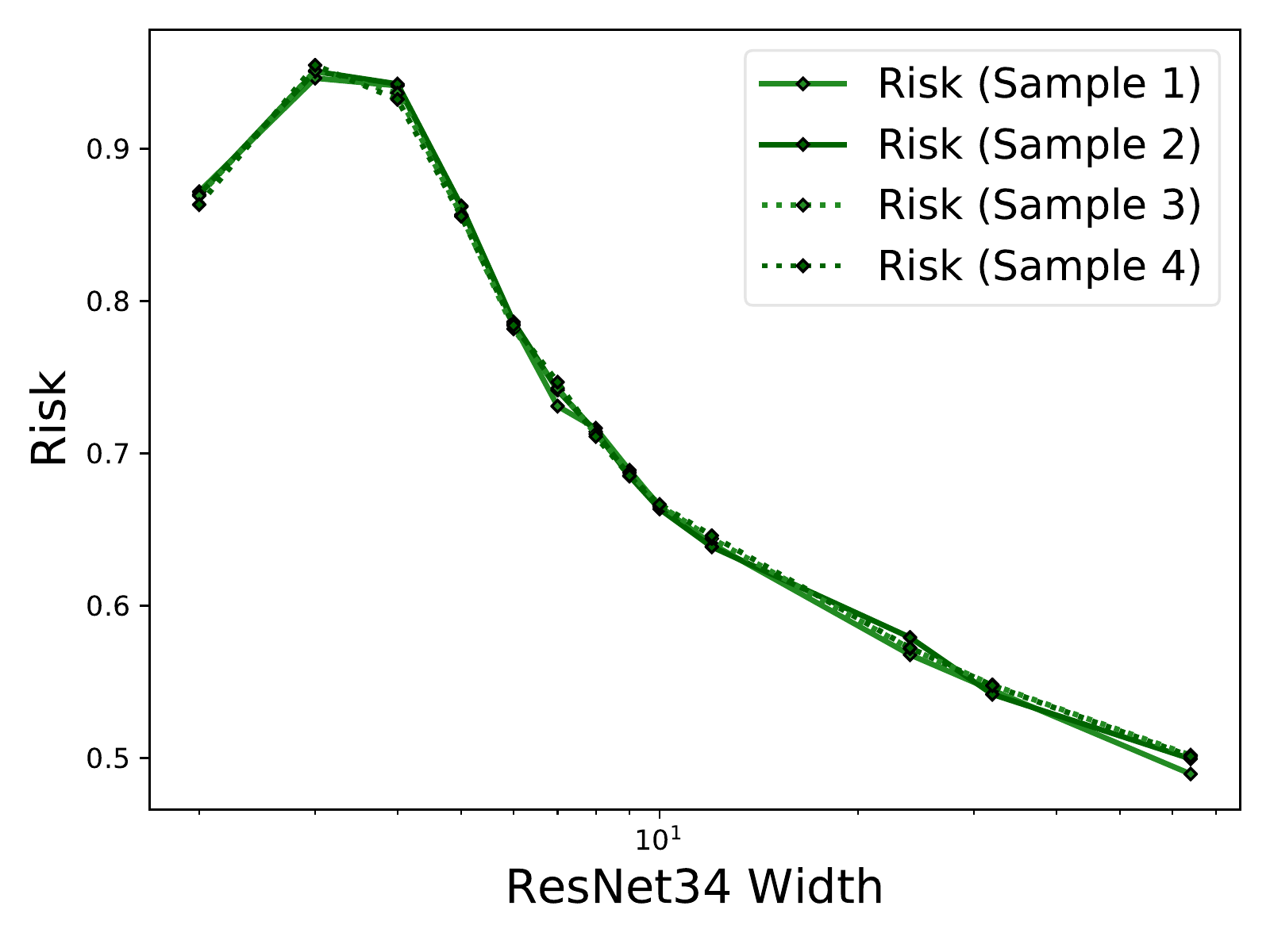}
    }
    \caption{Estimate of bias, variance, and risk using varying number of sample ($k$ in Algorithm~\ref{alg:example}). (\textbf{Left}) Bias (CE) for ResNet34. (\textbf{Middle}) Variance (CE) for ResNet34. (\textbf{Right}) Risk (CE) for ResNet34.}
    \label{fig:ce_error}
  \end{center}
  \vskip -0.2in
\end{figure}

For cross entropy loss, we are currently unable to obtain an unbiased estimator. We can access the quality of our estimator using the following scheme. We partition the dataset into five parts $\mc T_1, \dots, \mc T_5$, i.e., set $N=5$ in Algorithm~\ref{alg:example}. Then, we sequentially plot the estimate of bias and variance using $k=1, 2, 3, 4$ as described in Algorithm~\ref{alg:example}. Using larger $k$ gives better estimate. 
As shown in Figure \ref{fig:ce_error}, when $k$ is small, our estimator over-estimate the bias and under-estimate the variance, but the overall behavior of the curves are consistent.

\subsection{Effect of Depth on Bias and Variance for Out-Of-Distribution Data}\label{appendix:ood_depth}
\begin{figure}[ht]
  \begin{center}
    \subfigure{
    \includegraphics[width=.32\textwidth]{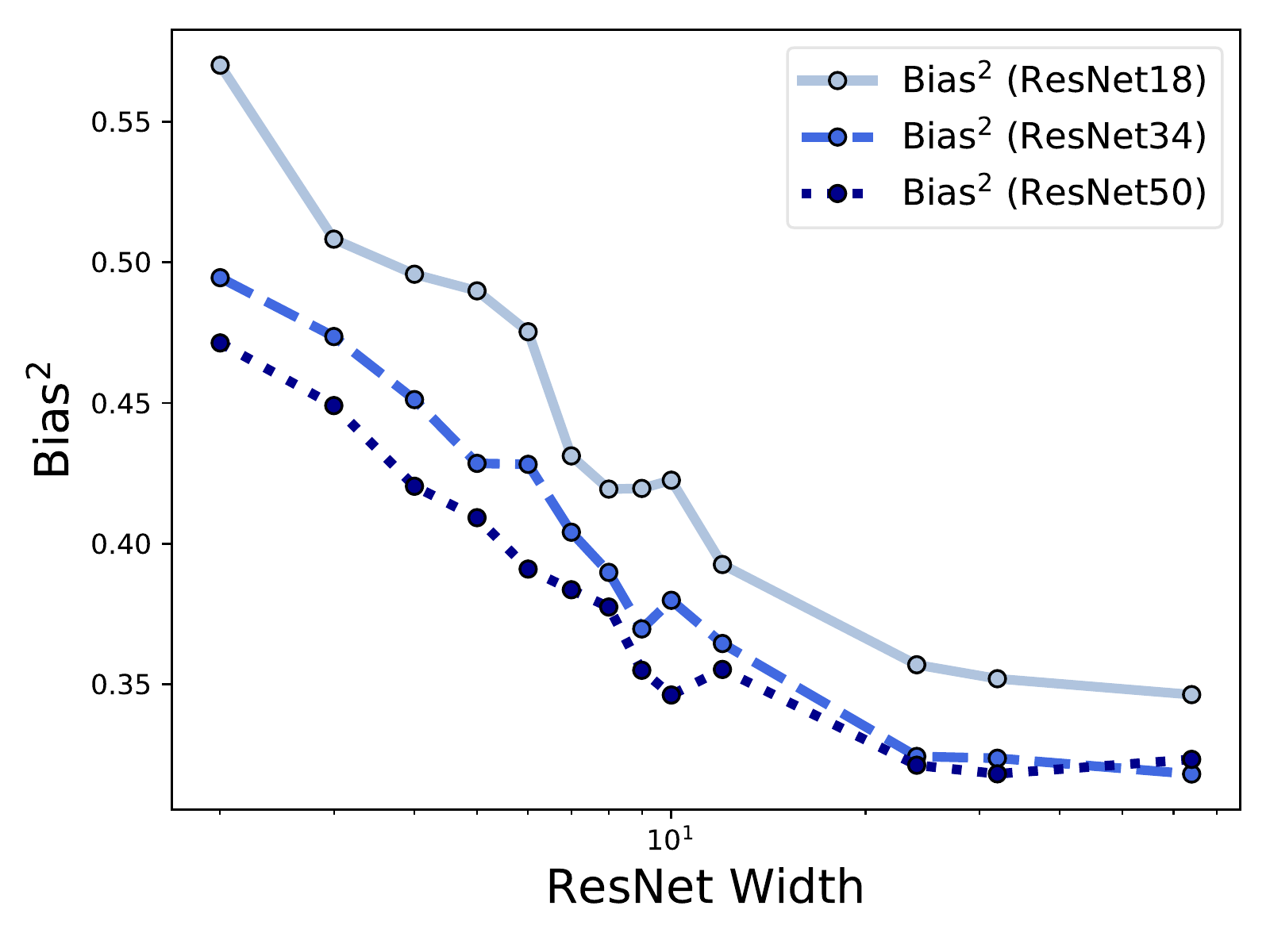}
    }
    \subfigure{\includegraphics[width=.32\textwidth]{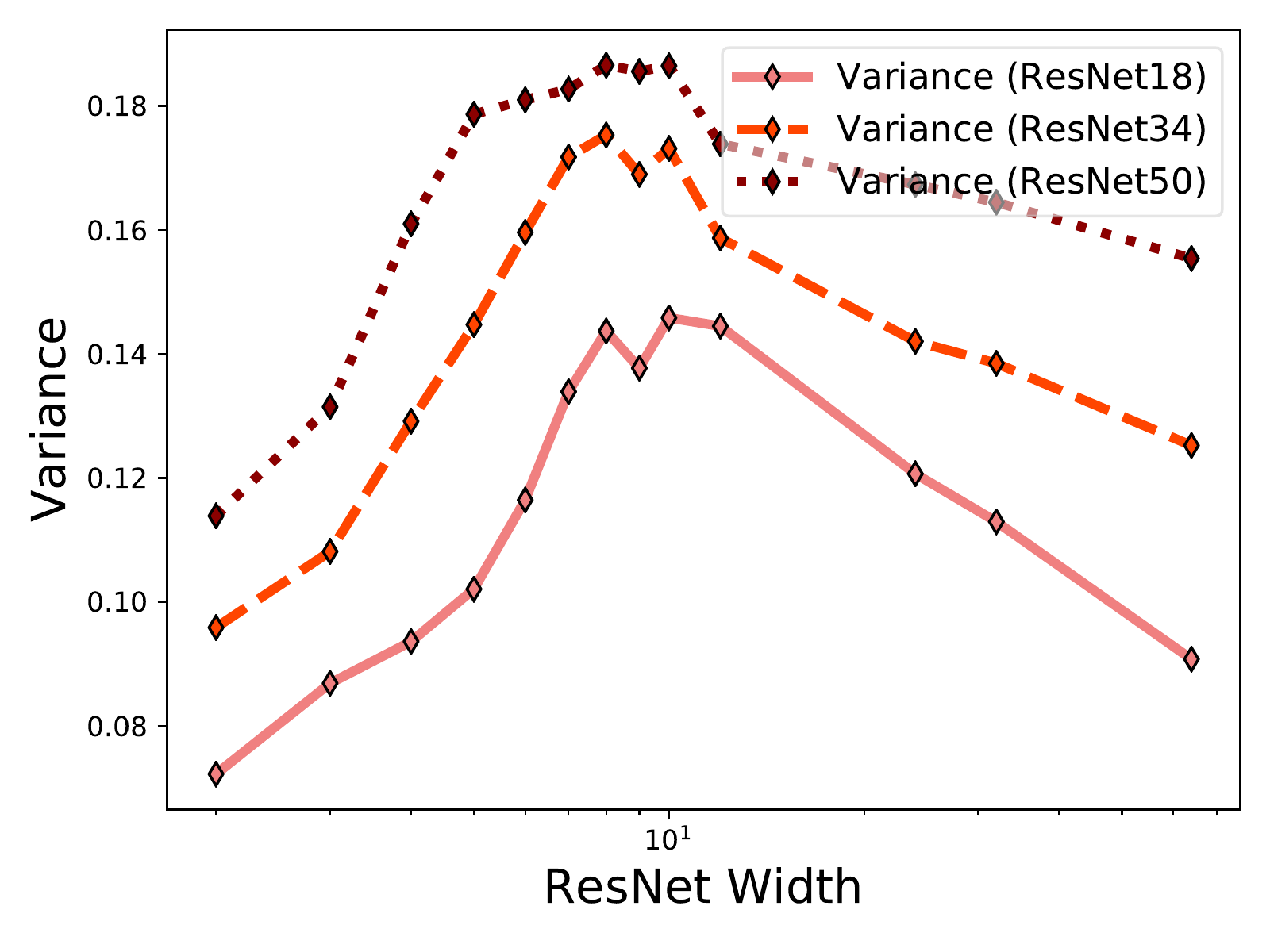}
    }
    \subfigure{\includegraphics[width=.32\textwidth]{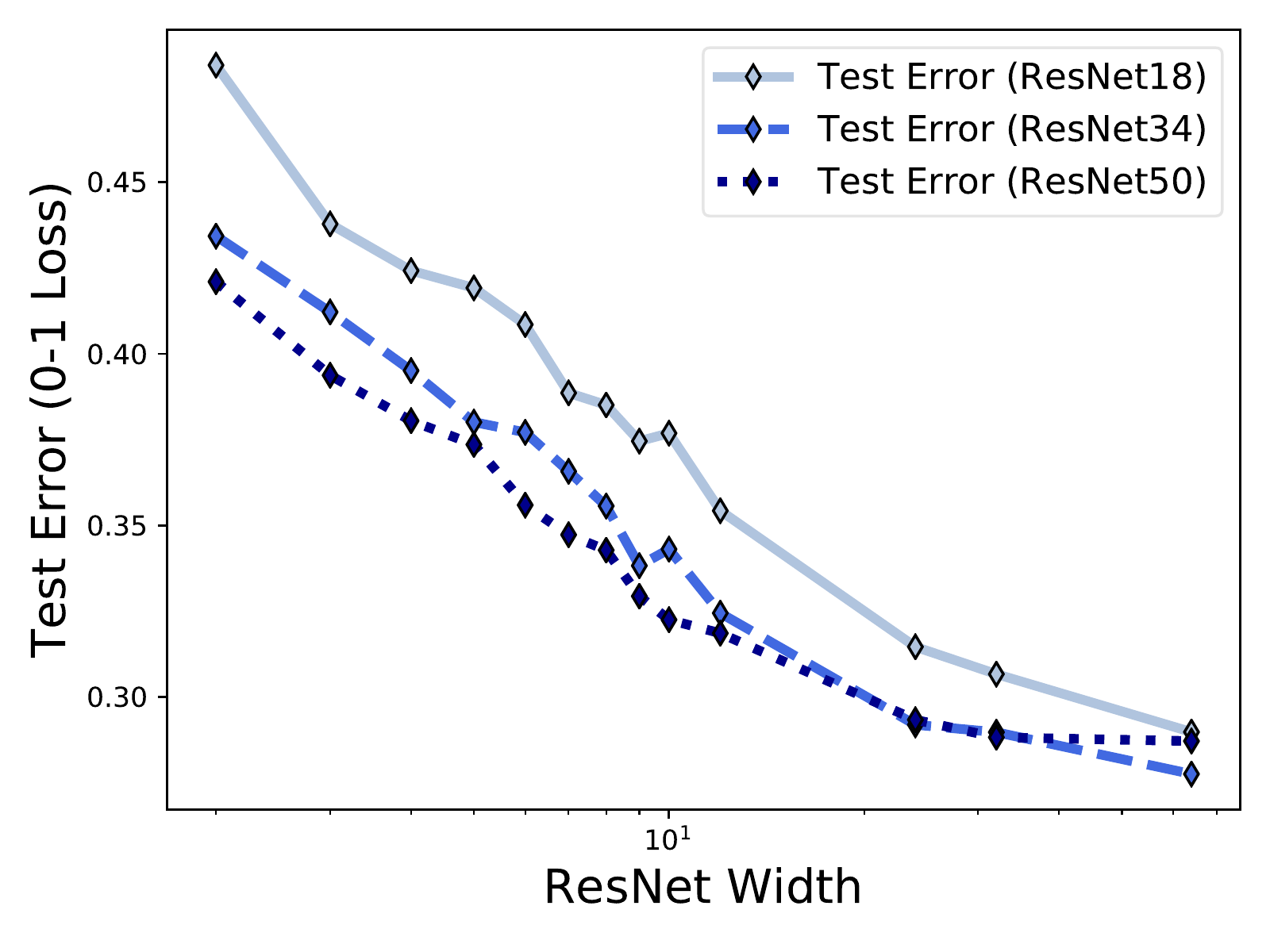}
    }
    \vskip -0.1in
    \caption{Bias, variance, and test error for ResNet with different depth (ResNet18, ResNet34 and ResNet50 trained by MSE loss on 25,000 CIFAR10 training samples) evaluated on out-of-distribution examples (CIFAR10-C dataset). (\textbf{Left}) Bias for ResNet18, ResNet34 and ResNet50. (\textbf{Middle}) Variance for ResNet18, ResNet34 and ResNet50. (\textbf{Right}) Test error for ResNet18, ResNet34 and ResNet50.}
    \label{fig:ood_depth}
  \end{center}
  \vskip -0.2in
\end{figure}

We study the role of depth on out-of-distribution test data. In Figure~\ref{fig:ood_depth}, we observe that increasing the depth can decrease the bias and increase the variance. Also, deeper ResNet can generalize better on CIFAR10-C dataset as shown in Figure~\ref{fig:ood_depth}.

\subsection{Effect of Depth on ResNet using Bottleneck Blocks}\label{appendix:bottleneck}
\begin{figure}[h]
  \begin{center}
    \subfigure{\includegraphics[width=.35\textwidth]{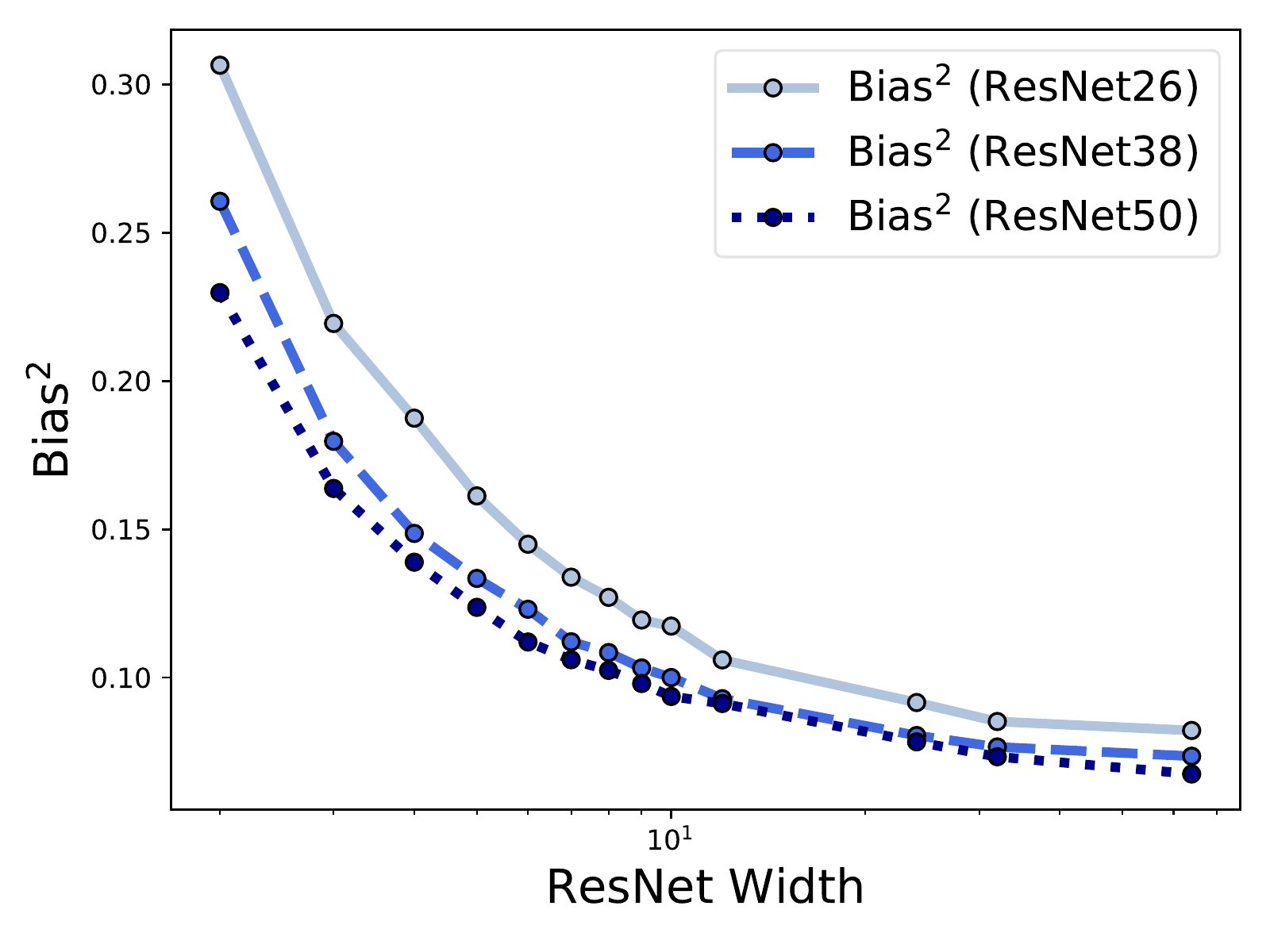}
    }
    \subfigure{
    \includegraphics[width=.35\textwidth]{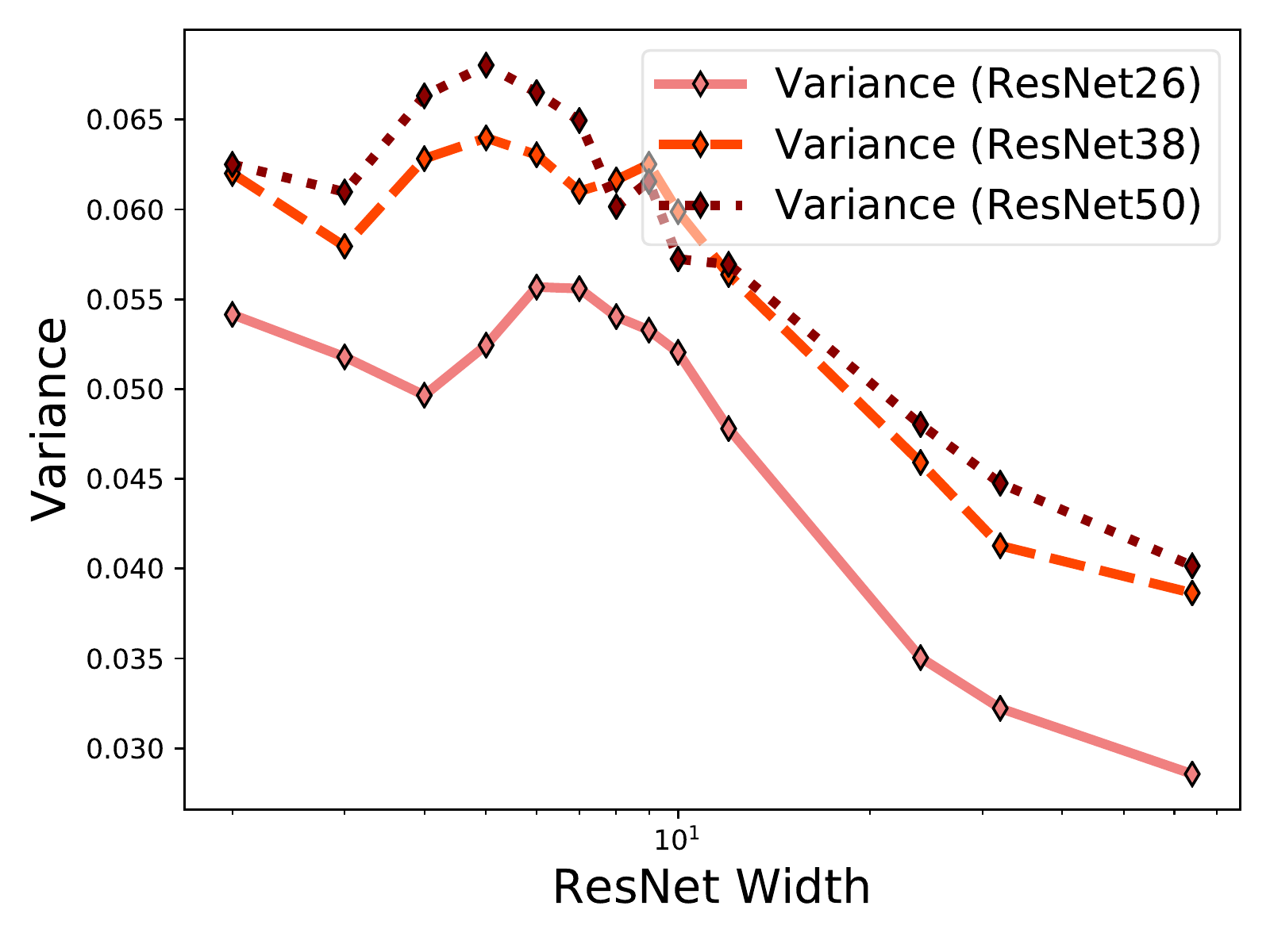}
    }
    \vskip -0.1in
    \caption{Bias and variance for ResNet (bottleneck block) with different depth. (\textbf{Left}) Bias for ResNet26, ResNet38 and ResNet50. (\textbf{Right}) Variance for ResNet26, ResNet38 and ResNet50. }
    \label{fig:depth-bottle}
  \end{center}
  \vskip -0.2in
\end{figure}

In order to study the role of depth for ResNet on bias and variance, we apply basic residual block for ResNet50. To better investigate the depth of ResNet, we use Bottleneck block for ResNet26, ResNet38, and ResNet50. More specifically, the number of 3-layer bottleneck blocks for ResNet26, ResNet38, and ResNet50 are $[2, 2, 2, 2]$, $[3, 3, 3, 3]$, and $[3, 4, 6, 3]$.  As shown in Figure~\ref{fig:depth-bottle}, we observe that deeper ResNet with Bottleneck blocks has lower bias and higher variance.

\subsection{Effect of Depth on VGG}\label{appendix:vgg}
\begin{figure}[h]
  \begin{center}
    \subfigure{\includegraphics[width=.35\textwidth]{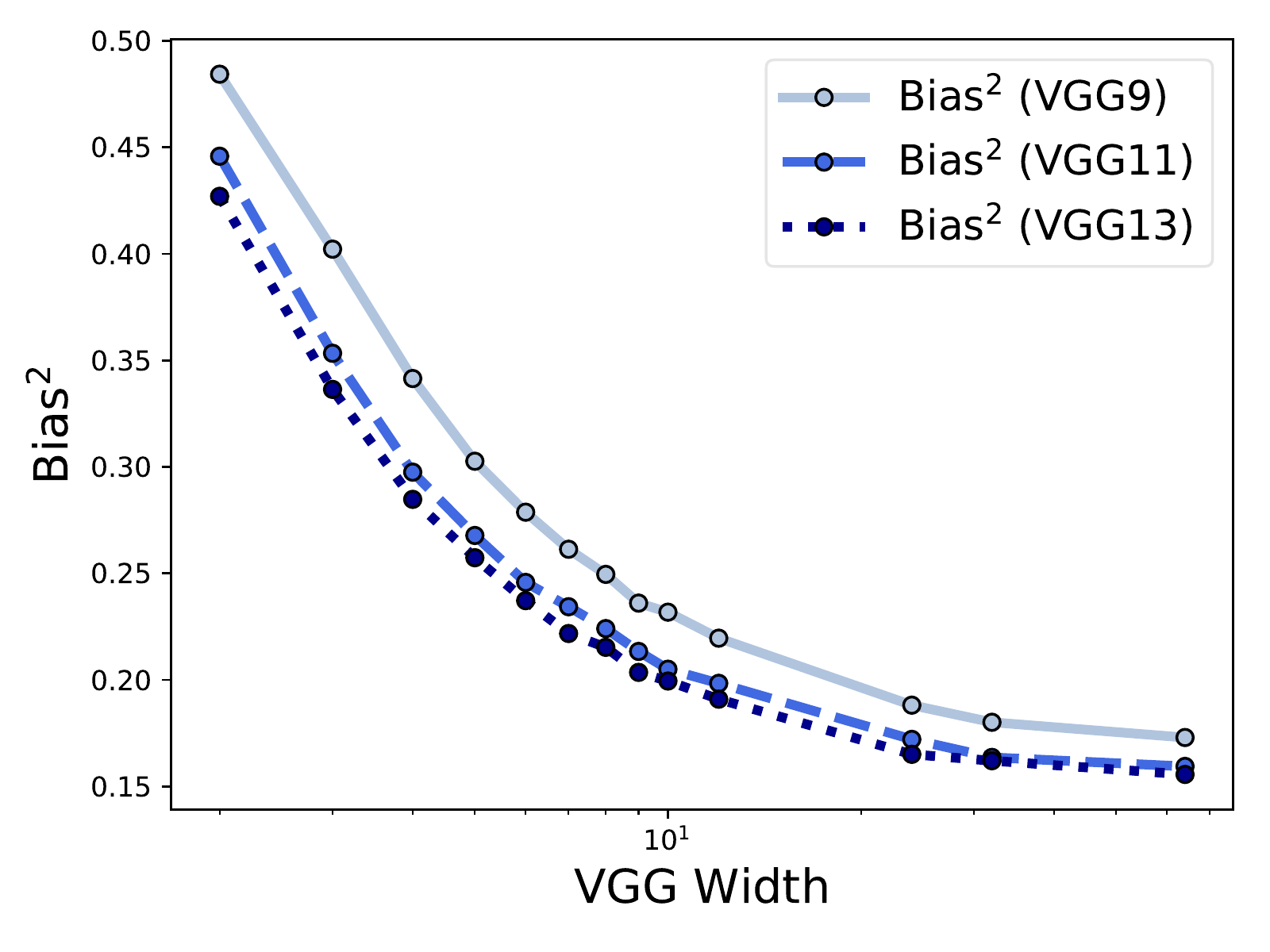}
    }
    \subfigure{
    \includegraphics[width=.35\textwidth]{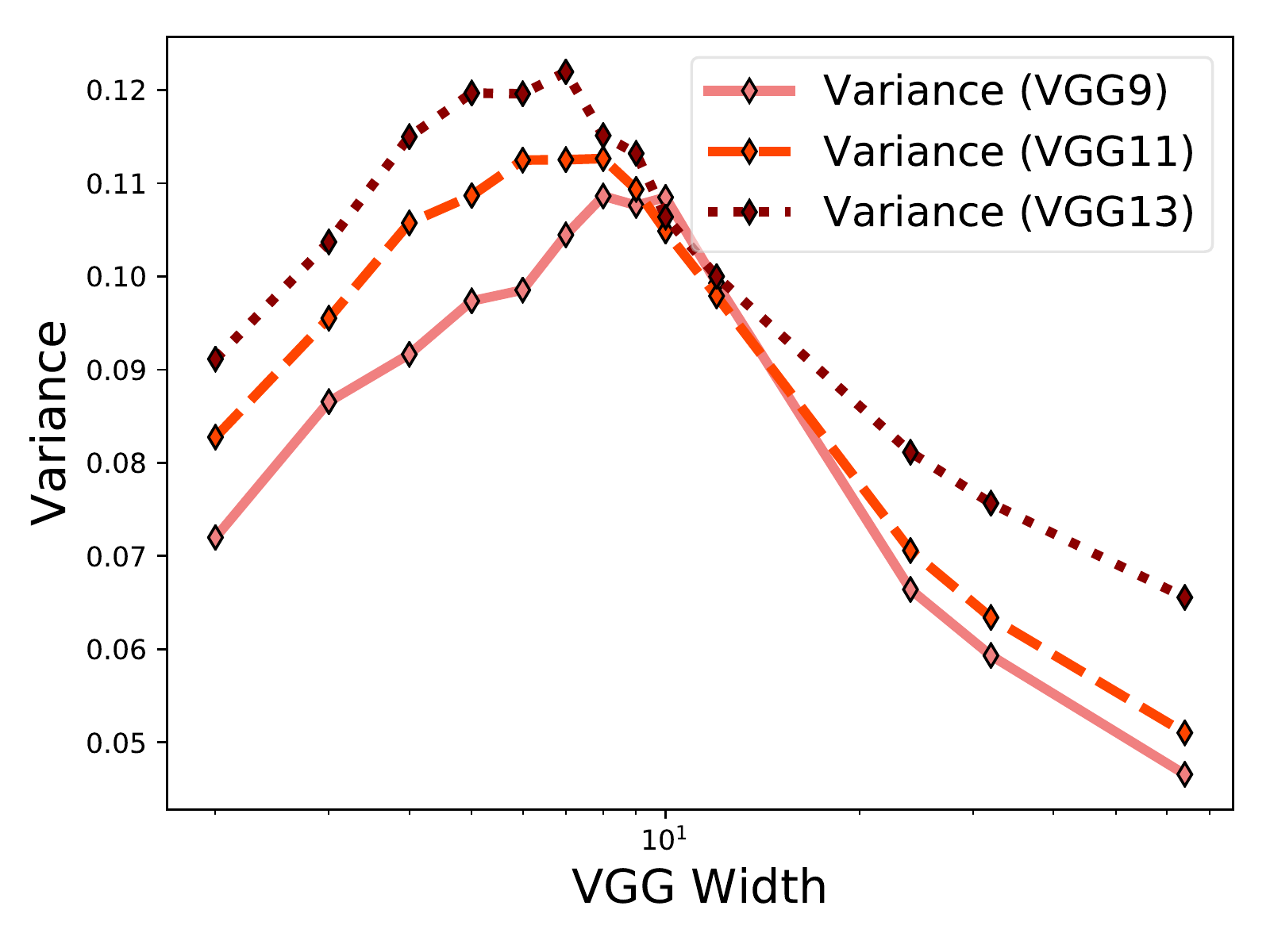}
    }
    \vskip -0.1in
    \caption{Bias and variance for VGG with different depth. (\textbf{Left}) Bias for VGG9, VGG11 and VGG13. (\textbf{Right}) Variance for VGG9, VGG11 and VGG13. }
    \label{fig:depth-vgg}
  \end{center}
  \vskip -0.2in
\end{figure}

We study the role of depth for VGG network on bias and variance.  As shown in Figure~\ref{fig:depth-vgg}, we observe that deeper VGG has lower bias and higher variance.

\subsection{Additional Synthetic Experiment}
\begin{figure}[h]
    \centering
    \includegraphics[width=.45\textwidth]{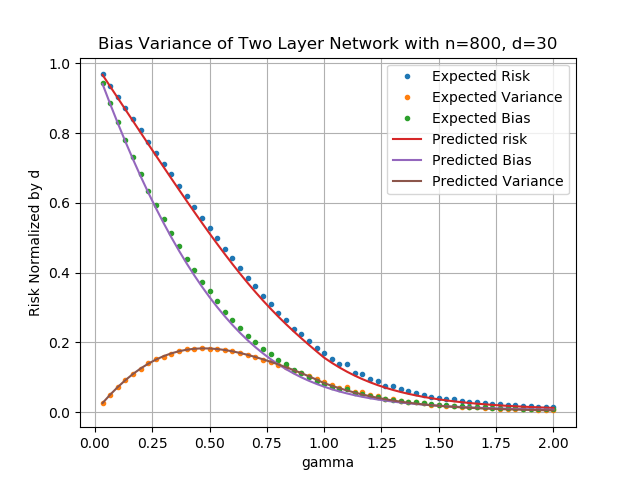}
    \vskip -0.1in
    \caption{Bias, Variance, and Risk for two layer linear network with parameters $n=800$ and $d=30$.}
    \label{fig:additional-simulation}
\end{figure}

In Figure~\ref{fig:additional-simulation}, we plot the result of performing regression on synthetic data using a two-layer linear fully connected linear network with varying width. The data are generated as $y={\mb\beta}^\transpose {\mb \x}$, $\x\sim\mc{N}(0, {\mb I}_d/d)$, where $\|\mb\beta\|_2=1$ is randomly generated and fixed weight vector. The first layer of the network is drawn from i.i.d. zero-mean Gaussian distribution with variance $1/\sqrt{d}$, and the second layer is trained using gradient descent with weight decay $0.1$. The horizontal axis is the number of parameters of the hidden layer normalized by the dimension of the data (i.e., $p/d$). The dots indicate actual experimental results, while the lines indicate theoretically predicted results. We can observe that they align well and the peak occurs at the predicted value.

\newpage
\clearpage
\section{Proof of Theorems in  \textsection\ref{section:two-layer}}\label{appendix:proof}
Throughout this section, we use $\|\cdot\|$ and $\|\cdot\|_2$ to denote the Frobenius norm and spectral norm of a matrix, respectively. Recall that for any given $\mb \theta$, the training set $\mc T = (\mb X, \mb y)$ satisfies the relation $\mb y = \mb X^\transpose \mb \theta$. 
By plugging this relation into \eqref{eq:exprf}, we get
\begin{equation}\label{eq:prf-f-M}
  f_\lambda(\mb x;\mc T, \mb W) = \mb x^\transpose \mb M_\lambda(\mc T, \mb W) \mb \theta, 
\end{equation}
where we define
\begin{equation}\label{eq: mmatrix}
  \mb M_\lambda(\mc T, \mb W) := \mb W^\transpose(\mb W\mb X\mb X^\transpose \mb W^\transpose+\lambda \mb I)^{-1}\mb W\mb X\mb X^\transpose.
\end{equation}
To avoid cluttered notations, we omit the dependency of $\mb M$ on $\lambda, \mc T$ and $\mb W$. 

By using \eqref{eq:prf-f-M}, the expected bias and expected variance in \eqref{eq:Ebias} and \eqref{eq:Evariance} can be written as functions on the statistics of $\mb M$.
This is stated in the following proposition. To proceed, we introduce the change of variable
\begin{equation*}
  \eta := \gamma^{-1} = \frac{d}{p}
\end{equation*}
in order to be consistent with conventions in random matrix theory.

\begin{proposition}[Expected Bias/Variance]\label{prop: biasvariance}
The expected bias and expected variance are given by
\begin{align*}
    \E \textbf{Bias}_\lambda^2 &= \frac{1}{d}\|\E \mb M -\mb I\|^2, ~\text{and}~\\
    \E \textbf{Variance}_\lambda &= \frac{1}{d}\E \|\mb M-\E \mb M\|^2,
\end{align*}
where $\mb M$ is defined in \eqref{eq: mmatrix}.
\begin{proof}
  By plugging \eqref{eq:prf-f-M} into \eqref{eq:Ebias}, and using the prior that $\mb x \sim \mc{N}(0, \mb I_d/d)$ and $\mb \theta \sim  \mc{N}(0, \mb I_d)$, we get
  \begin{align*}
    \E \textbf{Bias}_\lambda^2 &= \E\{ \E (\mb x^\transpose \mb M \mb \theta|\mb x, \mb \theta) -  \mb x^\transpose \mb\theta \}^2\\
    &= \E\left[\mb x^\transpose (\E \mb M-\mb I) \mb\theta \right]^2\\
    &=\E \mb x^\transpose (\E \mb M-\mb I) \mb\theta \mb\theta^\transpose (\E \mb M-\mb I) \mb x \\
    &= \E \text{tr}\Big[\mb x^\transpose (\E\mb M-\mb I) \mb\theta \mb\theta^\transpose (\E\mb M-\mb I)^\transpose \mb x\Big]\\
    & =  \text{tr}\Big[ (\E\mb M -\mb I) \E(\mb x\mb x^\transpose)(\E\mb M -\mb I)^\transpose \E(\mb\theta\mb\theta^\transpose) \Big]\\
    & =\frac{1}{d}\|\E\mb M -\mb I\|^2.
  \end{align*}
  Similarly, by plugging \eqref{eq:prf-f-M} into \eqref{eq:Evariance} we get
  \begin{align*}
    &~\E \textbf{Variance}_\lambda\\    
     = &~\E\Big\{\E \big[(\mb x^\transpose \mb M \mb\theta - \E (\mb x^\transpose \mb M\theta|\mb x, \mb\theta))^2|\mb x, \mb\theta\big] \Big\}\\
    =&~\E\Big\{\E\big[(\mb x^\transpose \mb M \mb\theta - \mb x^\transpose (\E \mb M)\mb\theta)^2 |\mb x, \mb\theta \big]\Big\}\\
    =&~ \E (\mb x^\transpose \mb M \mb\theta - \mb x^\transpose (\E \mb M)\mb\theta)^2 \\
    =&~ \E \Big[\mb x^\transpose(\mb M-\E\mb M)\mb\theta\Big]^2 \\
    =&~ \frac{1}{d}\E \|\mb M-\E\mb M\|^2.
  \end{align*}
\end{proof}
\end{proposition}
The risk is given by
\begin{equation*}
  \E \textbf{Bias}_\lambda^2 + \E \textbf{Variance}_\lambda = \frac{1}{d}\E \|\mb M-\mb I\|^2 = \frac{1}{d}\E\text{tr}(\mb M^\transpose \mb M) - \frac{2}{d}\E\text{tr}(\mb M) + 1.
\end{equation*}
First, we show that in the asymptotic setting defined in Assumption \ref{assumption:largesample}, the expected Bias and expected Variance can be calculated as functions on the statistics of the following matrix:
\begin{equation}\label{eq: mtmatrix}
\mb\mt_{\lambda_0}(\mb W) = \mb W^\transpose(\mb W\mb W^\transpose + \lambda_0\mb I)^{-1}\mb W.
\end{equation}
In the following, we omit the dependency of $\mb\mt$ on $\lambda_0$ and $\mb W$.
\begin{proposition}[Gap between $\mb M$ and $\mb\mt$] \label{prop: gap}
Under Assumption \ref{assumption:largesample} with $\lambda - \frac{n}{d}\lambda_0$, we have
\begin{align*}\label{eq: nogap}
  \frac{1}{d}\|\E \mb M -\mb I\|^2 &= \frac{1}{d}\|\E \mb\mt -\mb I\|^2,~\text{and}~\\
  \frac{1}{d}\E\|\mb M -\mb I\|^2 &= \frac{1}{d}\E\|\mb\mt -\mb I\|^2.
\end{align*}

\begin{proof}
  It suffices to show that $\|\mb M-\mb\mt\|_2=0$ almost surely. 
  From \eqref{eq: mmatrix} and \eqref{eq: mtmatrix}, we have
  \begin{equation*}
    \mb M-\mb\mt = \mb W^\transpose \mb\Omega \mb W +  \mb W^\transpose \mb\Omega \mb W\mb \Delta+ \mb W^\transpose(\mb W\mb W^\transpose+\lambda_0 \mb I)^{-1}\mb W\mb \Delta,
  \end{equation*}
  where $\mb\Delta :=  (d/n)\mb X\mb X^\transpose - \mb I$ and $\mb\Omega := (\mb W\mb W^\transpose+\lambda_0\mb I+\mb W\mb \Delta \mb W^\transpose)^{-1}-(\mb W\mb W^\transpose+\lambda_0\mb I)^{-1}.$
    
  By using triangle inequality and the sub-multiplicative property of spectral norm, we have
  \begin{equation}\label{eq:prf-gap-norm}
    \|\mb M-\mb\mt\|_2 \leq  \|\mb W\|_2^2 \cdot \|\mb\Omega\|_2 + \|\mb W\|_2^2 \cdot \|\mb\Omega\|_2 \cdot \|\mb\Delta\|_2 +\|\mb\mt\|_2 \cdot \|\mb\Delta\|_2.
  \end{equation}
  Furthermore, by a classical result on the perturbation of matrix inverse (see e.g., \citet[equation $(1.1)$]{ELGHAOUI2002171}), we have
  $$
    \|\mb\Omega\|_2 \leq \|(\mb W\mb W^\transpose+\lambda_0\mb I)^{-1}\|_2^2\|\mb W\|_2^2\|\mb\Delta\|_2 + O(\|\mb\Delta\|_2^2).
  $$
  Combining this bound with \eqref{eq:prf-gap-norm} gives
  \begin{equation*}
    \|\mb M-\mb\mt\|_2 \leq \|\mb W\|_2^4 \cdot \|(\mb W\mb W^\transpose+\lambda_0\mb I)^{-1}\|_2^2 \cdot\|\mb\Delta\|_2 + \|\mb\mt\|_2 \cdot \|\mb\Delta\|_2 + O(\|\mb\Delta\|_2^2).
  \end{equation*}
  It remains to show that $\|\mb\Delta\|_2 = 0$ and that $\|\mb W\|_2$, $\|(\mb W\mb W^\transpose+\lambda_0\mb I)^{-1}\|_2^2$, and $\|\mb\mt\|_2$ are bounded from above almost surely. By \citet[Example 6.2]{wainwright_2019}, $\forall \delta>0$ and $n>d$, 
  \begin{equation*}
    \mathbb{P}\Big(\|\mb\Delta\|_2\leq2\epsilon+\epsilon^2\Big)\geq 1-e^{-n\delta^2/2}, ~\text{where}~\epsilon = \delta + \sqrt{\frac{d}{n}}.
    \end{equation*}
  By letting $\delta = \sqrt{d/n}$ and taking the asymptotic limit as in Assumption \ref{assumption:largesample}, we have 
  \begin{equation*}
    \|\mb\Delta\|_2\overset{\text{a.s.}}{=}0.    
  \end{equation*}
  From \citet{geman1980}, the largest eigenvalue of $\mb W\mb W^\transpose$ is almost surely $(1+\sqrt{\eta})^2<\infty$. Therefore, we have
  $$
    \|\mb W\|_2 \overset{\text{a.s.}}{=} 1+\sqrt{\eta}< \infty.
  $$
  Finally, note that
  \begin{equation*}
      \|(\mb W\mb W^\transpose+\lambda_0 \mb I)^{-1}\|_2 \le \frac{1}{\lambda_0 + \sigma_{\min}(\mb W)^2} \le \frac{1}{\lambda_0} < \infty,
  \end{equation*}
  \begin{equation*}
      \|\mb\mt\|_2 = \frac{\sigma_{\max}(\mb W)^2}{\sigma_{\max}(\mb W)^2 + \lambda_0} \le 1.
  \end{equation*}
  We therefore conclude that $\|\mb M-\mb\mt\|_2=0$ almost surely, as desired.
\end{proof}
\end{proposition}
\begin{proposition}[Asymptotic Risk]\label{prop: largesamplerisk}
  Given the expression for Bias and Variance in Proposition \ref{thm:largesample}, under the asymptotic assumptions from Assumption \ref{assumption:largesample}, 
  \begin{equation*}
      \frac{1}{d}\E\|\mb \mt- \mb I\|^2 =
        \begin{cases}
             (1-\frac{1}{\eta })+ f_{\lambda_0^{-1}}(\frac{1}{\eta}),   & \text{if $d>p$}, \\
             f_{\lambda_0^{-1}}(\eta), & \text{if $d\leq p$},
        \end{cases}
  \end{equation*}
  where $\eta = d/p$, and for any $\eta, \alpha\in \R$,
  \begin{equation*}
      f_{\alpha}(\eta)= \frac{\alpha+\eta(1+\eta-2\alpha+\eta\alpha)}{2\eta\sqrt{\eta^2+2\eta\alpha(1+\eta)+\alpha^2(1-\eta)^2}} - \frac{1-\eta}{2
      \eta}.
  \end{equation*}
  \begin{proof}
      Recall that $\mb \mt = \mb W^\transpose(\mb W\mb W^\transpose+\lambda_0 \mb I)^{-1}\mb W$, by Sherman-Morrision,
      \begin{equation*}
          \mb\mt = \mb I-(\mb I+\lambda_0^{-1}\mb W^\transpose \mb W)^{-1},
      \end{equation*}
      where $(d/p)\mb W^\transpose \mb W\in\R^{d\times d}$. Let $\lambda_i\geq0, i=1,\dots,d$ be the eigenvalues of $(d/p)\mb W^\transpose \mb W$. For notational simplicity, let $\alpha=\lambda_0^{-1}$. Then
    \begin{align*}
      \|\mb\mt-\mb I\|^2 &= \|[\mb I+(\alpha/\eta)(d/p)\mb W^\transpose \mb W]^{-1}\|^2 =\sum_{i=1}^d \frac{1}{(1+\frac{\alpha}{\eta}\lambda_i)^2}.
    \end{align*}   
      \noindent Let $\mb A = (d/p)\mb W^\transpose \mb W$, and $\mu_{\mb A}$ be the spectral measure of $\mb A$. Then
      \begin{equation*}
          \frac{1}{d}\|\mb \mt-\mb I\|^2 = \int_{\R^+} \frac{1}{(1+\frac{\alpha}{\eta} x)^2} d\mu_{\mb A}(dx).
      \end{equation*}
      According to Marchenko-Pastur Law \cite{baibook}, in the limit when $d\rightarrow\infty$ when $\eta\leq1$,
      \begin{equation*}
          \frac{1}{d}\|\mb \mt-\mb I\|_F^2 \overset{\text{a.s.}}{=} \frac{1}{2\pi}\int_{\eta_-}^{\eta_+}\frac{\sqrt{(\eta_+-x)(x-\eta_-)}}{\eta x(1+\frac{\alpha}{\eta} x)^2} dx,
      \end{equation*}
      where $\eta_+ = (1+\sqrt{\eta})^2$, and $\eta_- = (1- \sqrt{\eta})^2$. For convenience, define
      \begin{align*}
          f_\alpha(\eta) = \frac{1}{2\pi}\int_{\eta_-}^{\eta_+}\frac{\sqrt{(\eta_+-x)(x-\eta_-)}}{\eta x(1+\frac{\alpha}{\eta} x)^2} dx =\frac{\alpha+\eta(1+\eta-2\alpha+\eta\alpha)}{2\eta\sqrt{\eta^2+2\eta\alpha(1+\eta)+\alpha^2(1-\eta)^2}} - \frac{1-\eta}{2\eta}.
      \end{align*}
      When $\eta>1$,
      \begin{equation*}
          \frac{1}{d}\|\mb \mt-\mb I\|_F^2 =\left(1-\frac{1}{\eta }\right)+ f_\alpha\left(\frac{1}{\eta}\right)\\.
      \end{equation*}
      Then, in the asymptotic regime, 
      \begin{equation*}
        \frac{1}{d}\|\mb \mt-\mb I\|_F^2 \overset{\text{a.s.}}{=}
            \begin{cases}
              (1-\frac{1}{\eta })+ f_\alpha(\frac{1}{\eta}),  & \text{if $d>p$}, \\
               f_\alpha(\eta), & \text{if $d<p$}. 
          \end{cases}
      \end{equation*}
  \end{proof}
\end{proposition}

\begin{proposition}[Asymptotic Bias]
  Given the expression for Bias in Proposition \ref{thm:largesample}, under the asymptotic assumptions in Assumption \ref{assumption:largesample}, the Bias for the model is given by
  \begin{align*}
    \frac{1}{d}\|\E \mb M-\mb I\|^2 = \Big[1-\frac{\lambda_0\eta+(1+\eta)-\sqrt{\lambda_0^2\eta^2+2\lambda_0\eta(1+\eta)+(1-\eta)^2}}{2\eta}\Big]^2.
  \end{align*}
  \begin{proof}
    Recall that
    \begin{equation*}
     \mb M = \mb W^\transpose(\mb W\mb X\mb X^\transpose \mb W^\transpose+\lambda \mb I)^{-1}\mb W\mb X\mb X^\transpose.
    \end{equation*}
    Recall that $\mb \mt = \mb I-(\mb I+\lambda_0^{-1}\mb W^\transpose \mb W)^{-1}$. Thus
    \begin{equation*}
        \frac{1}{d}\|\E\mb \mt-\mb I\|^2 = \frac{1}{d}\|\E(\mb I+\lambda_0^{-1}\mb W^\transpose \mb W)^{-1}\|^2.
    \end{equation*}
    By Neumann series,
    \begin{align*}
      \E(\mb I+\lambda_0^{-1}\mb W^\transpose \mb W)^{-1} = \sum_{m\geq 0}\E(-\lambda_0^{-1} \mb W^\transpose \mb W)^m = \mb I + \sum_{m\geq 1}(-1)^m(\lambda_0\eta)^{-m}\E \mb A^m,
    \end{align*}
    where $\eta = {d}/{p}, \mb A=(d/p)\mb W^\transpose \mb W$. According to Corollary 3.3 in \citet{bishop2018} (recall we are considering the asymptotic regime of $d, p\rightarrow\infty$), 
    \begin{equation*}
      \E \mb A^m = \sum_{k=1}^{m} \eta^{m-k}N_{m, k}\cdot \mb I,
    \end{equation*}
    where $$N_{m, k}=\frac{1}{k}\binom{m-1}{k-1}\binom{m}{k-1}$$ is the Narayana number. Therefore,
    \begin{equation*}
      \frac{1}{d}\|\E\mb \mt-\mb I\|^2 = \Big(1+\eta^{-1}\sum_{m=1}^\infty\sum_{k=1}^k (-\lambda_0^{-1})^m(\eta^{-1})^{k-1}N_{m, k}\Big)^2.
    \end{equation*}
    Observe that the double sum in the previous equation is just the generating series for the Narayana number,
    \begin{align*}
        \sum_{m=1}^\infty\sum_{k=1}^k (-\lambda_0^{-1})^m(\eta^{-1})^{k-1} N_{m, k} = -\frac{\lambda_0\eta+(1+\eta)-\sqrt{\lambda_0^2\eta^2+2\lambda_0\eta(1+\eta)+(1-\eta)^2}}{2\eta}.
    \end{align*}
    This completes the proof.
\end{proof}
\end{proposition}

Finally, the statement of Theorem \ref{thm:largesample} follows directly from the above propositions.


\end{document}